%% file: acl_latex.tex
\documentclass[11pt]{article}

\usepackage[preprint]{acl}

\usepackage{times}
\usepackage{latexsym}

\usepackage[T1]{fontenc}

\usepackage[utf8]{inputenc}

\usepackage{microtype}

\usepackage{inconsolata}

\usepackage{graphicx}

\usepackage{amsmath,amssymb,amstext} 
\usepackage{bm}
\usepackage{booktabs}
\usepackage{enumitem}
\usepackage{glossaries-extra}
\usepackage{makecell}
\usepackage{mathtools}
\usepackage{multirow}
\usepackage{listings}
\usepackage{siunitx}
\usepackage{tabularray}
\usepackage{tabularx}

\definecolor{codegray}{gray}{0.9}
\definecolor{codeblue}{rgb}{0.0, 0.0, 0.5}

\newcommand{\bftab}{\fontseries{b}\selectfont}
\newcommand{\alignedminus}{\mathmakebox[\widthof{+}]{-}}

\title{UniMaia: Steering Chess Policies with Language for Human-like Play}

\author{
 \textbf{Sherman Siu\textsuperscript{1}},
 \textbf{Lesley Istead\textsuperscript{1,2}}
\\
 \textsuperscript{1}University of Waterloo,
 \textsuperscript{2}Carleton University,
\\
 \small{
   \textbf{Correspondence:} \href{mailto:s8siu@uwaterloo.ca}{s8siu@uwaterloo.ca}
 }
}

\begin{document}
\maketitle
\begin{abstract}
Recent advances in large language models have enabled natural language to serve as a flexible interface for controlling complex systems, but often at the cost of large-scale multimodal training or weakened domain-specific inductive biases. In structured decision-making domains such as chess, specialized policy networks achieve strong performance but lack semantic controllability, while prompt-conditioned language models are more flexible yet typically exhibit weaker domain grounding. We propose \textbf{UniMaia}, a framework for prompt-conditioned policy modulation that adapts a frozen Lc0-based chess policy network using a parameter-efficient text encoder and a ControlNet-style conditioning mechanism. UniMaia enables semantic control over gameplay, including opening selection and player strength, while preserving the pretrained policy representations. We further introduce \textbf{UniMaia-Aux}, which incorporates auxiliary temporal conditioning and behavioral prediction objectives. To support this work, we construct a large-scale metadata-augmented Lichess dataset, develop a semi-automated prompt-generation pipeline, and introduce benchmarks spanning both prompt-conditioned and metadata-conditioned settings. UniMaia achieves state-of-the-art expected accuracy on several prompt-conditioned benchmarks and competitive top-move accuracy on general instruction-following tasks, while remaining competitive with dedicated metadata-conditioned approaches on human move prediction benchmarks. UniMaia-Aux further improves expected accuracy and behavioral modeling across several evaluation settings, with modest trade-offs in top-move accuracy. Overall, our results demonstrate that prompt-conditioned control of domain-specific policy networks is feasible without end-to-end multimodal training, while highlighting trade-offs between controllability and predictive performance.
\end{abstract}

\section{Introduction}
\label{sec:introduction}

Natural language has become an increasingly flexible interface for controlling machine learning systems. Recent advances in large language models have enabled prompt-based interaction across a wide range of tasks, including reasoning, planning, and tool use. However, in structured decision-making domains, purely language-based approaches often struggle to match the performance and inductive biases of specialized policy models. This creates a tension between \textbf{controllability} and \textbf{domain grounding}: prompt-conditioned systems are highly flexible and steerable, while domain-specific policy networks achieve stronger task performance but typically rely on rigid structured conditioning.

Chess provides a useful testbed for studying this trade-off. Prior work on human-like chess modeling has largely focused on metadata-conditioned policy networks that predict human moves from board states and structured signals such as player rating or time control. These approaches achieve strong move-prediction performance but limited semantic controllability. In contrast, language-based chess models can follow natural language instructions but often exhibit weaker domain grounding and lower move-prediction accuracy. End-to-end multimodal approaches provide a unified alternative, but typically require substantial compute and large-scale curated datasets.

In this work, we investigate whether natural language prompts can be used to steer a pretrained chess policy while preserving the inductive biases of a strong domain-specific model. We introduce \textbf{UniMaia}, a framework for \textbf{prompt-conditioned modulation of pretrained chess policies}. UniMaia combines a frozen Lc0-based policy network with a parameter-efficient text encoder and a ControlNet-style conditioning mechanism, enabling natural language control over gameplay behaviors such as opening selection and player strength while retaining the underlying chess representations of the base model.

To support this setting, we construct a large-scale metadata-augmented Lichess dataset and a prompt-generation pipeline for diverse natural language conditioning templates. We further introduce evaluation settings spanning both \textbf{prompt-conditioned} and \textbf{metadata-conditioned} benchmarks, enabling systematic comparison between language-conditioned and structured-conditioning approaches.

UniMaia achieves strong performance across both benchmark families. The model outperforms prior prompt-conditioned baselines on instruction-following and opening-control tasks while remaining competitive with metadata-conditioned approaches on human move prediction. We further show that the learned policy varies continuously with prompted Elo, suggesting that lightweight prompt-conditioned modulation can provide semantic control without discarding domain-specific inductive biases.

Our contributions are as follows:
\begin{enumerate}
    \item We propose UniMaia, a parameter-efficient framework for prompt-conditioned modulation of pretrained chess policy networks.

    \item We introduce a large-scale metadata-augmented chess dataset and prompt-generation pipeline for language-conditioned policy learning.

    \item We evaluate prompt-conditioned and metadata-conditioned policy models under a unified benchmark framework spanning instruction following, opening control, and human move prediction.

    \item We demonstrate that lightweight prompt-conditioned modulation enables controllable human-like chess behavior while remaining competitive with structured-conditioning approaches.
\end{enumerate}

\section{Background and Related Work}
\label{sec:background_related}

\paragraph{Chess Policy Networks.}
Modern neural chess engines model gameplay as policy learning over structured board representations. AlphaZero~\cite{silver2017alphazero} demonstrated that deep neural networks trained through self-play reinforcement learning could achieve superhuman performance when combined with Monte Carlo Tree Search (MCTS). Leela Chess Zero (Lc0), an open-source reproduction of AlphaZero for chess, later introduced stronger transformer-based ChessFormer (CF) architectures~\cite{monroe2024mastering,monroe2026chessformer}. These models encode the current board state together with a fixed history of previous positions using spatial feature planes and predict move probabilities over a fixed legal move space. Unlike language-only chess models, Lc0-style architectures incorporate strong domain-specific inductive biases through structured board encodings and specialized policy heads, and have proven highly effective for both strong play and human gameplay modeling~\cite{mcilroy2020aligning_maiachess,monroe2026chessformer,feng2024chessgpt}.

\paragraph{Human Move Prediction and Metadata Conditioning.}
Prior work on human move prediction adapts chess policies to model human gameplay distributions rather than optimal play. Maia~\cite{mcilroy2020aligning_maiachess} fine-tuned Lc0 CNN-based networks on human games across rating ranges, while Maia-2~\cite{tang2024maia2} extended this approach with unified skill conditioning using rating embeddings. Allie~\cite{zhang2024humanallie} further incorporated interpolated skill representations, time controls, and auxiliary behavioral objectives. Although effective, these approaches rely primarily on structured metadata rather than flexible natural-language control.

\paragraph{Prompt-Conditioned Chess Models.}
Other approaches instead model chess directly as a language task. ChessGPT~\cite{feng2024chessgpt} trains language models for chess generation and reasoning using specialized chess tokenization, while ChessCLIP~\cite{feng2024chessgpt} learns joint prompt--board representations. Additional work has explored large language models for gameplay and chess generation~\cite{karvonen2024emergent,zhang2024transcendence,haileystorm2024mambachess,carlini2023PlayingChess}. While these models provide flexible semantic control, they generally lack the structured inductive biases of specialized chess policy networks.

\paragraph{Multimodal Conditioning and Control.}
Our approach is also related to multimodal conditioning methods that adapt pretrained models using external signals. Cross-attention conditioning is widely used in vision--language and diffusion models~\cite{alayrac2022flamingo,liu2023visualllava,rombach2022high_stablediffusion}. In particular, ControlNet~\cite{zhang2023adding_controlnet,chen2024pixartdelta_controlnet_transformer} introduces lightweight conditioning through zero-initialized residual pathways while preserving a frozen base model. UniMaia applies a similar principle to chess by conditioning a pretrained policy network on natural-language prompts while preserving its underlying structured representations.

\section{Dataset and Prompt Construction}
\label{sec:dataset}

We construct a large-scale prompt-conditioned chess dataset from the Lichess Open Database~\cite{lichess-open-database}, designed to support natural-language control of pretrained chess policies.

\subsection{LichessGames}
\label{sec:lichess_games}

Training data consists of publicly available Lichess PGN archives from 2013--2023, totaling approximately 5.2 billion games. We process the raw PGNs into a structured Parquet dataset, \textsc{LichessGames}, containing move histories, player ratings, time controls, openings, termination details, and game metadata.

To support controllable policy learning, we augment games with normalized opening annotations and canonical opening move sequences. Because Lichess opening names and ECO classifications changed over time, we align historical annotations with the current Lichess opening database~\cite{github-lichess-org-chessopenings} using rule-based normalization, fuzzy matching, and manual verification. For each game, we recover the longest opening prefix consistent with both the annotated opening and played moves.

Games are additionally partitioned by rating range to improve sampling coverage across skill levels.

\subsection{Prompt Template Construction}
\label{sec:prompt_template_construction}

Rather than generating a unique prompt for every game, we construct synthetic natural-language templates instantiated from game metadata. This enables large-scale prompt-conditioned training without requiring human annotation for billions of examples.

Templates are generated using GPT-4 Turbo~\cite{achiam2023gpt4} and converted into parameterized Jinja2 templates through a post-processing pipeline. Metadata such as player ratings, titles, time controls, openings, and opening move sequences are inserted programmatically during training.

Template generation follows an iterative human-in-the-loop process: generated prompts are manually reviewed, and the conversion pipeline is refined as needed. Existing templates are updated accordingly, while later revisions prioritize backward compatibility over regeneration.

Although the templates are synthetically generated, the effective prompt space is substantially larger than the number of base templates due to combinatorial variation in metadata inclusion, prompt framing, opening descriptions, player attributes, time controls, and sampled game metadata. As a result, instantiated prompts often differ considerably in structure and lexical content despite sharing the same template family.

We construct two complementary template families: \textsc{LichessTemplates-Pretrain} and \textsc{LichessTemplates-Instruct}.

\paragraph{Pretraining templates.}
\textsc{LichessTemplates-Pretrain} consists of 160 metadata-rich templates designed to expose the model to diverse chess metadata during training. These templates include combinations of player skill levels, time controls, opening descriptions, and varying prompt verbosity, producing prompts that resemble descriptive game summaries while preserving most available metadata.

\paragraph{Instruction templates.}
\textsc{LichessTemplates-Instruct} consists of 2,048 instruction-oriented templates designed to encourage controllability under partial or naturalistic specifications. These templates vary whether player names, titles, ratings, skill descriptions, openings, and opening move sequences are included, and whether prompts are phrased as requests or game descriptions.

Unlike the pretraining templates, the instruction templates intentionally omit portions of the metadata, requiring the model to infer plausible gameplay behavior from incomplete natural-language specifications. For example, templates vary across \(2^{10}\) metadata inclusion configurations in addition to lexical and game-level variation.

During training, examples are formatted using either template family with equal probability.

\section{Method}
\label{sec:method}

We propose \textbf{UniMaia}, a prompt-conditioned chess policy model that adapts a pretrained Lc0 policy network using natural language prompts. Rather than training a language model to play chess end-to-end, UniMaia conditions a frozen domain-specific chess policy through lightweight residual modulation.

\subsection{Architecture}
\label{sec:architecture}

Given a board state \(s\) and natural language prompt \(p\), the model predicts a distribution over legal chess moves. UniMaia consists of:
(i) a frozen Lc0-CF backbone,
(ii) a LoRA-adapted ChessGPT-base text encoder~\cite{hu2022lora},
and (iii) a ControlNet-style conditioning branch that injects prompt-conditioned residual updates into the policy computation.

\begin{figure}[t]
    \centering
    \includegraphics[width=0.7\linewidth]{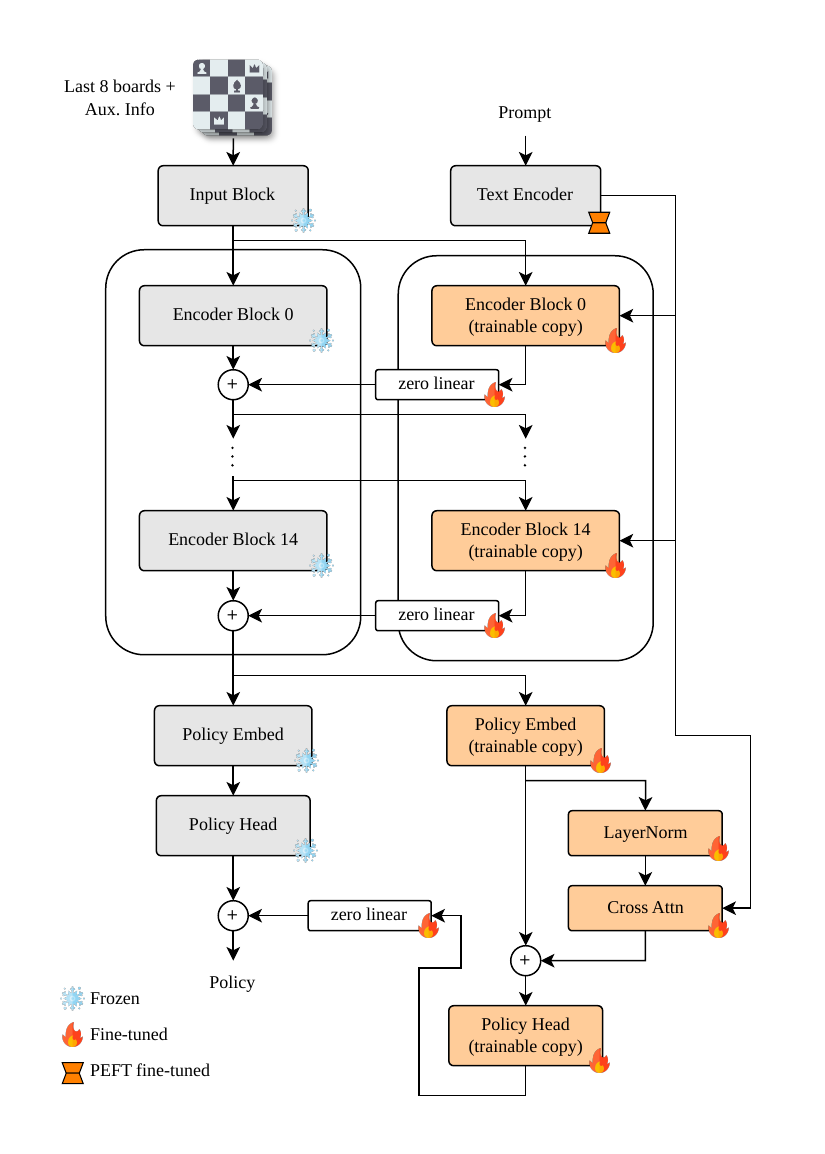}
    \caption{UniMaia architecture. A frozen Lc0-CF backbone encodes the board state, while a ControlNet-style branch incorporates prompt embeddings through cross-attention. Both the frozen backbone layer and the conditioning layer at depth \(n\) receive the residual-updated representation from layer \(n-1\).}
    \label{fig:method_architecture}
\end{figure}

Figure~\ref{fig:method_architecture} illustrates the architecture.

UniMaia builds on a pretrained Lc0-CF policy network initialized from BT4 weights. The backbone remains frozen during training and maps board states \(s\) to hidden representations via a sequence of transformer layers \(B_n\), a policy embedder \(\mathrm{PE}\), and a policy head \(\mathrm{PH}\).

The natural language prompt \(p\) is encoded by a text encoder
\[
e = T_\phi(p),
\]
which produces a sequence of prompt embeddings. To keep adaptation lightweight, the text encoder is fine-tuned using LoRA adapters together with rsLoRA and LoRA+~\cite{kalajdzievski2023rankrslora,hayou2024lora+}.

UniMaia conditions the frozen chess policy using a layer-coupled ControlNet-style branch. Both branches receive the residual-updated representation from the previous layer:
\[
h_n = B_n(h_{n-1} + r_{n-1}), \;
\tilde{h}_n = C_n(h_{n-1} + r_{n-1}, \; e),
\]
where \(C_n\) includes self-attention, cross-attention over the prompt embeddings \(e\), and a feed-forward block, and \(h_{-1} + r_{-1}\) denotes the output of the frozen input block (with \(r_{-1} = 0\)). The residual update is then
\[
r_n = \mathcal{Z}_n(\tilde{h}_n),
\]
projected through a zero-initialized layer \(\mathcal{Z}_n\), ensuring the model is initially equivalent to the pretrained backbone.

After the final layer, UniMaia applies prompt modulation directly to the intermediate policy representation using a \emph{controllable policy head}. The controllable policy head consists of a cloned policy embedder \(\widehat{\mathrm{PE}}\), a pre-LN cross-attention layer over the prompt embeddings, and a cloned policy head \(\widehat{\mathrm{PH}}\).

A cloned policy embedder produces
\[
z = \widehat{\mathrm{PE}}(h_N + r_N),
\]
which is passed through the controllable policy head and a final zero-initialized projection:
\[
r_\pi = \mathcal{Z}_\pi\!\left(\widehat{\mathrm{PH}}\!\left(\mathrm{CrossAttn}(\mathrm{LN}(z),\, e)\right)\right).
\]

The final policy logits are \(\mathrm{PH}(\mathrm{PE}(h_N + r_N)) + r_\pi\), over which we apply softmax to obtain the move distribution \(\pi(a \mid s, p)\).

\subsection{Training}
\label{sec:training}

UniMaia is trained on supervised move prediction using human games from the Lichess dataset, with only the prompt encoder adapters and conditioning branch updated; the Lc0 backbone remains frozen throughout. Training data consists of games from 2013--2022, while games from 2023 are held out for evaluation. Monthly Lichess partitions are streamed from disk or Hugging Face. Following the \textsc{LichessGames} partitioning scheme described in Section~3.1, we first apply Unimax downsampling~\cite{chung2023unimax} across the Elo partitions to improve coverage across skill levels while allocating a fixed training budget. Within each sampled Elo partition, games are then grouped by time control, and Unimax is applied again independently to balance the distribution of time controls without allowing dominant groups to overwhelm the dataset.

Prompts are drawn from the \textsc{LichessTemplates-Pretrain} and \textsc{LichessTemplates-Instruct} template families with equal probability. Unless otherwise specified, prompts may include opening information, player ratings, time controls, and partial move histories. Full benchmark details, prompting strategies, auxiliary evaluations, and ablations are provided in the appendix.

\textbf{UniMaia-Aux} denotes an additional fine-tuning stage using temporal metadata and auxiliary prediction targets, including game outcome, termination type, remaining plies, and move delay. Further details appear in Appendix~\ref{app:unimaia_aux_final_configuration}.

\subsection{Design Evolution}
\label{sec:design_evolution}

UniMaia was developed through an iterative sequence of architectural, optimization, and data-centric modifications. Rather than presenting a fully factorial ablation study, we summarize the primary design changes that led to the final model in Table~\ref{tab:design_evolution}. Each row represents a modification to the previous configuration, beginning with an early-fusion baseline containing a single cross-attention layer.

For model development, we used a small diagnostic benchmark suite consisting of opening-following, instruction-following, and human move prediction tasks. We report the \textit{development score} (Dev score), defined as the mean of the normalized accuracies across these tasks.

Each ablation was trained on 2013 data. We report the mean training loss over the December 2013 partition, denoted \(\mu_{\mathrm{loss},\,\mathrm{Dec13}}\).

\begin{table}[t]
\centering
\small
\begin{tabular}{lcc}
\toprule
\textbf{Change} & \textbf{\(\Delta\) Dev Score} & \textbf{\(\Delta \mu_{\mathrm{loss},\,\mathrm{Dec13}}\)} \\
\midrule
ControlNet + policy head & \bftab{+0.0684} & \bftab{\(\alignedminus\)0.1005} \\
NorMuon optimizer & +0.0110 & \underline{\(\alignedminus\)0.0349} \\
LoRA vs DoRA & +0.0016 & \(\alignedminus\)0.0049 \\
BT4 backbone & +0.0059 & \(\alignedminus\)0.0219 \\
Unimax sampling & \underline{+0.0195} & \(\alignedminus\)0.0295 \\
+ LoRA rank 16 & +0.0017 & +0.0011 \\
\midrule
Full configuration & \bftab{+0.1061} & \bftab{\(\alignedminus\)0.1811}\\
\bottomrule
\end{tabular}
\caption{Sequential design improvements during model development. Each row modifies the previous configuration and should therefore be interpreted as a development trace rather than an independent ablation.}
\label{tab:design_evolution}
\end{table}

The largest improvement arises from replacing the early-fusion baseline with a layer-coupled ControlNet architecture and controllable policy head, suggesting that explicit residual modulation of the pretrained chess policy is substantially more effective than shallow cross-attention conditioning. Data-centric modifications, particularly Unimax sampling across time controls, also provide substantial gains, while later scaling decisions exhibit diminishing returns.

Final evaluation results on the complete benchmark suite are reported in Section~\ref{sec:results}. Additional ablations, parameter analyses, training curves, and negative results are provided in Appendix~\ref{app:ablation_study}.

\section{Experimental Setup}
\label{sec:experimental_setup}

We evaluate UniMaia on benchmarks spanning prompt-conditioned and metadata-conditioned settings. Prompt-conditioned benchmarks measure adaptation to natural-language instructions, while metadata-conditioned benchmarks evaluate human move prediction given structured context such as player rating and time control. Across all settings, the primary objective is agreement with human moves.

\paragraph{Prompt-conditioned benchmarks.}
We evaluate opening controllability by creating the Lichess Openings Benchmark (LOB), which pairs positions with natural-language prompts describing target openings. We report results on two variants: LOB-Partial (LOB-P; 31{,}169 positions), where prompts contain only partial opening information up to the current position, and LOB-Canonical (LOB-C; 32{,}736 positions), where prompts additionally include the canonical continuation of the opening. LOB-P provides a more realistic and challenging setting, while LOB-C measures upper-bound opening-following performance.

To evaluate general instruction following, we introduce the Lichess Instruction-Following Benchmark (LIF; 25{,}000 positions), constructed from Lichess games formatted with diverse prompt templates. LIF evaluates adaptation to natural-language descriptions of player strength, time controls, and gameplay context.

We additionally introduce the Lumbra's Gigabase Benchmark (LGB; 25{,}000 positions), constructed from curated over-the-board and online games from Lumbra's Gigabase~\cite{lumbrasgigabaseFreeChess}. We use the OTB 2020--2024 subset to evaluate generalization beyond the Lichess training distribution. Compared to LIF, LGB contains games from a broader range of tournaments, federations, and historical sources, providing a stronger out-of-distribution evaluation.

Finally, we define temporally augmented variants of these benchmarks, denoted LIF-Aux and LGB-Aux, which append metadata describing the time control, remaining clock time, and previous-move delay when available. These benchmarks additionally evaluate the auxiliary behavioral targets introduced during auxiliary-target fine-tuning.

\paragraph{Metadata-conditioned benchmarks.}
To compare against prior human imitation models, we evaluate on metadata-conditioned human move prediction benchmarks. We report results on the Maia-1 Subset (M1-S; 11{,}225 positions), the full Maia-1 benchmark (M1; 11.3M positions)~\cite{mcilroy2020aligning_maiachess}, the Maia-2 Rapid Benchmark (M2R; 127{,}852 positions)~\cite{tang2024maia2}, and the Allie Blitz Benchmark (ABB)~\cite{zhang2024humanallie}.

M1-S is a stratified subset of the Maia-1 benchmark that we construct by grouping positions by player rating and time control. These benchmarks evaluate how well models predict human moves when contextual information is provided explicitly rather than through natural-language prompts. Additional benchmark details and model scores are provided in the appendix.

\paragraph{Comparability across model classes.}
Prompt-conditioned benchmarks use randomly sampled templates with varying metadata, whereas metadata-conditioned benchmarks use fixed templates with consistent metadata fields. Consequently, comparisons are most meaningful within each conditioning regime. Metadata-conditioned models may receive metadata absent from prompts, while prompts may contain information unavailable to metadata-conditioned models. Appendix~\ref{app:model_comparison} compares model inputs.

\paragraph{Evaluation metrics.}
We report both top-move accuracy (Acc@1) and expected accuracy (\(\mathbb{E}[\mathrm{Acc}]\)). Expected accuracy measures the probability assigned to the ground-truth move and is reported only for models that produce normalized move distributions. For auxiliary-target evaluations, we additionally report weighted F1 (\(\mathrm{F1}_{w}\)) for categorical targets.

Following prior work~\cite{mcilroy2020aligning_maiachess,zhang2024humanallie}, we additionally report results with the first 10 plies excluded on ABB and M1-S to focus on general gameplay. For LIF and LGB, we include all plies, since some prompts explicitly target opening play.

\paragraph{Baselines.}
We compare UniMaia against metadata-conditioned human imitation models, including Maia-1, Maia-2, and Allie, as well as prompt-conditioned approaches including ChessGPT and GPT-3.5 Turbo Instruct. We additionally compare against pretrained Lc0-CF policy networks.

For autoregressive language models, generated moves are greedily decoded, constrained to valid chess formats where possible, and illegal moves are replaced with uniformly sampled legal moves during evaluation. We focus on GPT-3.5 Turbo Instruct following prior observations that newer GPT models frequently generate illegal chess moves despite stronger general capabilities~\citep{carlini2023PlayingChess}. Additional prompting details and benchmark variants are provided in Appendices~\ref{app:experimental_details} and~\ref{app:prompt_templates}.

\section{Results}
\label{sec:results}

Tables~\ref{tab:main_results_top_move_accuracy} and~\ref{tab:main_results_expected_accuracy} summarize the main benchmark results. Overall, the models exhibit a clear trade-off: language models perform best on opening reconstruction, metadata-conditioned models achieve the strongest conventional human move prediction performance, and UniMaia performs best on prompt-conditioned instruction-following benchmarks.

\begin{table*}[h!]
\centering
\small
\begin{tabular}{l c c | c c | c c c c}
\toprule
{Model} & {LOB-P} & {LOB-C} & {LIF} & {LGB} & {ABB} & {M1-S} & {M1} & {M2R} \\
\midrule
Lc0-CF (BT4) & 0.5025 & 0.5039 & 0.3924 & 0.4770 & 0.4133 & 0.4255 & 0.4305 & 0.4079 \\
Maia-1\textsuperscript{*} & 0.5582 & 0.5602 & 0.4797 & 0.4857 & 0.5043 & 0.5072 & 0.5119 & 0.5094 \\
Maia-2 (Blitz) & 0.4058 & 0.4040 & 0.4452 & 0.4961 & 0.5230 & 0.5246 & \underline{0.5300} & 0.5284 \\
Allie-Policy & 0.6150 & 0.6164 & 0.5390 & 0.5572 & \bftab{0.5643} & \bftab{0.5733} & \bftab{0.5717} & \bftab{0.5598} \\
\midrule
ChessGPT-base\textsuperscript{\dag} & \bftab{0.7998} & \bftab{0.8955} & 0.4464 & 0.4788 & 0.4336 & 0.4375 & -- & 0.4305 \\
GPT-3.5\textsuperscript{\S} & 0.5748 & 0.6431 & 0.5312 & 0.5514 & 0.5498 & \underline{0.5667} & -- & 0.5489 \\
\midrule
UniMaia & \underline{0.7094} & \underline{0.8162} & \bftab{0.5587} & \bftab{0.6158} & \underline{0.5548} & \underline{0.5667} & -- & \underline{0.5536} \\
UniMaia-Aux & 0.7071 & 0.7801 & \underline{0.5481} & \underline{0.6086} & 0.5471 & 0.5624 & -- & 0.5493 \\
\bottomrule
\end{tabular}
\caption{Main benchmark comparison measured by Acc@1. Best values are in bold and second-best values are underlined. \textsuperscript{*}Evaluated with the model with the closest Elo to the active player. \textsuperscript{\dag}Opening benchmarks evaluated with the prompt and PGN. Remaining benchmarks evaluated with the Elo and time control PGN header. \textsuperscript{\S}Opening benchmarks evaluated with the GM PGN header. Remaining benchmarks evaluated with the Elo and time control PGN header. Further details in Appendix~\ref{app:baseline_details}.}
\label{tab:main_results_top_move_accuracy}
\end{table*}

\begin{table*}[h!]
\centering
\small
\begin{tabular}{l c c | c c | c c c c}
\toprule
{Model} & {LOB-P} & {LOB-C} & {LIF} & {LGB} & {ABB} & {M1-S} & {M1} & {M2R} \\
\midrule
Lc0-CF (BT4) & 0.3657 & 0.3655 & 0.2772 & 0.3436 & 0.3063 & 0.3175 & 0.3198 & 0.3036 \\
Maia-1\textsuperscript{*} & 0.4130 & 0.4138 & 0.3653 & 0.3651 & 0.3816 & 0.3840 & 0.3862 & 0.3827 \\
Maia-2 (Blitz) & 0.2990 & 0.2971 & 0.3423 & 0.3860 & 0.4102 & 0.4144 & \underline{0.4166} & 0.4137 \\
Allie-Policy & 0.4905 & 0.4906 & 0.4104 & 0.4480 & \bftab{0.4409} & \underline{0.4476} & \bftab{0.4475} & \underline{0.4354} \\
\midrule
UniMaia & \underline{0.5899} & \bftab{0.7140} & \underline{0.4297} & \underline{0.4834} & 0.4224 & 0.4317 & -- & 0.4182 \\
UniMaia-Aux & \bftab{0.5973} & \underline{0.6823} & \bftab{0.4331} & \bftab{0.4891} & \underline{0.4382} & \bftab{0.4488} & -- & \bftab{0.4427} \\
\bottomrule
\end{tabular}
\caption{Main benchmark comparison measured by \(\mathbb{E}[\mathrm{Acc}]\). Best values are in bold and second-best values are underlined. \textsuperscript{*}Evaluated with the model with the closest Elo to the active player.}
\label{tab:main_results_expected_accuracy}
\end{table*}

UniMaia achieves the strongest Acc@1 performance on the instruction-following benchmarks LIF and LGB while remaining competitive with metadata-conditioned approaches on conventional human move prediction tasks. In contrast, ChessGPT-base performs best on the opening-focused LOB benchmarks but degrades substantially on broader gameplay benchmarks, suggesting weaker generalization beyond opening continuation prediction.

Allie achieves the strongest overall metadata-conditioned performance on ABB and M1-S, whereas UniMaia provides substantially stronger prompt-conditioned controllability with only modest reductions on metadata-conditioned tasks. UniMaia also remains competitive with GPT-3.5 and Allie despite being optimized primarily for natural-language policy modulation.

Table~\ref{tab:main_results_expected_accuracy} reports expected accuracy results. UniMaia and UniMaia-Aux achieve the strongest overall performance across most settings, with UniMaia-Aux often outperforming UniMaia on \(\mathbb{E}[\mathrm{Acc}]\) despite slightly lower Acc@1. This suggests that auxiliary temporal conditioning improves policy calibration even when temporal metadata is unavailable at evaluation time.

Unlike autoregressive language models, UniMaia inherits the legality constraints and inductive biases of the pretrained chess policy, substantially reducing illegal move generation and improving gameplay stability.

\subsection{Auxiliary Temporal Metadata}
\label{sec:results_auxiliary_targets}

\begin{table}[h!]
\centering
\small
\makebox[\linewidth][c]{%
\begin{tabular}{llccccc}
\toprule
\multirow{2}{*}{Benchmark} & \multirow{2}{*}{Model} & \multirow{2}{*}{Acc@1} & \multirow{2}{*}{\(\mathbb{E}[\mathrm{Acc}]\)} & \multicolumn{1}{c}{\(\mathrm{F1}_{w}\)} \\
 & & & & {Resigned} \\
\midrule
\multirow{2}{*}{LIF-Aux} & Allie-Policy & \underline{0.5390} & \underline{0.4104} & \underline{0.6409} \\
 & UniMaia-Aux & \bftab{0.5549} & \bftab{0.4318} & \bftab{0.7279} \\
 \midrule
\multirow{2}{*}{LGB-Aux} & Allie-Policy & \underline{0.5572} & \underline{0.4480} & \underline{0.0933} \\
 & UniMaia-Aux & \bftab{0.5872} & \bftab{0.4637} & \bftab{0.6643} \\
\bottomrule
\end{tabular}%
}
\caption[{Auxiliary targets results.}]{Auxiliary targets results. Best values are in bold and second-best values are underlined.}
\label{tab:auxiliary_targets_model_results_abridged}
\end{table}

Table~\ref{tab:auxiliary_targets_model_results_abridged} reports results on the temporally augmented LIF-Aux and LGB-Aux benchmarks, which append temporal metadata such as time controls, remaining clock information, and previous move delay to the prompt.

UniMaia-Aux consistently outperforms Allie-Policy on both benchmarks in Acc@1 and expected accuracy while additionally supporting auxiliary behavioral prediction tasks. The largest gains occur for resignation prediction, suggesting improved modeling of human gameplay behavior beyond next-move prediction.

\subsection{Training Dynamics}
\label{sec:training_dynamics}

\begin{figure}[t]
    \centering
    \includegraphics[width=\linewidth]{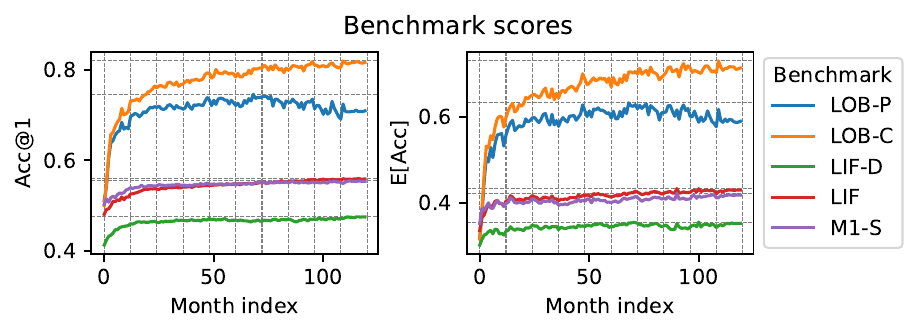}
    \caption{Benchmark performance across monthly checkpoints during training. Most benchmarks continue improving throughout training, although gains diminish over time. Vertical lines indicate year boundaries.}
    \label{fig:full_training_run_benchmark_scores}
\end{figure}

Figure~\ref{fig:full_training_run_benchmark_scores} shows benchmark performance across monthly checkpoints. Most benchmarks improve steadily throughout training, although gains diminish over time.

Interestingly, LOB-C continues improving, whereas LOB-P plateaus and slightly declines. Since LOB-C includes canonical opening continuations while LOB-P provides only partial opening information, this suggests that UniMaia increasingly benefits from stronger prompt constraints as training progresses.

These improvements come despite UniMaia being trained on a substantially smaller dataset than competing approaches: approximately 67M plies, compared to 9.15B for Maia-2 and over 262B tokens for Allie. This suggests that prompt-conditioned modulation of a pretrained backbone can be highly data-efficient, though the LLM backbone imposes higher computational costs due to memory and batch size constraints.

\section{Analysis}
\label{sec:analysis}

We analyze how prompt-conditioned modulation affects UniMaia's learned policy, focusing on policy continuity across Elo. Although UniMaia uses discrete Elo tokens, the induced policies vary continuously across rating levels, producing smooth behavioral changes under varying conditioning signals.

\begin{figure}[t]
    \centering
    \includegraphics[width=\linewidth]{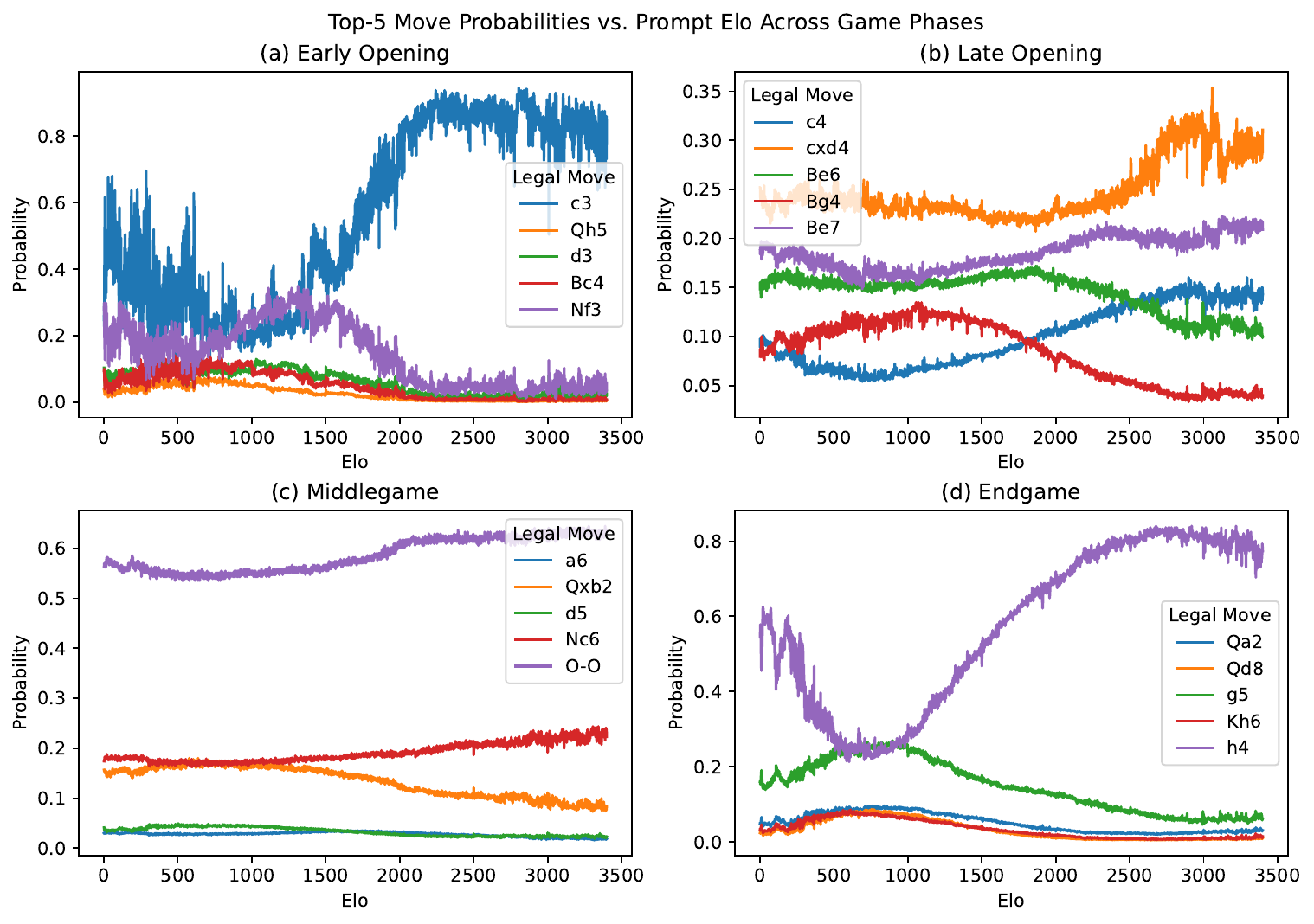}
    \caption{Top-5 move probabilities predicted by UniMaia as a function of Elo across representative game stages.}
    \label{fig:elo_plot_various_stages}
\end{figure}

Figure~\ref{fig:elo_plot_various_stages} shows top-5 move probabilities as a function of Elo for representative opening, middlegame, and endgame positions. Across positions, move distributions exhibit clear global trends as Elo increases, suggesting that UniMaia learns a continuous notion of player strength rather than isolated rating-specific behaviors. Local fluctuations remain, particularly in openings with many plausible continuations.

For Elo values far below the training distribution, the policy partially reverts toward stronger play, likely reflecting the influence of the frozen pretrained Lc0 backbone under limited low-Elo supervision. Ratings below 600 are absent during training due to a hard-coded minimum Elo threshold on Lichess~\cite{githubLowerMinimum,githubLowerMinRating}.

\begin{figure}[t]
    \centering
    \includegraphics[width=0.8\linewidth]{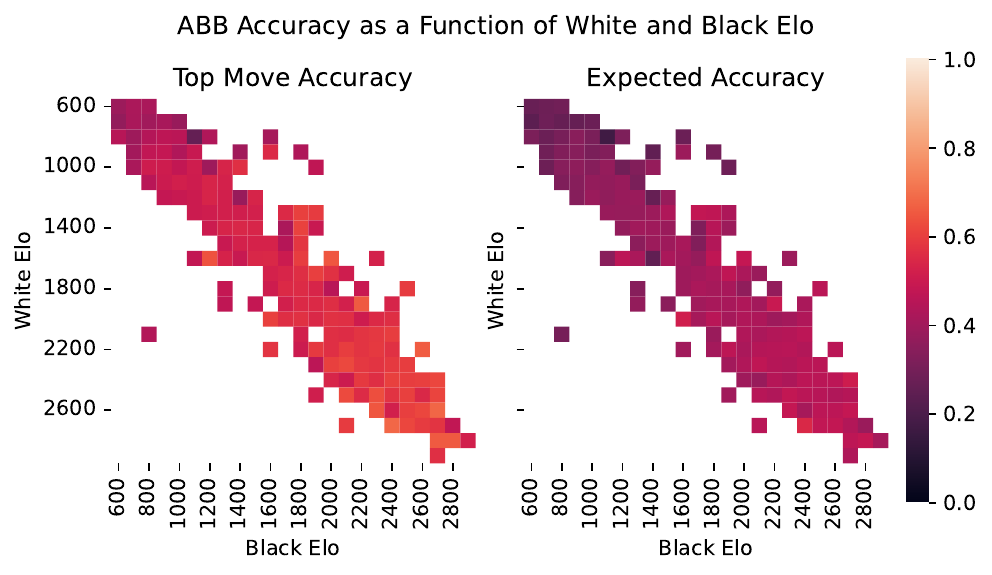}
    \caption{ABB accuracy as a function of White and Black Elo. Left: top-move accuracy. Right: expected accuracy.}
    \label{fig:abb_reduced_size_accuracy_heatmap_white_and_black_elo}
\end{figure}

We further analyze rating-dependent behavior using the metadata-conditioned ABB benchmark. Figure~\ref{fig:abb_reduced_size_accuracy_heatmap_white_and_black_elo} shows that both top-move accuracy and expected accuracy generally increase with the Elo of both players, consistent with higher-rated games being more predictable and lower-rated games exhibiting greater stochasticity.

Compared to prompt-conditioned benchmarks such as LIF, the ABB heatmap is substantially smoother across rating combinations. We attribute this to ABB using a fixed prompt template, whereas LIF samples diverse templates with varying metadata fields, introducing additional variability.

Overall, UniMaia exhibits coherent rating-dependent behavior while remaining sensitive to contextual factors such as opening structure and prompt formulation.

\section{Conclusion}
\label{sec:conclusion}

UniMaia demonstrates that pretrained domain-specific policy networks can be steered through natural language while preserving the inductive biases of specialized models. By combining a frozen Lc0-based chess policy with lightweight prompt-conditioned modulation, UniMaia achieves strong controllability over behaviors such as opening selection and player strength while remaining competitive on human move prediction benchmarks. Our results suggest that controllable policy modulation in structured domains depends more on conditioning design, including data composition and integration with pretrained expert policies, than on raw model scale. More broadly, this work highlights lightweight language-conditioned adaptation as a practical alternative to fully end-to-end multimodal training for structured decision-making systems, providing a general framework for controllable policies built on pretrained expert models.

\section*{Limitations}
\label{sec:limitations}

Despite strong performance across prompt-conditioned and metadata-conditioned benchmarks, UniMaia has several limitations.

First, the model remains sensitive to prompt phrasing and tokenization. Small lexical or formatting changes can noticeably alter the policy, especially for uncommon or out-of-distribution prompts. Cross-attention analysis also reveals widespread attention sink behavior, where many layers disproportionately attend to the first prompt token (Appendix~\ref{app:cross_attention_sinks}), suggesting both prompt-distribution and architectural limitations in robustness.

Second, the prompt data is synthetically generated using LLM-assisted templates with semi-automated validation and manual review. While scalable, this process may introduce artifacts and does not fully capture the variability or ambiguity of real human instructions. Some metadata dimensions are also underrepresented or omitted, such as time control information in the primary instruction-template distribution. This limitation affects UniMaia but not UniMaia-Aux, whose prompts are augmented with detailed temporal metadata. Maintaining correctness under iterative template modifications is also challenging: independently generated metadata snippets can introduce grammatical inconsistencies, duplicated phrases, or formatting errors across many prompt instantiations, and exhaustive validation became increasingly difficult as the template space expanded.

Third, the opening benchmarks rely on curated opening names and canonical continuations. While this improves reproducibility, it may partially reward memorization of opening terminology rather than deeper semantic understanding. Benchmark performance may also not fully reflect perceived human likeness or stylistic coherence during gameplay.

The evaluation scope is similarly limited. Experiments rely entirely on offline automated benchmarks rather than human studies or live gameplay, leaving open questions about perceived human likeness, long-term controllability, and robustness to paraphrased or adversarial prompts.

The training data is limited to Lichess games, introducing platform-specific biases related to player demographics, openings, time controls, and online play behavior. Generalization to over-the-board or correspondence chess may therefore be imperfect.

Architecturally, UniMaia prioritizes modularity and preservation of the pretrained Lc0 policy through a layer-coupled ControlNet design. However, later ablations suggest that deeper fusion and full fine-tuning can achieve stronger raw performance, indicating a trade-off between controllability, modularity, and peak accuracy. UniMaia-Aux also increases optimization complexity while yielding only modest gains in some settings.

Many design choices, including template construction and auxiliary objectives, were tuned specifically for chess. Although the overall approach could potentially generalize to other domains by pairing pretrained expert policies with domain-specific instruction data, such adaptation would require substantial changes to the data collection and conditioning pipeline.

Finally, UniMaia's frozen 3B-parameter LLM backbone introduces substantial computational overhead. The resulting batch size of 64, compared to 8{,}192 for Maia-2, requires many more gradient steps per training position. Full training took 31.6 days on two RTX Pro 6000 GPUs, exceeding the compute budgets of Maia-2 (13 days on two 80-GB A100s) and Allie (14 days on eight A6000s), even after accounting for per-GPU throughput differences.

However, smaller-scale experiments were considerably more accessible: the 2013-only ablation runs, which used the same architecture and parameter count, completed in roughly 7 hours on a single consumer-grade RTX 5090 with a smaller batch size. Similarly, fine-tuning UniMaia-Aux on 2018 data required approximately 1 day on two RTX Pro 6000 GPUs. These results suggest that meaningful experimentation, ablation studies, and downstream adaptation remain feasible on modern hardware, despite the high cost of full-scale training.

Nevertheless, the overall preprocessing pipeline and large-scale multi-GPU training may still limit reproducibility and accessibility. Future work could reduce this overhead through smaller language model backbones, more efficient conditioning mechanisms, or more aggressive quantization.

\section*{Acknowledgments}

Generative AI was used to proofread, reorganize, and improve the conciseness of the text. For code, autocomplete suggestions were used, and some brief example snippets were modified for our own use.


\bibliography{custom}

\appendix

\input{appendices/appendix-lc0}
\input{appendices/appendix-rating-system}
\input{appendices/appendix-dataset-construction}
\input{appendices/appendix-termination-statistics}
\input{appendices/appendix-ablation-study}
\input{appendices/appendix-full-training-configuration}
\input{appendices/appendix-training-dynamics}
\input{appendices/appendix-experimental-setup}

\input{appendices/appendix-prompt-templates}
\input{appendices/appendix-detailed-evaluation-results}
\input{appendices/appendix-extended-analysis}
\input{appendices/appendix-continuity-evaluation-details}
\input{appendices/appendix-performance-elo-range}
\input{appendices/appendix-future-work}

\end{document}

%% file: appendices/appendix-lc0.tex
\appendix
\section{Leela Chess Zero Encoding}
\label{app:lc0_encoding}

UniMaia builds upon pretrained Leela Chess Zero (Lc0) ChessFormer policy networks~\cite{monroe2024mastering,monroe2026chessformer}. Because UniMaia conditions a frozen Lc0 policy through a ControlNet-style conditioning branch, the underlying Lc0 input and output representations remain unchanged. We briefly summarize the encoding scheme used by Lc0 below.

\subsection{Input Encoding}
\label{app:lc0_input_encoding}

Lc0 represents each position using 112 spatial feature planes defined over the \(8 \times 8\) chess board~\cite{githubLc0srcneuralencoderccMaster}. The representation encodes the current board state together with a fixed history of the previous seven positions, following the AlphaZero-style formulation~\cite{silver2017alphazero}.

The first \(13 \times 8 = 104\) planes encode the 8 most recent board positions, ordered from most recent to oldest. For each position, the first six binary planes encode the locations of the white player's pawns, knights, bishops, rooks, queens, and king, respectively, followed by six corresponding planes for the black player's pieces in the same order. A thirteenth plane indicates whether the position has previously occurred in the game, enabling detection of repetition states. Missing history positions are padded with zeros.

The remaining 8 planes encode auxiliary game-state information using spatially constant values across all board squares. In order, these planes encode white queenside castling rights, white kingside castling rights, black queenside castling rights, black kingside castling rights, whether it is black's turn to move, the halfmove clock normalized by 99, a deprecated all-zero plane, and an all-ones plane used to help identify board boundaries.

As in AlphaZero, board states are always encoded from the perspective of the current player, so the side to move is consistently represented as playing from the bottom of the board~\cite{silver2017alphazero}. Lc0 additionally uses a simplified repetition encoding compared to AlphaZero, employing a single repetition plane per historical position rather than two separate repetition-count planes. Although this encoding does not distinguish higher-order repetitions such as fivefold repetition, such events are extremely rare in practice (see Appendix~\ref{app:termination_statistics}).

\subsection{Output Encoding}
\label{app:lc0_output_encoding}

Lc0 predicts move probabilities over a fixed output space of 1{,}858 legal UCI move tokens~\cite{githubLczerotrainingtflc0_az_policy_mappyMaster}. The encoding covers all legal moves from the perspective of the current player, including pawn promotions represented as distinct move types. Knight promotions are handled implicitly by treating ordinary pawn moves from the seventh to eighth rank as knight promotions when the moved piece is a pawn.

In addition to the policy head, modern Lc0 models predict a three-dimensional value distribution corresponding to win, draw, and loss probabilities, rather than the scalar value prediction used in AlphaZero~\cite{silver2017alphazero}. During Monte Carlo Tree Search, the scalar evaluation is computed as \(\mathbb{P}(\mathrm{win}) - \mathbb{P}(\mathrm{loss})\).

Lc0 models also include a moves-left head that predicts the expected number of remaining plies in the game~\cite{githubMovesLeft}. This auxiliary prediction helps reduce degenerate MCTS behavior in decisively winning or losing positions, where many continuations receive similar value estimates, by encouraging shorter wins and longer losses and reducing move shuffling.

%% file: appendices/appendix-rating-system.tex
\section{Lichess Rating System and Rating Ranges}
\label{app:rating_system}

Lichess uses the Glicko-2 rating system~\cite{glickman2012example,lichessChessRating}, an extension of the Elo rating system~\cite{elo1961uscfratingsystem} that additionally models rating uncertainty and volatility. In Glicko-2, each player is associated with a rating, a rating deviation measuring uncertainty, and a volatility parameter controlling expected rating fluctuation over time. Rating deviations decrease as more games are played and increase during inactivity.

Although Lichess internally uses Glicko-2, player ratings are still recorded in the public PGN archives using the conventional Elo terminology~\cite{lichess-open-database}. We therefore use the term ``Elo'' throughout the paper for consistency with prior chess literature and existing datasets.

Lichess maintains separate ratings for each time control category, including Ultrabullet, Bullet, Blitz, Rapid, Classical, and Correspondence chess~\cite{lichess2024FrequentlyAsked}. New accounts begin at a rating of 1500~\cite{lichessChessRating}. Historically, the minimum possible rating was 600, but this floor was lowered to 400 in March 2023~\cite{githubLowerMinRating}. The theoretical maximum rating is 4000, though observed ratings are substantially lower in practice.

As of early 2026, Lichess ratings span approximately 400--3400 depending on the time control and player population. The rating distribution is highly concentrated in the intermediate range, with median rapid ratings near 1400 and most active players falling roughly between 900 and 1900~\cite{lichessChessPlayers,lichessWeeklyRapid}. Extremely high ratings above 2500 are comparatively rare and correspond primarily to titled or professional-level players.

These rating ranges are important for interpreting both the training distribution and the controllability behavior of UniMaia. Because the underlying game data is heavily concentrated in intermediate rating ranges, the model receives substantially more supervision signal for common amateur and club-level play than for extreme low- or high-rating regimes.

%% file: appendices/appendix-dataset-construction.tex
\section{Dataset Construction Details}
\label{app:dataset_construction}

This appendix provides implementation details for constructing \textsc{LichessGames}, prompt templates, and opening normalization, complementing the high-level overview in Section~\ref{sec:dataset}.

\begin{figure*}[h!]
    \centering
    \includegraphics[width=0.9\textwidth]{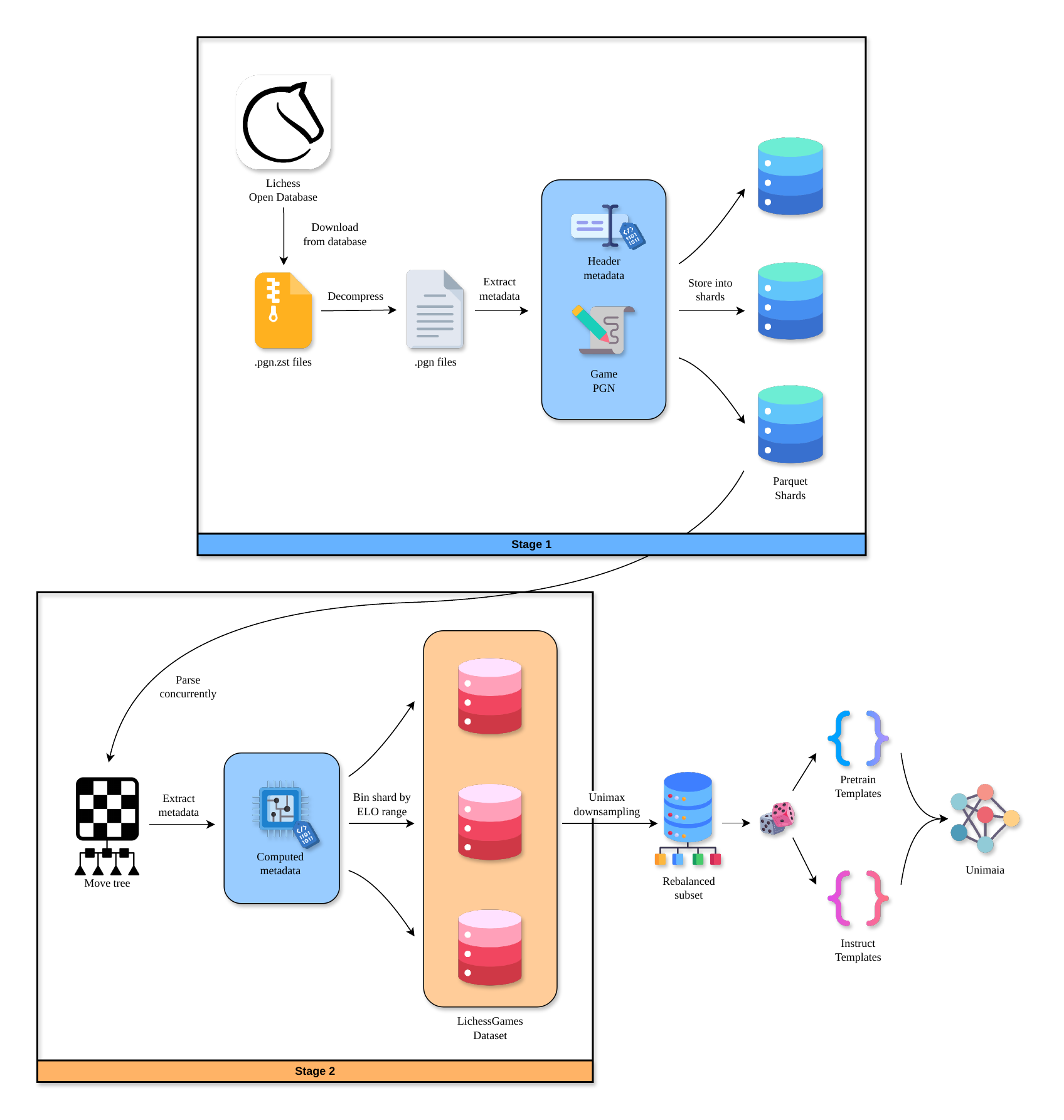}
    \caption{Overview of the data processing pipeline. Stage 1 performs lightweight string-based processing of PGNs into Parquet metadata shards, while Stage 2 parses games into move trees and extracts additional game-level features. During training, games are formatted using randomly sampled prompt templates from the pretraining and instruction-template families.}
    \label{fig:data_processing}
\end{figure*}

\subsection{Data Sources}
\label{app:data_sources}

Training data consists of publicly available Lichess PGN archives from 2013--2023. Additional auxiliary metadata sources are stored separately and joined dynamically when constructing training examples or prompt-conditioned benchmarks.\footnote{The version of the openings database used in this work contains 3,464 openings, although newer releases contain additional entries.} After preprocessing, the resulting \textsc{LichessGames} dataset contains approximately 5.2B games and 265.2B positions.

\begin{table}[t]
\centering
\small
\begin{tabular}{ll}
\toprule
\textbf{Statistic} & \textbf{Value} \\
\midrule
Total games          & \num{5.185e9} \\
Total positions      & \num{2.65197e11} \\
Years covered        & 2013--2023 \\
Source               & Lichess Open Database \\
License              & CC0 \\
ELO range            & 600--\(\approx 3400\) \\
Templates (pretrain) & 160 \\
Templates (instruct) & 2048 \\
\bottomrule
\end{tabular}
\caption{Summary statistics of the \textsc{LichessGames} dataset.}
\label{tab:lichessgames_summary}
\end{table}

\begin{table}[t]
\centering
\small
\begin{tabular}{p{2cm} p{2cm} p{0.8cm} p{1cm}}
\toprule
\textbf{Dataset} & \textbf{Purpose} & \textbf{Entries} & \textbf{License} \\
\midrule
Historical Peak Elo and Identity Data for Notable Lichess Users & Player name augmentation & 5,793 & CC0 \\
Lichess Chess Openings Database & Opening normalization and canonical move sequences & 3,464 & CC0 \\
\bottomrule
\end{tabular}
\caption{Auxiliary metadata sources used during training-example and benchmark construction. These sources are stored separately from \textsc{LichessGames} and joined dynamically as needed.}
\label{tab:auxiliary_metadata_sources}
\end{table}

\subsection{Creating \textsc{LichessGames}}
\label{app:creating_lichess_games}

As illustrated in Figure~\ref{fig:data_processing}, dataset construction proceeds in two stages.

\paragraph{Stage 1: Metadata extraction.}
The first stage performs lightweight string-only processing of PGNs without parsing move trees. Metadata headers and raw PGN strings are extracted and stored in Parquet shards using \texttt{python-chess}~\cite{github-niklasfpythonchess}, specifically with the cheaper \texttt{chess.pgn.read\_header} routine.

Games from the same month are stored sequentially in a single streaming pipeline with batched Parquet writes. Each shard stores at most \(4096 \times 1024\) games to avoid memory issues during later processing.

To support concurrent parsing in Stage 2, byte offsets are tracked for each PGN and the file pointer is rewound after reading headers.
Because UTF-8 characters may occupy multiple bytes, offsets must be adjusted accordingly:

\begin{lstlisting}[caption={Handling UTF-8 byte offsets when recovering PGN strings.}, label={lst:pgn-unicode-offset}]
diff = len(pgn.encode("utf-8")) - len(pgn)
f.seek(prev_offset)
pgn = f.read(offset - prev_offset - diff)
\end{lstlisting}

Table~\ref{tab:lichess_games_column_description_stage_1} summarizes the metadata fields extracted during Stage 1.

\begin{table}[h!]
    \centering
    \small
    \begin{tabularx}{\linewidth}{lX}
        \toprule
        \textbf{Field} & \textbf{Description} \\
        \midrule
        Event & The event name \\
        TimeControlName & The time control, if provided \\
        UTCDate & The date that the game was played \\
        Year & The year that the game was played \\
        Month & The month that the game was played \\
        Day & The day that the game was played \\
        TcDelay & Base time in seconds, if provided \\
        TcIncrement & Increment in seconds, if provided \\
        WhiteElo & White player Elo \\
        BlackElo & Black player Elo \\
        WhiteRatingDiff & White player rating change \\
        BlackRatingDiff & Black player rating change \\
        Round & Tournament round, if applicable \\
        Result & Game result \\
        Variant & Chess variant \\
        ECO & ECO opening code \\
        Opening & Opening name \\
        Termination & Coarse game termination info \\
        Annotator & PGN annotator \\
        IsFischerRandom & Whether the game is Fischer random chess \\
        EvalDepth & Stockfish evaluation depth, if available \\
        OriginalPGN & Original PGN string \\
        \bottomrule
    \end{tabularx}
    \caption{Metadata fields extracted during Stage 1.}
    \label{tab:lichess_games_column_description_stage_1}
\end{table}

\paragraph{Stage 2: Move-tree parsing and partitioning.}
The second stage fully parses PGNs into move trees using \texttt{chess.pgn.read\_game}, which is substantially more CPU-intensive. 
Because PGNs are already separated into arrays during Stage 1, this stage can be parallelized efficiently across workers on a SLURM cluster.

Additional features extracted during this stage are summarized in
Table~\ref{tab:lichess_games_column_description_stage_2}.

\begin{table}[h!]
    \centering
    \small
    \begin{tabularx}{\linewidth}{lX}
        \toprule
        \textbf{Field} & \textbf{Description} \\
        \midrule
        Length & Game length in halfmoves \\
        TerminationOutcome & Outcome under standard chess rules \\
        TerminationWinner & Winning player \\
        LastComment & Final PGN comment \\
        TerminationReason & Combined termination information \\
        MeanElo & Mean player Elo \\
        DiffElo & Absolute Elo difference \\
        \bottomrule
    \end{tabularx}
    \caption{Additional features extracted during Stage 2.}
    \label{tab:lichess_games_column_description_stage_2}
\end{table}

Games are partitioned into 100-point Elo bins based on mean player Elo and Elo difference.
These partitions are later combined with Unimax sampling~\cite{chung2023unimax} during training.

\subsection{Prompt Template Construction}
\label{app:prompt_template_construction}

This section describes implementation details of the prompt-template
generation pipeline summarized in
Section~\ref{sec:prompt_template_construction}.

Prompt templates are initially generated using \texttt{gpt-4-turbo-2024-04-09}~\cite{achiam2023gpt4}, then converted into parameterized Jinja2 templates by replacing metadata spans with programmatically generated snippets.

Template construction follows an iterative human-in-the-loop workflow:
\begin{enumerate}
    \item Generate natural-language prompts conditioned on game metadata.
    \item Convert metadata spans into parameterized Jinja2 fields.
    \item Instantiate templates across many games.
    \item Manually review outputs and refine formatting rules.
\end{enumerate}

A major challenge is maintaining correctness under iterative modifications. 
Because templates combine independently generated metadata snippets, small changes can introduce grammatical or formatting errors across many instantiations.\footnote{Substantial effort was invested in designing metadata formatting rules that produce well-formed text across edge cases, including correct pluralization, handling zero versus nonzero increments, and decomposing time controls into hours, minutes, and seconds. However, as the template space expanded, the combinatorial growth in template instantiations made exhaustive validation increasingly difficult.}

Consequently, some prompts contain minor artifacts such as duplicated words or awkward phrasing.
These artifacts are retained in both training and evaluation for reproducibility and do not appear to materially affect model behavior.

\subsubsection{\textsc{LichessTemplates-Pretrain}}
\label{app:lichess_templates_pretrain}

\textsc{LichessTemplates-Pretrain} contains 160 templates generated from combinations of:
\begin{itemize}
    \item 5 seed games,
    \item 4 time controls (bullet, blitz, rapid, classical),
    \item common versus uncommon openings,
    \item concise versus regular verbosity.
\end{itemize}

Two prompts are generated per configuration, yielding:
\[
2 \times 5 \times 4 \times 2 \times 2 = 160
\]
templates.

\subsubsection{\textsc{LichessTemplates-Instruct}}
\label{app:lichess_templates_instruct}

\textsc{LichessTemplates-Instruct} contains 2,048 templates varying:
\begin{itemize}
    \item player titles,
    \item Elo ratings,
    \item textual skill descriptions,
    \item usernames,
    \item real names,
    \item opening names,
    \item opening move sequences,
    \item white versus black perspective,
    \item request versus descriptive framing.
\end{itemize}

Combined with common versus uncommon openings and prompt paraphrases, this yields:
\[
2^{9} \times 2 \times 2 = 2048
\]
templates.

\subsection{Enriching Player Metadata with Real Names}
\label{app:player_metadata_real_names}

The player identity mapping data was obtained from a publicly distributed, anonymized file (\texttt{famous\_lichess.tsv}) originally shared on an online chess community forum. Because the original hosting platform and author are no longer active or identifiable, the dataset has been independently preserved and re-hosted for reproducibility purposes~\cite{unknown_2026_20389754_famouslichess}.

\subsection{Opening Normalization and Canonical Move Sequences}
\label{app:opening_normalization}

Opening normalization uses ECO codes and opening names to align historical Lichess annotations with the current Lichess Chess Openings database~\cite{github-lichess-org-chessopenings}.

Unresolved openings are matched using \texttt{thefuzz}~\cite{github-Seatgeekthefuzz} and \texttt{rapidfuzz}~\cite{github-Rapidfuzz}, where token-sort ratio performed best empirically.

When automated matching failed, openings were resolved manually by:
\begin{itemize}
    \item inspecting repository history using \texttt{git log -S},
    \item searching for the opening sequence externally,
    \item verifying the resulting sequence using the Lichess analysis board.
\end{itemize}

At prompt-construction time, the historical opening name is preserved in the text prompt, while the normalized opening entry is used to recover the canonical move sequence.

Because multiple move sequences may correspond to the same opening name, we identify the longest common prefix consistent with the played PGN and use this prefix as the canonical opening continuation.

%% file: appendices/appendix-termination-statistics.tex
\section{Termination Statistics}
\label{app:termination_statistics}

This appendix reports the frequencies of termination reasons for representative dataset shards used in the analysis of repetition-related outcomes discussed in Appendix~\ref{app:lc0_encoding}. We include shards from March 2014, October 2022, and January 2023 to demonstrate that five-fold repetition remains consistently rare across time, including both early data and data collected after changes to Lichess draw-claim behavior. Specifically, Lichess introduced automatic draw claims for threefold repetition under low time conditions in April 2014 and later made automatic claiming the default in November 2022~\cite{githubAutoThreefold,githubChangeDefaultThreefoldToAlways}.

\begin{table}[h!]
\centering
\small
\begin{tabular}{lrr}
\toprule
Termination Reason & Counts & Percentage (\%) \\
\midrule
Resigned & \num{266449} & \num{37.425} \\
Time forfeit & \num{231410} & \num{32.503} \\
Checkmate & \num{193718} & \num{27.209} \\
Threefold repetition & \num{5615} & \num{0.789} \\
Insufficient material & \num{5098} & \num{0.716} \\
Stalemate & \num{4868} & \num{0.684} \\
Draw by agreement & \num{3966} & \num{0.557} \\
Fivefold repetition & \num{640} & \num{0.090} \\
Fifty moves & \num{196} & \num{0.028} \\
\midrule
Total & \num{711960} & \num{100.000} \\
\bottomrule
\end{tabular}
\caption[Termination reason frequencies for the only shard of March 2014 games in \textsc{LichessGames}.]{Termination reason frequencies for the only shard of March 2014 games in \textsc{LichessGames}, after applying the filtering criteria described in Appendix~\ref{app:ablation_baseline_configuration}.}
\label{tab:mar_2014_first_shard_termination_reason_frequencies}
\end{table}

\begin{table}[h!]
\centering
\small
\begin{tabular}{lrr}
\toprule
Termination Reason & Counts & Percentage (\%) \\
\midrule
Resigned & \num{1326763} & \num{34.911} \\
Time forfeit & \num{1288942} & \num{33.915} \\
Checkmate & \num{1061309} & \num{27.926} \\
Threefold repetition & \num{50813} & \num{1.337} \\
Stalemate & \num{31683} & \num{0.834} \\
Insufficient material & \num{26885} & \num{0.707} \\
Draw by agreement & \num{9675} & \num{0.255} \\
Fivefold repetition & \num{3576} & \num{0.094} \\
Fifty moves & \num{735} & \num{0.019} \\
Abandoned & \num{50} & \num{0.001} \\
Rules infraction & \num{27} & \num{0.001} \\
\midrule
Total & \num{3800458} & \num{100.000} \\
\bottomrule
\end{tabular}
\caption[Termination reason frequencies for the first shard of October 2022 games in \textsc{LichessGames}.]{Termination reason frequencies for the first shard of October 2022 games in \textsc{LichessGames}, after applying the filtering criteria described in Appendix~\ref{app:ablation_baseline_configuration}.}
\label{tab:oct_2022_first_shard_termination_reason_frequencies}
\end{table}

\begin{table}[h!]
\centering
\small
\begin{tabular}{lrr}
\toprule
Termination Reason & Counts & Percentage (\%) \\
\midrule
Resigned & \num{1321464} & \num{34.705} \\
Time forfeit & \num{1294684} & \num{34.002} \\
Checkmate & \num{1064219} & \num{27.949} \\
Threefold repetition & \num{59097} & \num{1.552} \\
Stalemate & \num{30763} & \num{0.808} \\
Insufficient material & \num{26698} & \num{0.701} \\
Draw by agreement & \num{9385} & \num{0.246} \\
Fivefold repetition & \num{731} & \num{0.019} \\
Fifty moves & \num{626} & \num{0.016} \\
Rules infraction & \num{34} & \num{0.001} \\
Abandoned & \num{2} & \num{0.000} \\
\midrule
Total & \num{3807703} & \num{100.000} \\
\bottomrule
\end{tabular}
\caption[Termination reason frequencies for the first shard of January 2023 games in \textsc{LichessGames}.]{Termination reason frequencies for the first shard of January 2023 games in \textsc{LichessGames}, after applying the filtering criteria described in Appendix~\ref{app:ablation_baseline_configuration}.}
\label{tab:jan_2023_first_shard_termination_reason_frequencies}
\end{table}

%% file: appendices/appendix-ablation-study.tex
\section{Additional Design Evolution Details}
\label{app:additional_ablation_details}

Table~\ref{tab:design_evolution_parameters} summarizes the parameter changes introduced by each sequential design modification from Section~\ref{sec:design_evolution}. Architectural scaling contributes the majority of the parameter increase.

\begin{table}[t]
\centering
\small
\begin{tabular}{lcc}
\toprule
\textbf{Change} & \makecell{\textbf{Trainable LLM} \\ \textbf{Parameters}} & \makecell{\textbf{Lc0} \\ \textbf{Parameters}} \\
\midrule
ControlNet + policy head & -- & +124.486M \\
NorMuon optimizer & -- & -- \\
LoRA vs DoRA & \(\alignedminus\)0.328M & -- \\
BT4 backbone & -- & +279.038M \\
Unimax sampling & -- & -- \\
+ LoRA rank 16 & +3.932M & -- \\
\midrule
Full configuration & \bftab{+3.604M} & \bftab{+403.524M} \\
\bottomrule
\end{tabular}
\caption{Parameter changes associated with each design modification. The BT4 backbone accounts for most of the parameter increase (+279M of +404M total), despite comparatively modest gains in development score.}
\label{tab:design_evolution_parameters}
\end{table}

\begin{figure}[t]
    \centering
    \includegraphics[width=\linewidth]{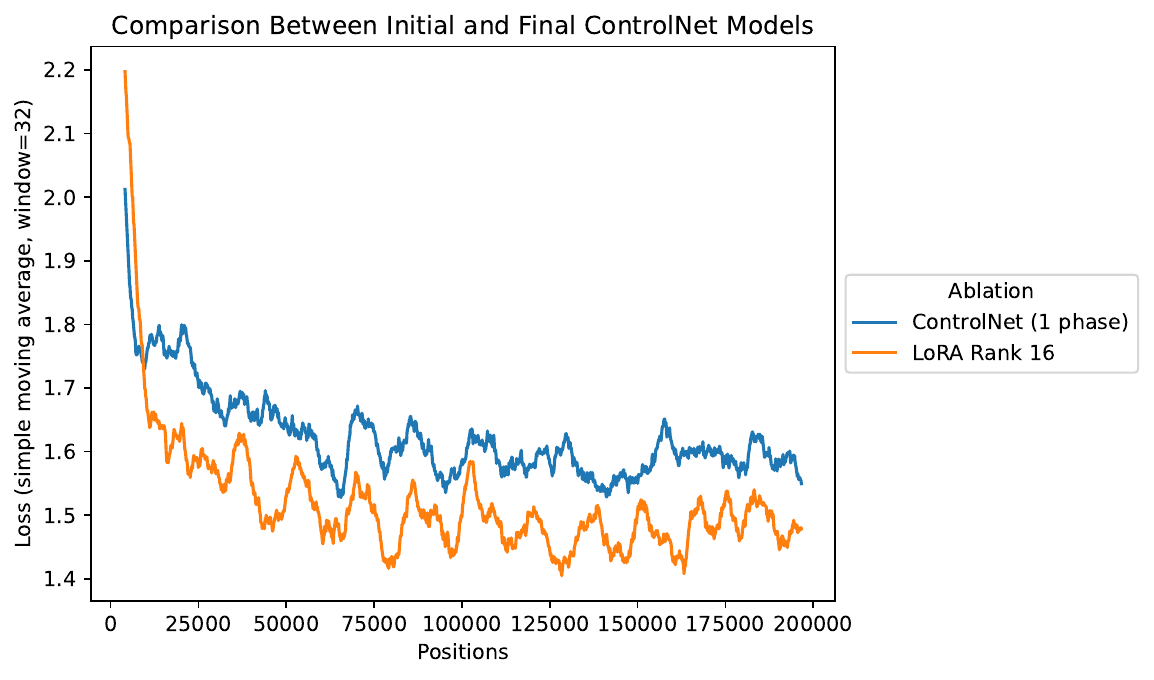}
    \caption{Comparison of training losses between the initial single-phase, layer-coupled ControlNet and the final model.}
    \label{fig:comparison_initial_final_controlnet_losses_appendix}
\end{figure}

Figure~\ref{fig:comparison_initial_final_controlnet_losses_appendix} compares the training losses of the initial and final configurations. Although most modifications add little or no parameter overhead, their combined effect substantially improves optimization and final loss.

Overall, most improvements are parameter-efficient, while later scaling decisions contribute the majority of the parameter growth for comparatively smaller gains.

\section{Ablation Study}
\label{app:ablation_study}

This appendix provides additional details on the ablation studies used during UniMaia development. The main paper summarizes the sequential design changes leading to the final model; here, we describe the shared experimental setup and detailed ablation results. Full-scale training dynamics are discussed separately in Appendix~\ref{app:training_dynamics}.

All ablations are trained on Lichess games from January--December 2013. This subset preserves realistic game distributions while enabling rapid iteration over architecture, optimization, parameter-efficient fine-tuning, backbone scaling, and data-sampling strategies. Unless otherwise specified, all ablations use the same preprocessing pipeline, prompt construction procedure, optimizer settings, and evaluation protocol.

\subsection{Training Data and Procedure}
\label{app:ablation_training_data_and_procedure}

Ablations follow the same monthly partitioning scheme as the full training run, restricted to the 2013 subset. Since globally shuffling the dataset is impractical, each monthly partition is divided into independently shuffled shards processed sequentially. Although this may introduce mild temporal drift, it enables efficient checkpointing and incremental training.

Within each shard, games are grouped by mean player Elo and absolute Elo difference. Unimax downsampling~\cite{chung2023unimax} is applied across Elo buckets to maintain coverage across skill levels while enforcing a fixed training budget. Additional data-sampling ablations (Appendix~\ref{app:data_sampling}) are applied within each Elo bucket to isolate the effects of balancing by opening, time control, or game length.

Each ablation trains on \(2^{14}\) plies per shard, totaling \num{196608} training plies across 2013. Since the subset contains one shard per month, training proceeds sequentially from January through December. To reduce prompt-template overfitting, prompts are lowercased with probability \(0.1\) and stripped of diacritics with probability \(0.5\).

To ensure hardware stability during long training runs, we insert a cooldown period after processing each month to prevent GPU thermal throttling and system crashes.

\subsection{Optimization}
\label{app:ablation_optimization}

Unless otherwise specified, ablations use the same optimization hyperparameters as the corresponding compared configuration. We use weight decay \(0.01\) and \(\epsilon=\num{1e-8}\). Learning rates follow the development schedule: linear warmup to \(\num{5e-5}\), followed by staged monthly decay. Optimizer-specific changes, including the transition from AdamW~\cite{loshchilov2017decoupledadamw} to NorMuon, are evaluated separately in Appendix~\ref{app:optimizer_ablations}.

The ablations are designed for rapid iteration rather than full-scale training. In practice, the 2013-only runs complete in several hours on modern consumer GPUs such as the RTX 3090 or RTX 5090, depending on batch size and memory constraints.

\subsection{Evaluation Protocol}
\label{app:ablation_evaluation_protocol}

Ablations are compared using both training loss and downstream benchmark performance. Since per-step loss curves are noisy, we summarize the training loss using the mean loss over the December 2013 partition, denoted \(\mu_{\mathrm{loss},\,\mathrm{Dec13}}\).

For downstream evaluation, we use a compact diagnostic benchmark suite:

\begin{itemize}
    \item \textbf{LOB-P}: opening-following under partial opening information;
    \item \textbf{LIF-D}: diagnostic instruction-following over a restricted Elo range;
    \item \textbf{M1-S}: human move prediction on a stratified subset of the Maia-1 benchmark.
\end{itemize}

LIF-D is used only during development because it partially overlaps with the training distribution. Final evaluation instead uses the full Lichess Instruction-Following Benchmark described in Section~\ref{sec:experimental_setup}. As described in the main paper, we summarize ablation performance using the development score: the mean normalized accuracy across these diagnostic tasks.

\subsection{Initial Baseline Configuration}
\label{app:ablation_baseline_configuration}

The initial ablation baseline is an early-fusion model with a single cross-attention conditioning layer. It combines prompt embeddings from ChessGPT-base~\cite{feng2024chessgpt} with a frozen Lc0-CF policy backbone~\cite{monroe2024mastering,monroe2026chessformer} initialized from the \texttt{t3} checkpoint.

Unless otherwise stated, ablations exclude bot games, Lichess Master games\footnote{Lichess Master was a rare discretionary title formerly awarded by Lichess~\cite{lichessTitleLichess}; all other titles are officially awarded by FIDE.}, correspondence games, games missing Elo or time-control metadata, and games shorter than 32 plies unless they end in checkmate or forced draw. We retain the standard Lc0 input representation, including its repetition handling, since five-fold repetition is extremely rare in the dataset.

The following sections evaluate the main design choices explored during development. Each ablation isolates one factor where possible, while the sequential design table in the main paper should be interpreted as a development trace rather than a fully factorial study.

\subsection{Architecture Design}
\label{app:architecture_design}

We study how conditioning effectiveness depends on both the frequency of cross-modal interaction and the integration of prompt signals into the frozen chess policy. We compare early fusion, deep fusion, and ControlNet-style conditioning.

\subsubsection{Fusion-Based Conditioning}
\label{app:fusion_based_conditioning}

We first evaluate two standard fusion approaches that inject prompt information directly into the Lc0 encoder.

\paragraph{Single-step cross-attention (early fusion).}
Inspired by early-fusion multimodal models such as LLaVA~\cite{liu2023visualllava}, this variant performs a single cross-attention operation between the prompt embeddings and the shallow board representation before the Transformer stack. The fused representation is then processed by all encoder layers and output heads without further prompt interaction.

For training stability, we adopt a three-stage procedure:
\begin{enumerate}[label=(\roman*)]
    \item both the text encoder and Lc0 encoder are frozen,
    \item the text encoder is adapted using DoRA~\cite{liu2024dora},
    \item all non-policy components are frozen while the policy outputs are fine-tuned.
\end{enumerate}

Each stage is trained on disjoint month ranges (Jan--Mar, Apr--Jun, Jul--Dec).

\paragraph{Per-layer cross-attention (deep fusion).}
This variant inserts cross-attention blocks into every Transformer layer, enabling iterative prompt-conditioned refinement of the board representation. All cross-attention output projections are zero-initialized for stability, while the pretrained Lc0 parameters remain frozen.

\subsubsection{ControlNet-Based Conditioning}
\label{app:controlnet_conditioning}

We next evaluate ControlNet-style conditioning~\cite{zhang2023adding_controlnet}, which introduces a parallel trainable pathway alongside the frozen backbone.

\paragraph{Layer-coupled ControlNet.}
A trainable copy of the encoder is added in parallel to the frozen Lc0 encoder. After each layer, the ControlNet activations are projected and injected residually into the corresponding backbone layer. Cross-attention modules inserted between the self-attention and feed-forward blocks enable prompt-conditioned modulation of the hidden board representation.

Following the PixArt-\(\delta\) ControlNet design~\cite{chen2024pixartdelta_controlnet_transformer}, both cross-attention outputs and residual injections are zero-initialized for stability.

Unlike the original sequential ControlNet formulation, both the base layer and the ControlNet layer at depth \(n\) receive the base hidden state from depth \(n-1\). We refer to this as the \emph{layer-coupled} design.

We evaluate:
\begin{itemize}[noitemsep, topsep=-\parskip, after=\vspace{1ex}]
    \item a two-stage variant, where the text encoder is initially frozen and later adapted with DoRA, and
    \item a single-stage variant, where the text encoder is adapted throughout training.
\end{itemize}

\paragraph{Controllable policy head.}
To enable direct modulation of the output distribution, we additionally evaluate a controllable policy head (CPH). A pre-normalized cross-attention block is inserted between the cloned policy embedding and the cloned policy head. Given a policy embedding \(x\), cross-attention over the prompt embeddings produces an update \(y\), which is added residually before the cloned policy head:
\[
x \leftarrow x + y.
\]

This allows prompt information to directly influence the policy logits.

\paragraph{Gated attention.}
We further evaluate gated attention~\cite{qiu2025gated} within the controllable policy head to mitigate attention sinks observed during preliminary experiments.

\paragraph{Sequential ControlNet.}
Finally, we evaluate a sequential ControlNet variant closer to the original ControlNet connectivity pattern, where each ControlNet layer consumes the previous ControlNet hidden state rather than the base hidden state.

\subsubsection{Training Regime}
\label{app:architecture_training_regime}

Unless otherwise specified, all architecture ablations freeze the pretrained Lc0 backbone and train only the newly introduced conditioning components. This setting was initially chosen for stability, as early experiments with full joint optimization produced large loss spikes and unstable convergence.

Although later experiments partially mitigated these issues, we retain the frozen-backbone regime throughout the primary architecture study to isolate the effect of the conditioning mechanisms themselves.

\subsubsection{Results}
\label{app:architecture_results}

\begin{figure}[h!]
    \centering
    \includegraphics[width=\linewidth]{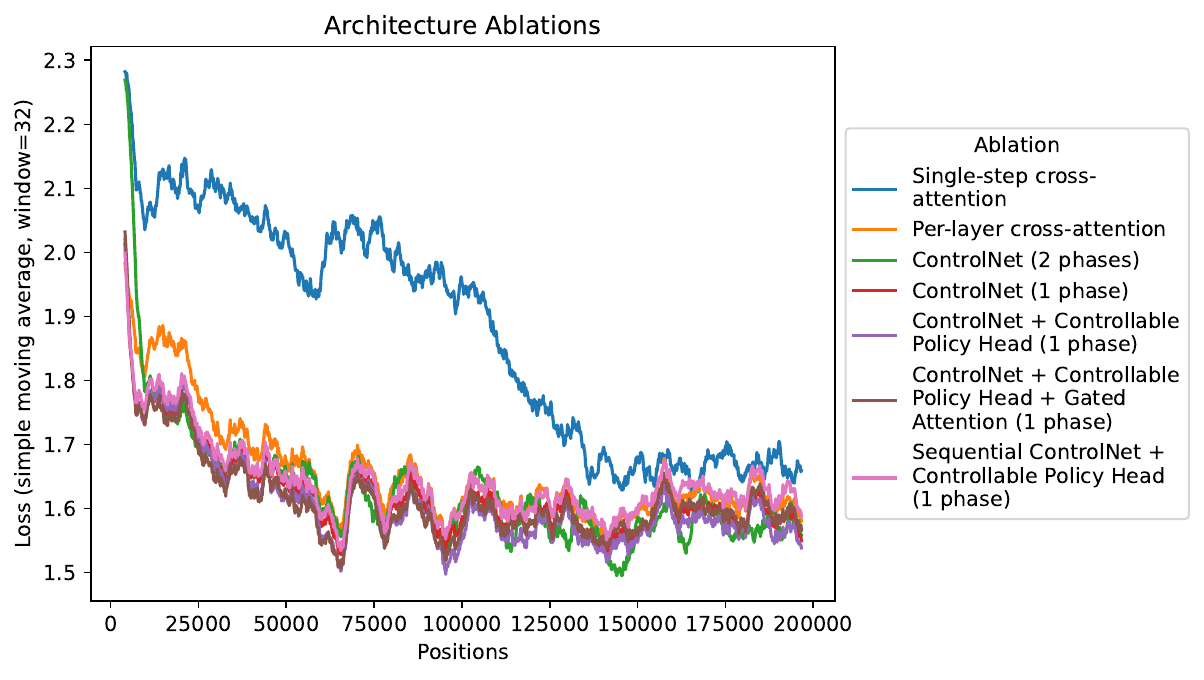}
    \caption[Training losses for the architecture ablations.]{Simple moving averages of the training losses for the architecture ablations. Due to the small batch size (\(8\)), the raw losses are highly volatile.}
    \label{fig:architecture_ablation_losses}
\end{figure}

Figure~\ref{fig:architecture_ablation_losses} shows the training losses for the architecture ablations. Fusion-based approaches consistently exhibit higher losses than ControlNet variants, particularly during early training.

\begin{table*}[h!]
\centering
\small
\begin{tabular}{l S[table-format=1.4] S[table-format=1.4] S[table-format=1.4] S[table-format=1.4] | S[table-format=1.4]}
\toprule
{Experiment} & {LOB-P} & {LIF-D} & {M1-S} & {Mean} & {\(\mu_{\mathrm{loss},\,\mathrm{Dec13}}\)} \\
\midrule
Single-step cross-attention (early fusion) & 0.4696 & 0.4176 & 0.4315 & 0.4396 & 1.6685 \\
Per-layer cross-attention (deep fusion) & \bftab{0.5745} & 0.4325 & 0.3379 & 0.4483 & 1.6092 \\
ControlNet (2 phases) & 0.5384 & 0.4332 & 0.5118 & 0.4944 & \underline{1.5753} \\
ControlNet & 0.5469 & \underline{0.4351} & 0.5122 & 0.4981 & 1.5865 \\
ControlNet + controllable policy head & \underline{0.5704} & \bftab{0.4358} & \underline{0.5179} & \bftab{0.5080} & \bftab{1.5680} \\
ControlNet + CPH + gated attention & 0.5059 & 0.4253 & \bftab{0.5228} & 0.4847 & 1.5942 \\
Sequential ControlNet + CPH & 0.5653 & 0.4311 & 0.5116 & \underline{0.5027} & 1.6223 \\
\bottomrule
\end{tabular}
\caption[Architecture ablation results.]{Architecture ablation results. Best values are shown in bold and second-best values are underlined.}
\label{tab:architecture_ablations}
\end{table*}

Table~\ref{tab:architecture_ablations} summarizes downstream benchmark performance. Overall, ControlNet variants substantially outperform both early and deep fusion in mean benchmark score and training loss.

While deep fusion achieves the strongest opening-following performance on LOB-P, it performs substantially worse on M1-S, indicating weaker alignment with human move distributions. In contrast, ControlNet variants perform consistently well across benchmarks, suggesting that structured residual conditioning is more effective than repeated cross-attention alone in the frozen-backbone regime.

Adding a controllable policy head further improves both benchmark performance and loss, supporting the hypothesis that direct policy modulation is beneficial. Gated attention slightly improves M1-S performance, but reduces instruction-following accuracy overall. The sequential ControlNet variant also underperforms the layer-coupled formulation and exhibits higher training loss.

Overall, the layer-coupled ControlNet with a controllable policy head achieves the best trade-off between convergence and downstream performance, and is therefore selected as the baseline architecture for subsequent experiments.

\subsubsection{Full Fine-Tuning (Retrospective Study)}
\label{app:architecture_full_finetuning}

We additionally conduct a retrospective study in which selected fusion architectures are trained with full joint optimization rather than a frozen backbone.

\begin{table*}[h!]
\centering
\small
\begin{tabular}{l S[table-format=1.4] S[table-format=1.4] S[table-format=1.4] S[table-format=1.4] | S[table-format=1.4]}
\toprule
{Experiment} & {LOB-P} & {LIF-D} & {M1-S} & {Mean} & {\(\mu_{\mathrm{loss},\,\mathrm{Dec13}}\)} \\
\midrule
Early fusion (frozen) & 0.4696 & 0.4176 & 0.4315 & 0.4396 & 1.6685 \\
Early fusion (full FT) & 0.5393 & 0.4294 & 0.5094 & 0.4927 & 1.6329 \\
Deep fusion (frozen) & \underline{0.5745} & 0.4325 & 0.3379 & 0.4483 & \bftab{1.6092} \\
Deep fusion (full FT) & 0.5742 & \bftab{0.4400} & \underline{0.5115} & \underline{0.5086} & 1.6205 \\
Deep fusion + CPH (full FT) & \bftab{0.5822} & \underline{0.4389} & \bftab{0.5122} & \bftab{0.5111} & \underline{1.6189} \\
\bottomrule
\end{tabular}
\caption[Full fine-tuning results for selected architectures.]{Full fine-tuning results for selected architectures. Best values are shown in bold and second-best values are underlined.}
\label{tab:architecture_ablations_full_finetune}
\end{table*}

Under full fine-tuning, both early and deep fusion improve substantially, particularly on \mbox{M1-S}. Deep fusion with a controllable policy head achieves the strongest overall benchmark performance.

These results are not directly comparable to the primary architecture study, which assumes a frozen backbone. Instead, they suggest that the relative effectiveness of conditioning strategies depends strongly on whether the pretrained chess policy is fixed or jointly optimized.

Finally, fusion-based methods remain substantially more parameter-efficient than ControlNet, which approximately doubles the number of trainable Lc0 parameters, highlighting a trade-off between parameter efficiency and conditioning modularity.

\subsection{Optimization Strategy}
\label{app:optimizer_ablations}

We next evaluate whether alternative optimizers improve convergence and downstream performance. Unless otherwise specified, all optimizer ablations use the same architecture, training data, and hyperparameters.

We compare three optimizers:

\begin{enumerate}
    \item \textbf{AdamW:} AdamW~\cite{loshchilov2017decoupledadamw}, used in our initial experiments.

    \item \textbf{Muon:} Muon~\cite{jordan2024muon} orthogonalizes updates for 2D non-embedding and non-output parameters. Following prior work~\cite{jordan2024muon,liu2025muon}, we use AdamW for all remaining parameters.

    \item \textbf{NorMuon:} NorMuon~\cite{li2025normuon} extends Muon with neuron-wise normalization using second-order momentum statistics.
\end{enumerate}

\begin{figure}[h!]
    \centering
    \includegraphics[width=\linewidth]{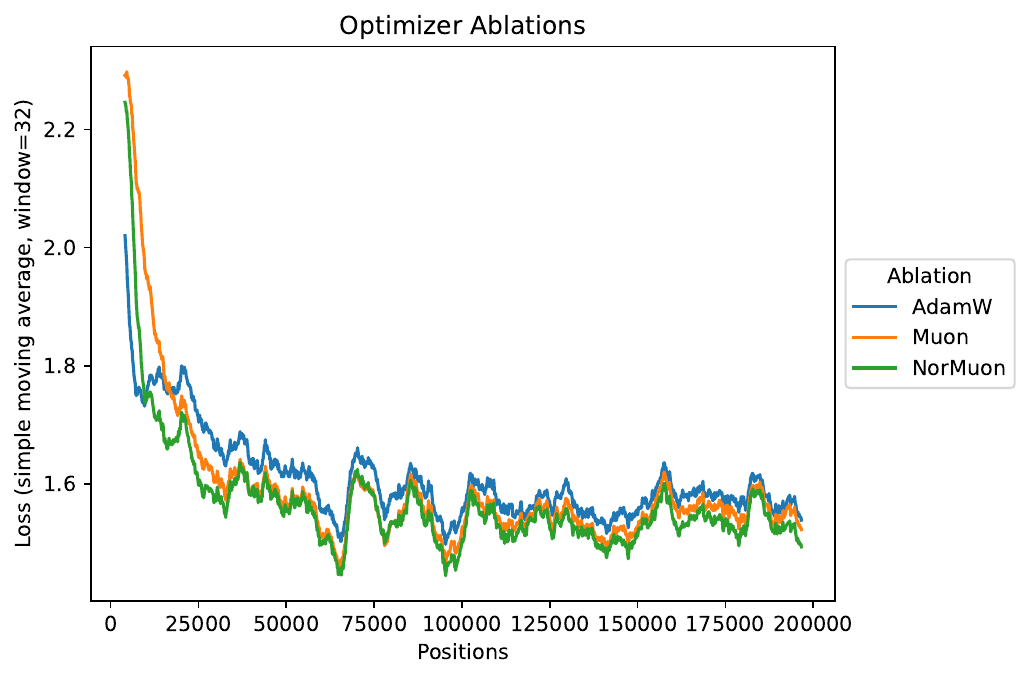}
    \caption{Simple moving averages of the training losses for each optimizer.}
    \label{fig:optimizer_ablation_losses}
\end{figure}

Figure~\ref{fig:optimizer_ablation_losses} shows the training losses. Although Muon and NorMuon initially converge more slowly, both quickly outperform AdamW and maintain lower losses throughout training.

These improvements also transfer to downstream performance (Table~\ref{tab:optimizer_ablations}). Muon improves opening-following performance and training loss over AdamW, while NorMuon achieves the best overall results. Based on these results, we use NorMuon in all subsequent experiments.

\begin{table*}[h!]
\centering
\small
\begin{tabular}{l S[table-format=1.4] S[table-format=1.4] S[table-format=1.4] S[table-format=1.4] | S[table-format=1.4]}
\toprule
{Experiment} & {LOB-P} & {LIF-D} & {M1-S} & {Mean} & {\(\mu_{\mathrm{loss},\,\mathrm{Dec13}}\)} \\
\midrule
AdamW\(^*\) & 0.5704 & \underline{0.4358} & \underline{0.5179} & 0.5080 & 1.5680 \\
Muon & \underline{0.5906} & 0.4346 & 0.5171 & \underline{0.5141} & \underline{1.5525} \\
NorMuon & \bftab{0.5919} & \bftab{0.4432} & \bftab{0.5219} & \bftab{0.5190} & \bftab{1.5331} \\
\bottomrule
\end{tabular}
\caption[Optimizer ablation results.]{Optimizer ablation results. Best values are in bold and second-best values are underlined.}
\label{tab:optimizer_ablations}
\end{table*}

\subsection{Parameter-Efficient Fine-Tuning}
\label{app:peft_ablations}

We next evaluate parameter-efficient fine-tuning (PEFT) strategies for adapting the text encoder. The baseline uses DoRA with rsLoRA and LoRA+, motivated by prior work showing improved adaptation with minimal additional overhead.

We compare:

\begin{enumerate}
    \item \textbf{DoRA:} Decomposes low-rank updates into magnitude and direction components~\cite{liu2024dora}. Unless otherwise specified, DoRA additionally uses rsLoRA and LoRA+.

    \item \textbf{LoRA:} Standard LoRA adapters~\cite{hu2022lora} without DoRA magnitude decomposition.

    \item \textbf{No rsLoRA:} Removes rsLoRA scaling~\cite{kalajdzievski2023rankrslora}.

    \item \textbf{No LoRA+:} Removes LoRA+~\cite{hayou2024lora+}, using identical learning rates for the LoRA matrices \(A\) and \(B\) instead of the default ratio \(\frac{\eta_B}{\eta_A}=2\).

    \item \textbf{Frozen text encoder:} Fully freezes the text encoder while retaining the remaining ControlNet conditioning architecture.
\end{enumerate}

\begin{figure}[t]
    \centering
    \includegraphics[width=\linewidth]{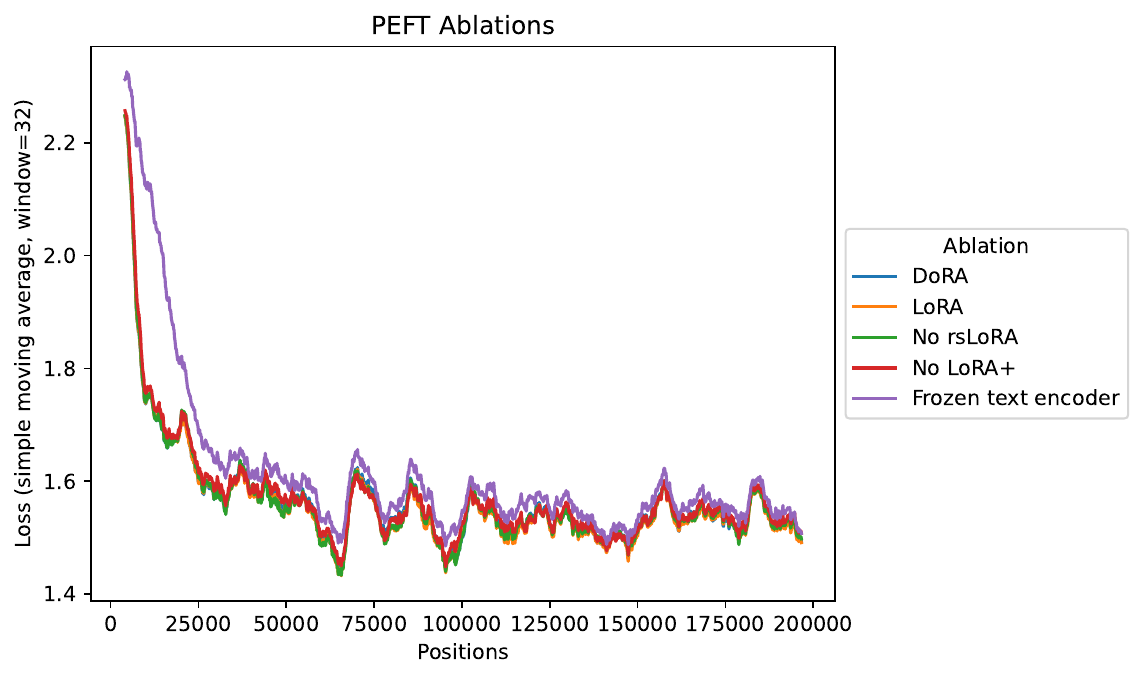}
    \caption{Simple moving averages of the training losses for the PEFT ablations.}
    \label{fig:peft_ablation_losses}
\end{figure}

\begin{table*}[t]
\centering
\small
\begin{tabular}{l S[table-format=1.4] S[table-format=1.4] S[table-format=1.4] S[table-format=1.4] | S[table-format=1.4]}
\toprule
{Experiment} & {LOB-P} & {LIF-D} & {M1-S} & {Mean} & {\(\mu_{\mathrm{loss},\,\mathrm{Dec13}}\)} \\
\midrule
DoRA\(^*\) & \underline{0.5919} & 0.4432 & \underline{0.5219} & \underline{0.5190} & \underline{1.5331} \\
LoRA & \bftab{0.6015} & 0.4408 & 0.5196 & \bftab{0.5206} & \bftab{1.5282} \\
No rsLoRA & 0.5799 & \underline{0.4441} & 0.5200 & 0.5147 & 1.5333 \\
No LoRA+ & 0.5781 & \bftab{0.4471} & \bftab{0.5246} & 0.5166 & 1.5401 \\
Frozen text encoder & 0.5545 & 0.4350 & 0.5145 & 0.5013 & 1.5501 \\
\bottomrule
\end{tabular}
\caption{PEFT ablation results. Best values are in bold and second-best values are underlined.}
\label{tab:peft_ablations}
\end{table*}

\begin{table}[t]
\centering
\small
\begin{tabular}{lc}
\toprule
{Experiment} & {Trainable LLM Parameters} \\
\midrule
DoRA\(^*\) & 4.260M \\
\textbf{LoRA} & 3.932M \\
Frozen text encoder & 0 \\
\bottomrule
\end{tabular}
\caption{Trainable text-encoder parameter counts for the PEFT variants.}
\label{tab:peft_ablation_parameter_counts}
\end{table}

Figure~\ref{fig:peft_ablation_losses} shows that all LoRA-based variants converge similarly. Although the frozen text encoder reaches a comparable loss, it consistently underperforms on downstream benchmarks, indicating that adapting the language representation remains beneficial even with a frozen chess backbone.

Among trainable variants, standard LoRA slightly outperforms DoRA in both mean benchmark score and final loss (Table~\ref{tab:peft_ablations}) while using fewer trainable parameters (Table~\ref{tab:peft_ablation_parameter_counts}), suggesting limited benefit from DoRA's additional magnitude parameters.

Removing rsLoRA or LoRA+ slightly reduces overall performance, though the effect is small. Since both introduce negligible overhead while modestly improving stability and accuracy, we retain them in the final configuration. Overall, these results suggest that lightweight low-rank adaptation is sufficient for conditioning the frozen chess policy, with more complex PEFT variants providing diminishing returns.

\subsection{Lc0 Scaling}
\label{app:lc0_scaling}

We next investigate whether scaling the frozen Lc0-ChessFormer backbone improves downstream controllability and human move prediction performance. Since the ControlNet branch mirrors the backbone architecture, increasing the size of the underlying Lc0 model approximately doubles the additional parameters introduced by the conditioning network. Consequently, backbone scaling substantially increases total model size and memory requirements.

We compare three pretrained Lc0-CF checkpoints from~\citet{monroe2024mastering,monroe2026chessformer}. All models use 15 Transformer layers with 32-dimensional attention heads, while larger variants increase the number of attention heads and embedding dimensions. All checkpoints additionally use stochastic weight averaging~\cite{izmailov2018averaging_swa}.

\begin{enumerate}
\item \textbf{t3:} A 0.10B-parameter model initialized from \texttt{t3-512x15x16h-swa-2815000}.

\item \textbf{BT3:} A 0.12B-parameter model initialized from \texttt{BT3-768x15x24h-swa-2790000}.

\item \textbf{BT4:} A 0.21B-parameter model initialized from \texttt{BT4-1024x15x32h-swa-5000000}.
\end{enumerate}

\begin{table*}[h!]
\centering
\small
\begin{tabular}{l S[table-format=1.4] S[table-format=1.4] S[table-format=1.4] S[table-format=1.4] | S[table-format=1.4]}
\toprule
{Experiment} & {LOB-P} & {LIF-D} & {M1-S} & {Mean} & {\(\mu_{\mathrm{loss},\,\mathrm{Dec13}}\)} \\
\midrule
Lc0-CF-t3 (0.10B)\(^*\) & 0.6015 & 0.4408 & 0.5196 & 0.5206 & 1.5282 \\
Lc0-CF-BT3 (0.12B) & \bftab{0.6154} & \underline{0.4432} & \bftab{0.5249} & \bftab{0.5278} & \underline{1.5136} \\
\textbf{Lc0-CF-BT4 (0.21B)} & \underline{0.6071} & \bftab{0.4449} & \underline{0.5244} & \underline{0.5255} & \bftab{1.5063} \\
\bottomrule
\end{tabular}
\caption[Lc0 backbone scaling ablation results.]{Lc0 backbone scaling ablation results. Best values are shown in bold and second-best values are underlined.}
\label{tab:lc0_ablations}
\end{table*}

Increasing the backbone width generally improves performance. While BT3 achieves the highest mean benchmark score in this ablation, BT4 attains the lowest loss and remains competitive across all benchmarks.

We note that BT3 also outperforms BT4 in the standalone Lc0 evaluation (see Appendix~\ref{app:detailed_evaluation_results}), particularly on human move matching. This likely contributes to its stronger benchmark performance in the ControlNet setting.

Nevertheless, we select BT4 as the baseline for subsequent experiments, prioritizing lower loss and higher capacity despite a small drop in benchmark performance. We do not further disentangle the effect of backbone quality versus ControlNet adaptation and leave this to future work.

\subsection{Data Sampling}
\label{app:data_sampling}

We use Unimax downsampling~\cite{chung2023unimax} to allocate a fixed training budget of plies across the dataset while preventing large groups from dominating the sample distribution. Unimax combines approximately uniform allocation with a cap on the number of samples drawn from each source; unless otherwise specified, we use a maximum of one epoch per group, allowing smaller groups to be fully utilized while truncating larger groups to satisfy the global budget. In all experiments, Unimax is first applied across Elo buckets. This ablation studies how to perform additional downsampling \emph{within} each Elo bucket.

We consider two independent design choices:
\begin{enumerate}
    \item \textbf{Group-level balancing}: whether to partition games into groups (e.g., by opening or time control) and allocate separate budgets to each group using Unimax.
    \item \textbf{Per-game budget allocation}: whether to allocate plies proportionally to game length or using Unimax, which caps the contribution of long games.
\end{enumerate}

In all settings, plies are sampled uniformly within each game after budgets are assigned. Unless otherwise specified, the baseline uses Unimax grouped by opening.

\begin{table*}[h!]
\centering
\small
\begin{tabular}{l S[table-format=1.4] S[table-format=1.4] S[table-format=1.4] S[table-format=1.4] | S[table-format=1.4]}
\toprule
{Experiment} & {LOB-P} & {LIF-D} & {M1-S} & {Mean} & {\(\mu_{\mathrm{loss},\,\mathrm{Dec13}}\)} \\
\midrule
Proportional (global) & 0.6357 & 0.4469 & 0.5312 & 0.5380 & 1.5500 \\
Proportional (opening) & 0.6154 & 0.4492 & 0.5319 & 0.5321 & 1.5651 \\
Proportional (opening + time control) & 0.6384 & \underline{0.4503} & 0.5260 & 0.5382 & 1.6006 \\
Proportional (time control) & \underline{0.6503} & \bftab{0.4503} & 0.5312 & \underline{0.5440} & 1.5192 \\
Unimax (global) & 0.6228 & 0.4472 & \bftab{0.5322} & 0.5341 & \bftab{1.4716} \\
Unimax (opening)\(^*\) & 0.6071 & 0.4449 & 0.5244 & 0.5255 & 1.5063 \\
Unimax (opening + time control) & 0.6182 & 0.4494 & 0.5272 & 0.5316 & 1.5562 \\
\textbf{Unimax (time control)} & \bftab{0.6580} & 0.4450 & \underline{0.5320} & \bftab{0.5450} & \underline{1.4768} \\
\bottomrule
\end{tabular}
\caption[Downsampling ablation results.]{Downsampling ablation results. Best values are in bold and second-best values are underlined.}
\label{tab:downsampling_ablations}
\end{table*}

Table~\ref{tab:downsampling_ablations} shows that balancing across time controls consistently outperforms balancing across openings. This suggests that improving coverage across time controls improves generalization, whereas enforcing a uniform opening distribution may distort the naturally skewed opening frequencies observed in Lichess. Although proportional allocation occasionally improves individual metrics, Unimax-based per-game allocation yields better overall benchmark performance. Consequently, we adopt Unimax (time control) for subsequent experiments.

\subsection{LoRA Rank Ablations}
\label{app:lora_rank_ablations}

We evaluate LoRA ranks \(2^k\) for \(k \in [0,7]\) while keeping the LoRA scaling factor (\(\alpha\)) fixed.

\begin{table*}[h!]
\centering
\small
\begin{tabular}{l S[table-format=1.4] S[table-format=1.4] S[table-format=1.4] S[table-format=1.4] | S[table-format=1.4]}
\toprule
{Experiment} & {LOB-P} & {LIF-D} & {M1-S} & {Mean} & {\(\mu_{\mathrm{loss},\,\mathrm{Dec13}}\)} \\
\midrule
LoRA rank = 1 & 0.6494 & 0.4433 & 0.5264 & 0.5397 & 1.4863 \\
LoRA rank = 2 & 0.6340 & 0.4425 & 0.5299 & 0.5355 & 1.4874 \\
LoRA rank = 4 & 0.6504 & 0.4442 & \underline{0.5300} & 0.5415 & 1.4866 \\
LoRA rank = 8\(^*\) & \underline{0.6559} & \bftab{0.4457} & \bftab{0.5304} & \underline{0.5440} & 1.4863 \\
\textbf{LoRA rank = 16} & \bftab{0.6625} & 0.4448 & 0.5297 & \bftab{0.5457} & 1.4874 \\
LoRA rank = 32 & 0.6540 & \underline{0.4453} & \underline{0.5300} & 0.5431 & 1.4892 \\
LoRA rank = 64 & 0.6545 & 0.4428 & 0.5299 & 0.5424 & \bftab{1.4832} \\
LoRA rank = 128 & 0.6467 & 0.4429 & 0.5247 & 0.5381 & \underline{1.4843} \\
\bottomrule
\end{tabular}
\caption[LoRA rank ablation results.]{LoRA rank ablation results. Best values are in bold and second-best values are underlined.}
\label{tab:lora_rank_ablations}
\end{table*}

Table~\ref{tab:lora_rank_ablations} shows that benchmark performance peaks at rank 16, with rank 8 performing similarly. In contrast, training loss does not vary monotonically with rank and is only weakly correlated with downstream performance.

Higher LoRA ranks increase the number of trainable parameters, but yield diminishing returns beyond moderate ranks. Combined with the PEFT ablations in Table~\ref{tab:peft_ablations}, these results suggest that most gains arise from introducing low-rank adaptation itself, while increasing the adaptation capacity beyond a moderate rank provides comparatively small improvements.

\subsection{Learning Rate Schedule Ablations}
\label{app:learning_rate_schedule_ablations}

We compare our custom learning rate schedule against simplified Warmup--Stable--Decay (WSD-S) schedules proposed by Wen et al.~\cite{wen2024understanding}. Since the ablations are restricted to 2013 data, the custom schedule behaves similarly to a cosine-like decay.

We evaluate two WSD-S variants:
\begin{enumerate}
    \item \textbf{WSD-S (monthly):} warmup and decay are applied independently each month;
    \item \textbf{WSD-S (annual):} a single warmup--stable--decay cycle is applied across all of 2013.
\end{enumerate}

Because the original schedule spends nearly all of January 2013 in warmup, we shorten the warmup phase to 512 steps and allocate an additional 512 decay steps, leaving \num{1024} stable-training steps.

\begin{figure}[h!]
    \centering
    \includegraphics[width=\linewidth]{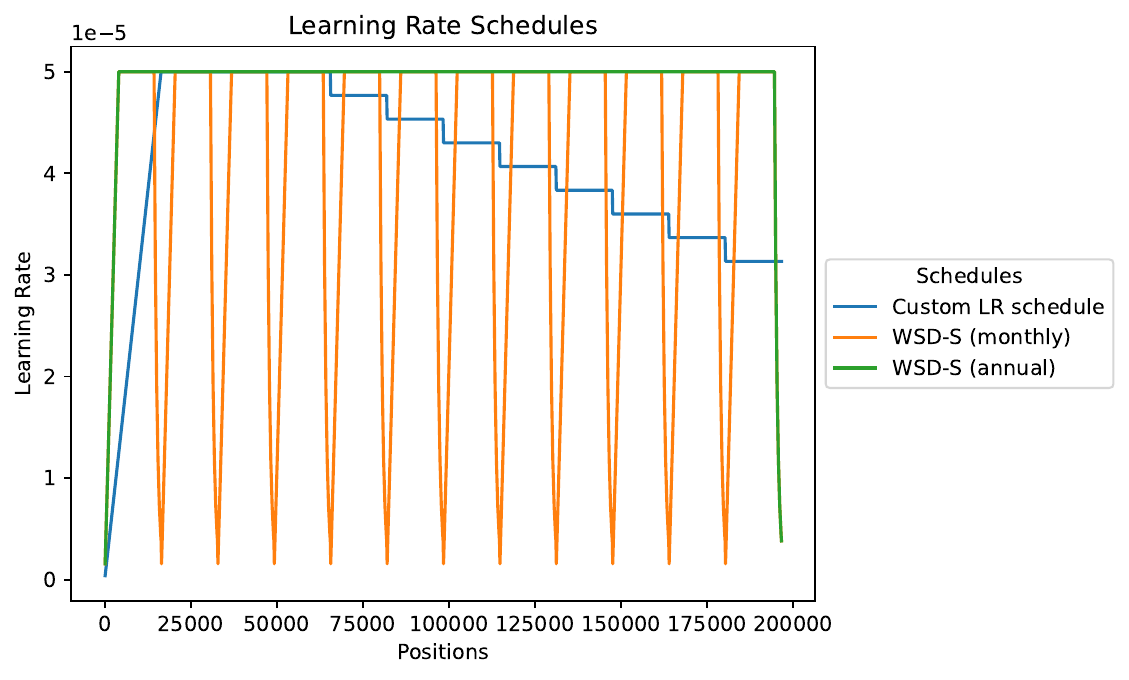}
    \caption{Learning rate schedules evaluated in the ablation study.}
    \label{fig:learning_rate_schedules}
\end{figure}

\begin{figure}[h!]
    \centering
    \includegraphics[width=\linewidth]{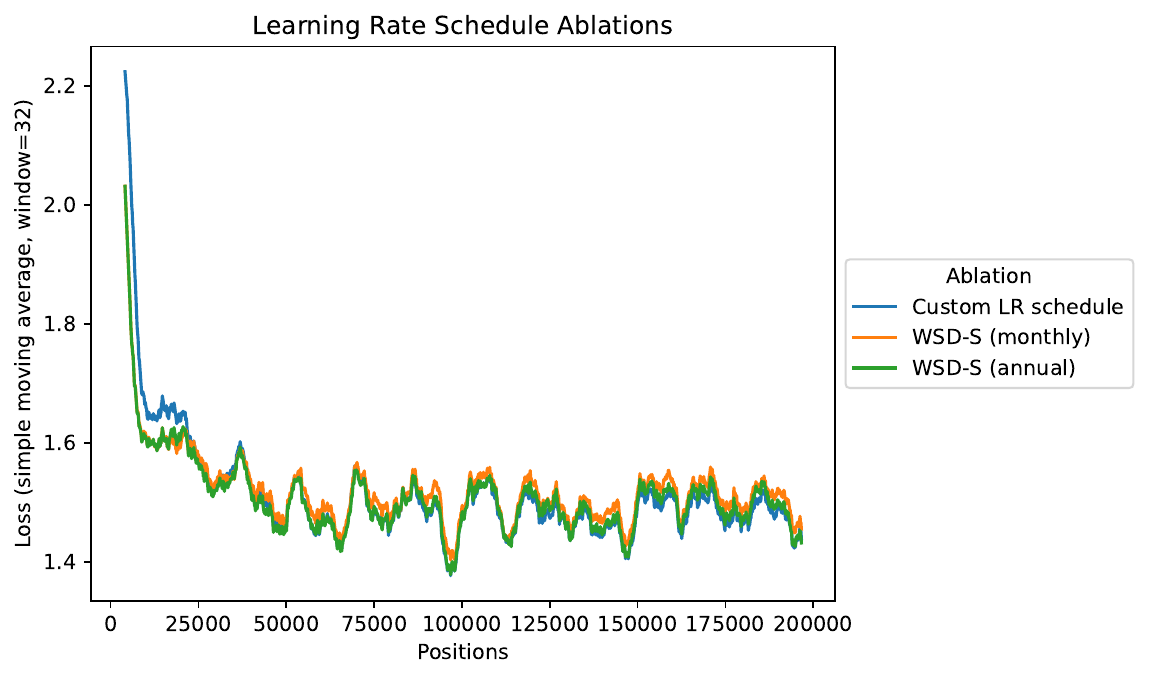}
    \caption[Simple moving averages of the training losses for the learning rate schedule ablations.]{Simple moving averages of the training losses for the learning rate schedule ablations. The WSD-S schedules converge faster initially due to their shorter warmup, but the custom schedule catches up during the decay phase.}
    \label{fig:learning_rate_schedule_ablation_losses}
\end{figure}

\begin{table*}[h!]
\centering
\small
\begin{tabular}{l S[table-format=1.4] S[table-format=1.4] S[table-format=1.4] S[table-format=1.4] | S[table-format=1.4]}
\toprule
{Experiment} & {LOB-P} & {LIF-D} & {M1-S} & {Mean} & {\(\mu_{\mathrm{loss},\,\mathrm{Dec13}}\)} \\
\midrule
Custom LR schedule\(^*\) & \bftab{0.6580} & \bftab{0.4450} & \bftab{0.5320} & \bftab{0.5450} & \bftab{1.4768} \\
WSD-S (monthly) & 0.6019 & 0.4407 & 0.5222 & 0.5216 & 1.5014 \\
WSD-S (annual) & \underline{0.6389} & \underline{0.4434} & \underline{0.5226} & \underline{0.5350} & \underline{1.4845} \\
\bottomrule
\end{tabular}
\caption[Learning rate schedule ablation results.]{Learning rate schedule ablation results. Best values are in bold and second-best values are underlined.}
\label{tab:learning_rate_schedule_ablations}
\end{table*}

As shown in Figure~\ref{fig:learning_rate_schedule_ablation_losses}, the WSD-S schedules reduce loss more rapidly early in training due to their shorter warmup phases. However, the custom schedule achieves lower final loss and better benchmark performance overall (Table~\ref{tab:learning_rate_schedule_ablations}), particularly after the WSD-S variants enter their decay phase.

Although the custom schedule performs best overall, the ablation suggests that the original warmup duration is unnecessarily long. Consequently, we reduce the warmup length from \num{2048} to \num{1024} steps in the full training run.

\subsection{Negative Results}
\label{app:negative_results}

This section summarizes exploratory ablations that did not affect the final UniMaia configuration, but nevertheless provide insight into training behavior and data composition effects.

\subsubsection{Template Mixture Ablations}
\label{app:template_mixture_ablations}

We evaluate different sampling mixtures between \textsc{LichessTemplates-Pretrain} and \textsc{LichessTemplates-Instruct}:

\begin{enumerate}
    \item \textbf{50/50 mixture}: sample from both template families with equal probability;
    \item \textbf{Pretrain only}: sample exclusively from \textsc{LichessTemplates-Pretrain};
    \item \textbf{Instruct only}: sample exclusively from \textsc{LichessTemplates-Instruct}.
\end{enumerate}

\begin{table*}[h]
\centering
\small
\begin{tabular}{l S[table-format=1.4] S[table-format=1.4] S[table-format=1.4] S[table-format=1.4] | S[table-format=1.4]}
\toprule
{Experiment} & {LOB-P} & {LIF-D} & {M1-S} & {Mean} & {\(\mu_{\mathrm{loss},\,\mathrm{Dec13}}\)} \\
\midrule
\textbf{50/50 pretrain, instruct templates}\(^*\) & \underline{0.5704} & 0.4358 & \underline{0.5179} & \underline{0.5118} & \underline{1.5680} \\
Pretrain templates & 0.4401 & \bftab{0.4422} & 0.5104 & 0.4794 & \bftab{1.5507} \\
Instruct templates & \bftab{0.5819} & \underline{0.4402} & \bftab{0.5221} & \bftab{0.5173} & 1.5842 \\
\bottomrule
\end{tabular}
\caption[Template mixture ablation results.]{Template mixture ablation results. Best values are in bold and second-best values are underlined.}
\label{tab:data_ablations}
\end{table*}

Using only instruct templates yields the strongest average benchmark performance, while the 50/50 mixture achieves comparable downstream accuracy with lower training loss. We therefore adopt the mixed-template configuration in the final model.

One possible explanation is that the instruct template pool (\num{2048} templates) is substantially larger than the pretrain pool (160 templates), making the pretrain-only setting more susceptible to overfitting. Nevertheless, we retain the mixed setting because the pretrain templates contain richer structured metadata, which may provide a stronger semantic grounding signal during training.

We also experiment with restricting training to rapid games and omitting the first 10 plies, following Maia-2 Rapid~\cite{tang2024maia2}. Although this substantially lowers training loss, it produces only marginal gains on rapid-only human move prediction benchmarks. Since UniMaia is intended as a unified prompt-conditioned model across time controls, we do not pursue this direction further. This result is also consistent with prior work such as Allie~\cite{zhang2024humanallie}, which shows that unified multi-time-control models can match or outperform specialized rapid-only systems.

\subsubsection{Text Model}
\label{app:llm_ablations}

We evaluate several text encoders differing in architecture family, parameter scale, and training data composition. Specifically, we compare pretrained and chess-specialized models from the GPT-NeoX~\cite{black2022gpt}, Gemma~\cite{gemmateam2025gemma3technicalreport}, and Qwen~\cite{zhang2025qwen3embedding} families, including autoregressive, encoder-decoder, and embedding-focused variants.

Within the GPT-NeoX family, we compare RedPajama INCITE Base (3B)~\cite{weber2024redpajama}, ChessGPT-base, and ChessGPT-chat~\cite{feng2024chessgpt}. We additionally evaluate Gemma 3 (1B), T5 Gemma 2 encoder variants~\cite{tay2022ul2,zhang2025t5gemma2}, Embedding Gemma~\cite{kusupati2022matryoshka,vera2025embeddinggemma}, and Qwen 3 Embedding (0.6B).

\begin{table*}[h!]
\centering
\small
\begin{tabular}{l S[table-format=1.4] S[table-format=1.4] S[table-format=1.4] S[table-format=1.4] | S[table-format=1.4]}
\toprule
{Experiment} & {LOB-P} & {LIF-D} & {M1-S} & {Mean} & {\(\mu_{\mathrm{loss},\,\mathrm{Dec13}}\)} \\
\midrule
Gemma 3 (1B) & 0.4817 & 0.4275 & 0.5049 & 0.4714 & 1.6049 \\
T5Gemma 2 Encoder (1B) & 0.4555 & 0.4287 & 0.5056 & 0.4632 & 1.6111 \\
T5Gemma 2 Encoder (0.27B) & 0.5168 & 0.4329 & 0.5161 & 0.4886 & 1.5873 \\
Embedding Gemma (0.30B) & 0.5273 & 0.4381 & \bftab{0.5212} & 0.4955 & \underline{1.5512} \\
Qwen 3 Embedding (0.60B) & 0.5772\rlap{\textsuperscript{\dag}} & \bftab{0.4441}\rlap{\textsuperscript{\dag}} & \underline{0.5207}\rlap{\textsuperscript{\dag}} & \underline{0.5140}\rlap{\textsuperscript{\dag}} & 1.5733\rlap{\textsuperscript{\dag}} \\
Redpajama Incite Base (3B) & 0.5108 & 0.4341 & 0.5100 & 0.4850 & 1.5916 \\
\textbf{ChessGPT-base v1 (3B)}\(^*\) & \bftab{0.6015} & 0.4408 & 0.5196 & \bftab{0.5206} & \bftab{1.5282} \\
ChessGPT-chat v1 (3B) & \underline{0.5778} & \underline{0.4433} & 0.5147 & 0.5119 & 1.5617 \\
\bottomrule
\end{tabular}
\caption[LLM ablation results.]{LLM ablation results. Best values are in bold and second-best values are underlined. \textsuperscript{\dag}The corresponding LoRA model produced NaN losses, so benchmark scores and loss are instead reported using DoRA.}
\label{tab:llm_ablations}
\end{table*}

Model scale alone does not consistently improve performance. For example, the larger T5 Gemma 2 encoder underperforms its smaller counterpart. Instead, training data composition appears more important: models trained on both chess and natural language data consistently outperform general-purpose pretrained models. Embedding-oriented models also perform competitively despite their smaller size.

Overall, ChessGPT-base achieves the best average performance and lowest training loss, suggesting that domain-specialized pretraining is more beneficial than increasing model scale alone. We therefore retain ChessGPT-base for all subsequent experiments.

\subsubsection{Fine-tuning ChessGPT-base}
\label{app:fine_tuning_llm_ablations}

We investigate whether fine-tuning the language model prior to ControlNet training improves downstream instruction-following performance.

\begin{figure}[h!]
    \centering
    \includegraphics[width=\linewidth]{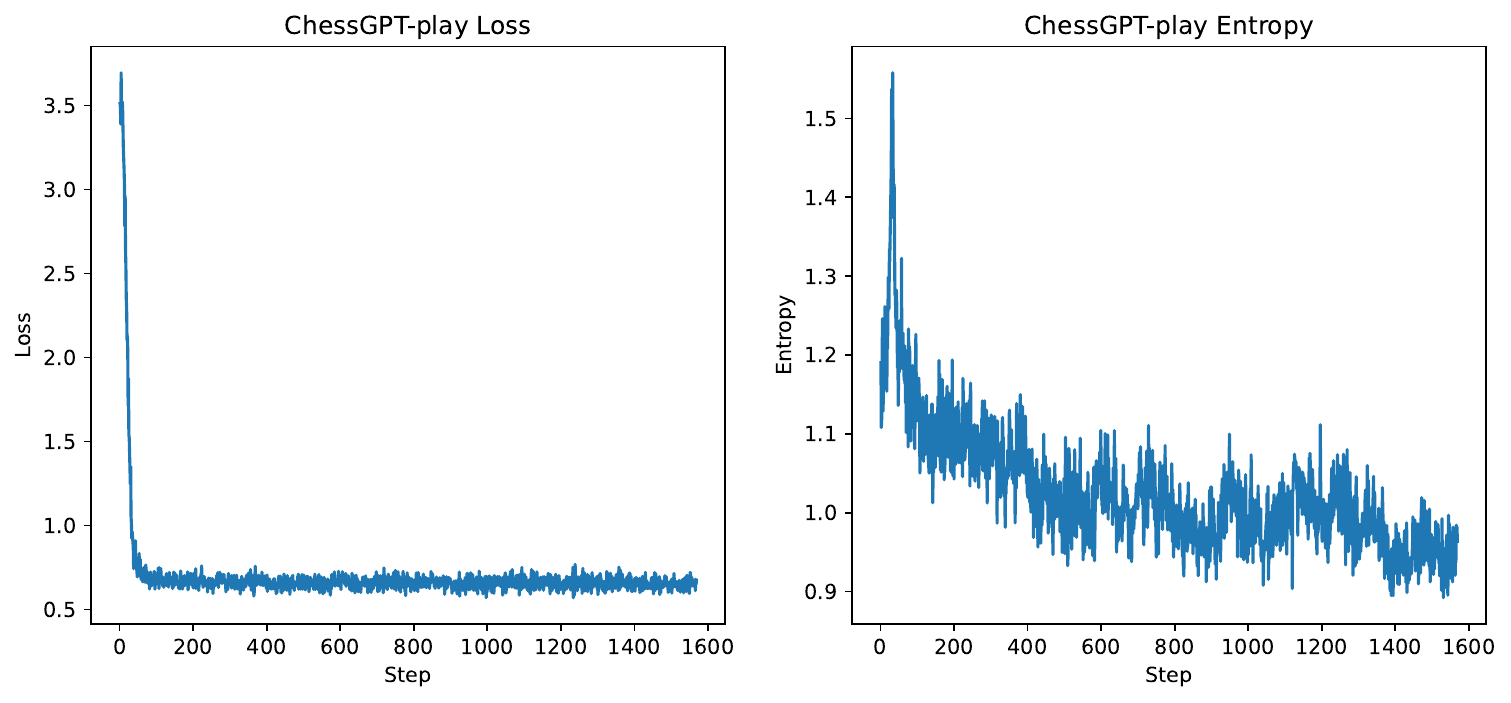}
    \caption[Training loss (left) and entropy (right) during ChessGPT-play training.]{Training loss (left) and entropy (right) during ChessGPT-play training. While loss plateaus early, entropy continues to decrease, indicating increasingly confident predictions.}
    \label{fig:chessgpt_play_training_stats}
\end{figure}

We evaluate four variants:

\begin{enumerate}
\item \textbf{ChessGPT-play:} ChessGPT-base fine-tuned directly for next-move prediction on the 2013 split using the template \texttt{\{prompt\}<|endoftext|>\{pgn\}}.

\item \textbf{ControlNet (ChessGPT-base):} The standard ControlNet configuration using the pretrained ChessGPT-base encoder.

\item \textbf{ControlNet (ChessGPT-play):} The same ControlNet configuration, but replacing the text encoder with ChessGPT-play.

\item \textbf{ControlNet (ChessGPT-play), end-of-text PGNs:} A variant using the explicit end-of-text separator \texttt{\{prompt\}<|endoftext|>\{pgn\}} during ControlNet training to reduce distribution shift relative to ChessGPT-play pretraining.
\end{enumerate}

ChessGPT-play converges rapidly within the first \(\sim50\) steps (Figure~\ref{fig:chessgpt_play_training_stats}). Interestingly, prediction entropy continues decreasing after the loss plateaus, suggesting increasing confidence without corresponding improvements in accuracy.

\begin{table*}[h!]
\centering
\small
\begin{tabular}{l S[table-format=1.4] S[table-format=1.4] S[table-format=1.4] S[table-format=1.4] | S[table-format=1.4]}
\toprule
{Experiment} & {LOB-P} & {LIF-D} & {M1-S} & {Mean} & {\(\mu_{\mathrm{loss},\,\mathrm{Dec13}}\)} \\
\midrule
ChessGPT-play & \bftab{0.8080} & 0.3975 & 0.4343 & \bftab{0.5466} & \bftab{0.6530} \\
ControlNet (ChessGPT-base)\(^*\) & \underline{0.6625} & \underline{0.4448} & \underline{0.5297} & \underline{0.5457} & 1.4874 \\
ControlNet (ChessGPT-play) & 0.6355 & 0.4445 & \bftab{0.5314} & 0.5371 & 1.4975 \\
ControlNet (ChessGPT-play), end-of-text PGN & 0.6494 & \bftab{0.4453} & 0.5281 & 0.5409 & \underline{1.4872} \\
\bottomrule
\end{tabular}
\caption[Fine-tuned base model ablation results.]{Fine-tuned base model ablation results. Best values are in bold and second-best values are underlined.}
\label{tab:fine-tuned_base_model_ablations}
\end{table*}

Although ChessGPT-play performs strongly on the opening-following benchmark, it substantially underperforms on LIF-D and M1-S. Replacing the pretrained ChessGPT-base encoder with ChessGPT-play similarly degrades overall ControlNet performance, suggesting that specialization toward direct move prediction harms general instruction-following behavior.

To further evaluate opening-following performance, we additionally test the models on canonical opening reconstruction benchmarks.

\begin{table*}[h!]
\centering
\small
\begin{tabular}{l S[table-format=1.4] S[table-format=1.4] S[table-format=1.4] S[table-format=1.4] | S[table-format=1.4]}
\toprule
{Experiment} & {LOB-P} & {LOB-C} & {LOB-P\rlap{\textsuperscript{\dag}}} & {LOB-C\rlap{\textsuperscript{\dag}}} & {Mean} \\
\midrule
ChessGPT-play & \bftab{0.8080} & \bftab{0.8799} & \bftab{0.8305} & \bftab{0.8327} & \bftab{0.8378} \\
ControlNet (ChessGPT-base)\(^*\) & \underline{0.6625} & \underline{0.6844} & \underline{0.6569} & \underline{0.6608} & \underline{0.6661} \\
ControlNet (ChessGPT-play) & 0.6355 & 0.6523 & 0.6119 & 0.6151 & 0.6287 \\
ControlNet (ChessGPT-play), end-of-text PGN & 0.6494 & 0.6715 & 0.6388 & 0.6420 & 0.6504 \\
\bottomrule
\end{tabular}
\caption[Fine-tuned base model ablations evaluated on opening instruction-following benchmarks.]{Fine-tuned base model ablations evaluated on opening instruction-following benchmarks. Best values are in bold and second-best values are underlined. \textsuperscript{\dag}Prompted with the PGN header.}
\label{tab:fine-tuned_base_model_ablations_opening}
\end{table*}

As shown in Table~\ref{tab:fine-tuned_base_model_ablations_opening}, the strong opening-following performance of ChessGPT-play does not transfer when the model is used as the ControlNet text encoder. Adding an explicit end-of-text separator slightly improves performance, but remains below the original ChessGPT-base configuration. Interestingly, inference performance for the end-of-text variant remains nearly unchanged even when the separator and PGN are omitted, although we do not report these results.

\subsubsection{LoRA Target Modules Ablations}
\label{app:lora_target_modules_ablations}

We vary the set of modules targeted by LoRA in the text encoder:

\begin{enumerate}
\item \textbf{QKV projections:} Apply LoRA only to the query, key, and value projections.
\item \textbf{QKV + O projections:} Additionally adapt the output projection.
\item \textbf{QKV + O + FFN:} Further extend LoRA to the feedforward layers.
\end{enumerate}

\begin{table*}[h!]
\centering
\small
\begin{tabular}{l S[table-format=1.4] S[table-format=1.4] S[table-format=1.4] S[table-format=1.4] | S[table-format=1.4]}
\toprule
{Experiment} & {LOB-P} & {LIF-D} & {M1-S} & {Mean} & {\(\mu_{\mathrm{loss},\,\mathrm{Dec13}}\)} \\
\midrule
LoRA target = \(W_{QKV}\) & \bftab{0.6455} & \bftab{0.4445} & 0.5260 & \underline{0.5387} & \underline{1.4867} \\
\textbf{LoRA target} = \(W_{QKV},\,W_O\)\(^*\) & \underline{0.6443} & \underline{0.4426} & \bftab{0.5315} & \bftab{0.5394} & 1.4871 \\
LoRA target = \(W_{QKV},\,W_O,\,FFN\) & 0.6407 & 0.4419 & \underline{0.5288} & 0.5371 & \bftab{1.4855} \\
\bottomrule
\end{tabular}
\caption[LoRA target modules ablation results.]{LoRA target modules ablation results. Best values are in bold and second-best values are underlined.}
\label{tab:lora_target_modules_ablations}
\end{table*}

Table~\ref{tab:lora_target_modules_ablations} shows that all configurations perform similarly. Targeting both QKV and O projections achieves the highest mean score, while adding FFN adapters slightly reduces downstream performance despite a marginally lower loss. The differences are small and likely within the variance induced by random initialization. Overall, expanding LoRA beyond attention projections provides limited benefit in this setting and increases parameter count with diminishing returns.

\subsubsection{LoRA Rank with Scaled Alpha Ablations}
\label{app:lora_rank_scaled_alpha_ablations}

We additionally scale the LoRA scaling factor with rank while keeping the same rsLoRA multiplier used in the rank-16 setting from Appendix~\ref{app:lora_rank_ablations}. Specifically, we evaluate ranks \(2^k\) with \(\alpha = 4 \cdot 2^{k/2}\) for \(k \in [2, 6]\).

\begin{table*}[h!]
\centering
\small
\begin{tabular}{l S[table-format=1.4] S[table-format=1.4] S[table-format=1.4] S[table-format=1.4] | S[table-format=1.4]}
\toprule
{Experiment} & {LOB-P} & {LIF-D} & {M1-S} & {Mean} & {\(\mu_{\mathrm{loss},\,\mathrm{Dec13}}\)} \\
\midrule
LoRA rank = 4, \(\alpha = 8\) & 0.6511 & 0.4438 & 0.5272 & 0.5407 & \bftab{1.4866} \\
LoRA rank = 8, \(\alpha = 8\sqrt{2}\) & 0.6441 & 0.4427 & 0.5288 & 0.5385 & 1.4908 \\
LoRA rank = 16, \(\alpha = 16\)\(^*\) & \bftab{0.6625} & \underline{0.4448} & 0.5297 & \bftab{0.5457} & 1.4874 \\
LoRA rank = 32, \(\alpha = 16\sqrt{2}\) & \underline{0.6546} & \bftab{0.4456} & \bftab{0.5338} & \underline{0.5447} & 1.4870 \\
LoRA rank = 64, \(\alpha = 32\) & 0.6468 & 0.4406 & \underline{0.5327} & 0.5400 & \underline{1.4868} \\
\bottomrule
\end{tabular}
\caption[LoRA rank with scaled \(\alpha\) ablation results.]{LoRA rank with scaled \(\alpha\) ablation results. Best values are in bold and second-best values are underlined.}
\label{tab:lora_rank_scaled_alpha_ablations}
\end{table*}

Table~\ref{tab:lora_rank_scaled_alpha_ablations} shows that performance again peaks at moderate ranks, with rank 16 achieving the highest mean score. Increasing the rank beyond this point provides only marginal or inconsistent improvements despite the additional parameters.

Compared to the fixed-\(\alpha\) setting (Table~\ref{tab:lora_rank_ablations}), scaling \(\alpha\) with rank produces a similar trend, suggesting that the effective capacity of the adapters saturates in this regime. Although scaling \(\alpha\) stabilizes training across ranks, it does not substantially change the optimal rank, reinforcing that moderate-rank adapters provide the best trade-off between performance and parameter efficiency.

\subsubsection{NorMuon Hyperparameter Ablations}
\label{app:normuon_hyperparameter_ablations}

We additionally evaluate Nesterov momentum~\cite{dozat2016incorporating}, but observe slightly worse performance. Replacing the Newton--Schulz iteration with the Polar Express algorithm~\cite{amsel2025polar} for approximating \(\textrm{polar}(\cdot)\) occasionally introduces gradient spikes in the LoRA adapters, slightly increasing loss and making NaNs more likely in some execution environments. Consistent with observations from the Kimi Team~\cite{team2025kimi}, Muon also produces exploding attention logits more frequently than AdamW. DoRA adapters are generally more stable than LoRA under Polar Express.

\subsubsection{Key Padding Mask Ablations}
\label{app:key_padding_mask_ablations}

We evaluate whether masking padding tokens in cross-attention improves performance. Table~\ref{tab:key_padding_mask_ablations} shows that applying a key padding mask slightly degrades overall performance despite a marginal improvement on \textsc{LIF-D}. Adding gated attention~\cite{qiu2025gated} further reduces performance and increases loss. We therefore omit both modifications in the final configuration.

\begin{table*}[h!]
\centering
\small
\begin{tabular}{l S[table-format=1.4] S[table-format=1.4] S[table-format=1.4] S[table-format=1.4] | S[table-format=1.4]}
\toprule
{Experiment} & {LOB-P} & {LIF-D} & {M1-S} & {Mean} & {\(\mu_{\mathrm{loss},\,\mathrm{Dec13}}\)} \\
\midrule
Without key padding mask\(^*\) & \bftab{0.6625} & \underline{0.4448} & \bftab{0.5297} & \bftab{0.5457} & \bftab{1.4874} \\
With key padding mask & \underline{0.6496} & \bftab{0.4459} & \underline{0.5287} & \underline{0.5414} & \underline{1.4900} \\
With key padding mask and gated attention & 0.5338 & 0.4298 & 0.5256 & 0.4964 & 1.5354 \\
\bottomrule
\end{tabular}
\caption[Key padding mask ablation results.]{Key padding mask ablation results. Best values are in bold and second-best values are underlined.}
\label{tab:key_padding_mask_ablations}
\end{table*}

\subsection{Reproducibility and Sources of Non-Determinism}
\label{app:reproducibility}

Despite fixing random seeds for Python, NumPy, and PyTorch (CPU and CUDA), we observe small differences in training trajectories across execution environments. Runs performed locally and on university servers exhibit consistent high-level behavior, with major loss peaks and dips aligning with dataset shard boundaries. However, within-shard dynamics differ systematically between environments, likely due to differences in data loading pipelines (e.g., streaming versus local parquet access with an \texttt{IterableDataset}) that alter sample ordering within shards.

Within a fixed environment, repeated runs produce nearly identical trajectories, with only minor numerical variation attributable to known GPU and multi-worker data-loading non-determinism. Overall, reproducibility is preserved at a coarse level, although exact optimization dynamics remain sensitive to data ordering and system-level factors.

\begin{figure}[h!]
    \centering
    \includegraphics[width=\linewidth]{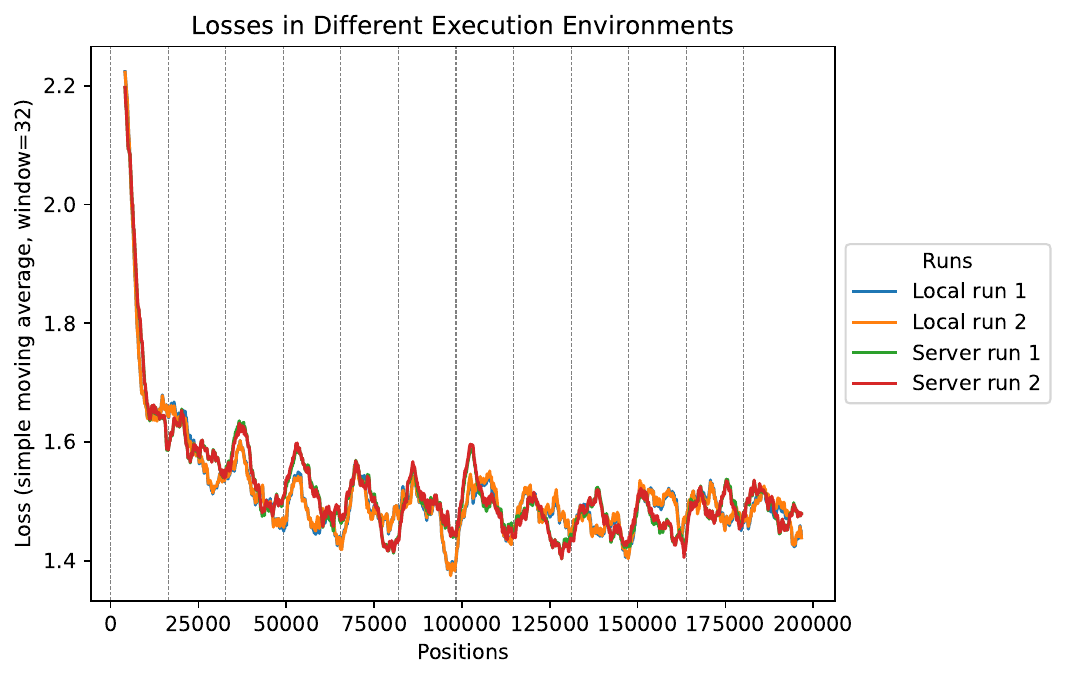}
    \caption[Training losses across repeated local and server runs.]{
    Training losses for two local and two server runs with identical training configurations. Runs within the same environment exhibit highly similar trajectories, while cross-environment runs differ slightly in their within-shard dynamics. Each shard contains \(2^{14}\) plies.
    }
    \label{fig:reproducibility_losses}
\end{figure}

Figure~\ref{fig:reproducibility_losses} illustrates these effects, while Table~\ref{tab:reproducibility_runs} shows that benchmark performance remains highly consistent across repeated runs despite slight differences in final losses.

\begin{table*}[h!]
\centering
\small
\begin{tabular}{l S[table-format=1.4] S[table-format=1.4] S[table-format=1.4] S[table-format=1.4] | S[table-format=1.4]}
\toprule
{Experiment} & {LOB-P} & {LIF-D} & {M1-S} & {Mean} & {\(\mu_{\mathrm{loss},\,\mathrm{Dec13}}\)} \\
\midrule
Local run 1 & \bftab{0.6580} & 0.4450 & \bftab{0.5320} & \bftab{0.5450} & \bftab{1.4768} \\
Local run 2 & 0.6429 & \bftab{0.4470} & 0.5254 & 0.5385 & \underline{1.4813} \\
Server run 1 & \underline{0.6559} & \underline{0.4457} & 0.5304 & \underline{0.5440} & 1.4863 \\
Server run 2 & 0.6443 & 0.4426 & \underline{0.5315} & 0.5394 & 1.4871 \\
\bottomrule
\end{tabular}
\caption[Results across repeated local and server runs.]{
Results across repeated local and server runs. Best values are in bold and second-best values are underlined.
}
\label{tab:reproducibility_runs}
\end{table*}

%% file: appendices/appendix-full-training-configuration.tex
\section{Final Training Configuration}
\label{app:final_training_configuration}

This section summarizes the configurations used for the final \textbf{UniMaia} and \textbf{UniMaia-Aux} training runs.

\subsection{UniMaia}
\label{app:unimaia_final_configuration}

UniMaia is trained on Lichess games from January 2013 to December 2022, corresponding to approximately \num{66.978}B plies. Games from 2023 are reserved for evaluation. Following Appendix~\ref{app:data_sampling}, we apply two-stage Unimax downsampling~\cite{chung2023unimax}: first across Elo partitions, and then independently within each time-control group using a fixed turn budget. Prompts are sampled uniformly from \textsc{LichessTemplates-Pretrain} and \textsc{LichessTemplates-Instruct}.

The model uses the BT4 Lc0-CF backbone~\cite{monroe2024mastering,monroe2026chessformer} within the layer-coupled ControlNet architecture described in Section~\ref{sec:method}. The frozen Lc0 policy network is conditioned through a text encoder adapted using LoRA with rsLoRA and LoRA+, using rank~16 adapters.

Optimization uses NorMuon for all 2D non-embedding, non-output parameters and AdamW for the remaining parameters. We use weight decay \(0.01\), \(\epsilon=\num{1e-8}\), gradient clipping with norm \(1.0\), and a LoRA+ learning-rate ratio of \(2.0\). Training uses a batch size of \(64\).

The learning rate is linearly warmed up over the first \num{1024} steps to a peak learning rate of \num{5e-5}. From April 2013 to June 2014, the learning rate decays linearly through monthly updates to \num{1.5e-5}. It is then held constant until 2022, after which it decays monthly toward \num{5e-7} by the end of training.

Training proceeds sequentially over monthly dataset partitions, each divided into independently shuffled shards to support efficient streaming and checkpoint recovery. Since training consists of a single epoch over the sampled data, the training loss is equivalent to the validation loss; we therefore report average monthly loss values throughout the paper. Monthly checkpoints are evaluated on the benchmark suite described in Appendix~\ref{app:baseline_details}. All experiments are conducted on a multi-GPU SLURM setup using two RTX Pro 6000 GPUs with fixed Python, NumPy, and PyTorch random seeds.

\begin{table}[t]
\centering
\small
\begin{tabularx}{\linewidth}{lX}
\toprule
\textbf{Setting} & \textbf{Value} \\
\midrule
Training data & Lichess 2013--2022 \\
Evaluation data & Lichess 2023 \\
Backbone & BT4 Lc0-CF \\
Architecture & ControlNet + controllable policy head \\
Frozen backbone & Yes \\
Text adaptation & LoRA \& rsLoRA \& LoRA+ \\
LoRA rank & 16 \\
Batch size & 64 \\
Peak learning rate & \num{5e-5} \\
Warmup steps & 1{,}024 \\
Optimizer & NorMuon + AdamW \\
Weight decay & 0.01 \\
Optimizer \(\epsilon\) & \num{1e-8} \\
Gradient clipping & 1.0 \\
Downsampling & Unimax by time control \\
Hardware & 2\(\times\) RTX Pro 6000 \\
\bottomrule
\end{tabularx}
\caption[UniMaia training configuration.]{UniMaia training configuration.}
\label{tab:unimaia_training_configuration}
\end{table}

\subsection{UniMaia-Aux}
\label{app:unimaia_aux_final_configuration}

UniMaia-Aux performs an additional fine-tuning stage starting from the final UniMaia checkpoint trained through December 2022. Auxiliary training uses games from 2018 only, selected to ensure reliable clock metadata and termination annotations.

The auxiliary targets consist of:
\begin{enumerate}[label=(\roman*)]
\item the final game result, represented as a categorical terminal-outcome distribution \(Z\) over \(\{-1,0,1\}\), where a white win corresponds to \(1\), a draw corresponds to \(0\), and a black win corresponds to \(-1\);
\item the number of plies remaining until the end of the game, represented as \(N-i-1\), where \(N\) is the total number of plies in the game and \(i\) is the current ply index;
\item the termination reason (e.g., checkmate, timeout, resignation, or draw by threefold repetition); and
\item the move delay of the active player, denoted by \(\Delta t_a\).
\end{enumerate}

The scalar value estimate \(V\) can then be recovered from the predicted terminal-outcome distribution by taking its expectation:
\[
V = \mathbb{E}[Z].
\]

The value and moves-left heads are implemented as trainable residual updates to the frozen Lc0 heads, while additional heads predict termination type and move delay.

Auxiliary training uses the same BT4-based ControlNet architecture, LoRA rank, optimizer setup, batch size, and downsampling strategy as UniMaia. We filter out bot and correspondence games, games missing Elo or time-control metadata, and games without clock annotations; the move-delay objective additionally requires \texttt{\%clk} PGN annotations.

All auxiliary objectives are optimized jointly. Outcome and termination predictions use cross-entropy losses, while moves-left and move-delay predictions use Huber losses. The learning rate schedule follows the same warmup as UniMaia, followed by a lower constant learning rate during fine-tuning.

\begin{table}[t]
\centering
\small
\begin{tabularx}{\linewidth}{lX}
\toprule
\textbf{Setting} & \textbf{Value} \\
\midrule
Initialization & UniMaia 2022 checkpoint \\
Auxiliary training data & Lichess 2018 \\
Backbone & BT4 Lc0-CF \\
Architecture & ControlNet + controllable policy head \\
Frozen backbone & Yes \\
LoRA rank & 16 \\
Batch size & 64 \\
Peak learning rate & \num{5e-5} \\
Warmup steps & 1{,}024 \\
Optimizer & NorMuon + AdamW \\
Auxiliary losses & Outcome, moves-left, \\
& termination, move delay \\
Move-delay source & PGN \texttt{\%clk} annotations \\
\bottomrule
\end{tabularx}
\caption[UniMaia-Aux training configuration.]{UniMaia-Aux training configuration.}
\label{tab:unimaia_aux_training_configuration}
\end{table}

%% file: appendices/appendix-training-dynamics.tex
\section{Training Dynamics}
\label{app:training_dynamics}

This appendix provides additional training dynamics analyses for UniMaia and UniMaia-Aux. The UniMaia benchmark trajectories are shown in Figure~\ref{fig:full_training_run_benchmark_scores} in the main paper and are not repeated here.

\subsection{UniMaia}
\label{app:training_dynamics_unimaia}

\begin{figure}[h!]
    \centering
    \includegraphics[width=\linewidth]{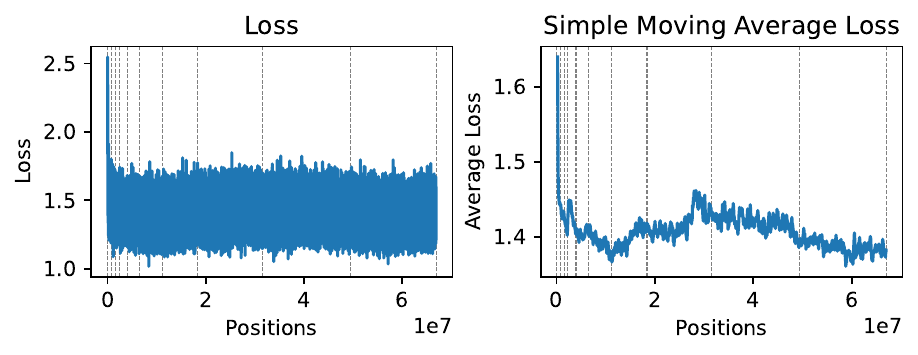}
    \caption[Training losses for the full UniMaia training run.]{
    Left: raw training loss.
    Right: simple moving average (window size \num{1024}).
    Vertical lines indicate year boundaries.
    }
    \label{fig:full_training_run_losses}
\end{figure}

Figure~\ref{fig:full_training_run_losses} shows the UniMaia training loss over the full run. The loss decreases rapidly early in training before gradually plateauing, while retaining noticeable shard-dependent variance. The smoothed loss also exhibits clear year-dependent structure, consistent with the reproducibility effects discussed in Appendix~\ref{app:reproducibility}.

\begin{figure}[h!]
    \centering
    \includegraphics[width=\linewidth]{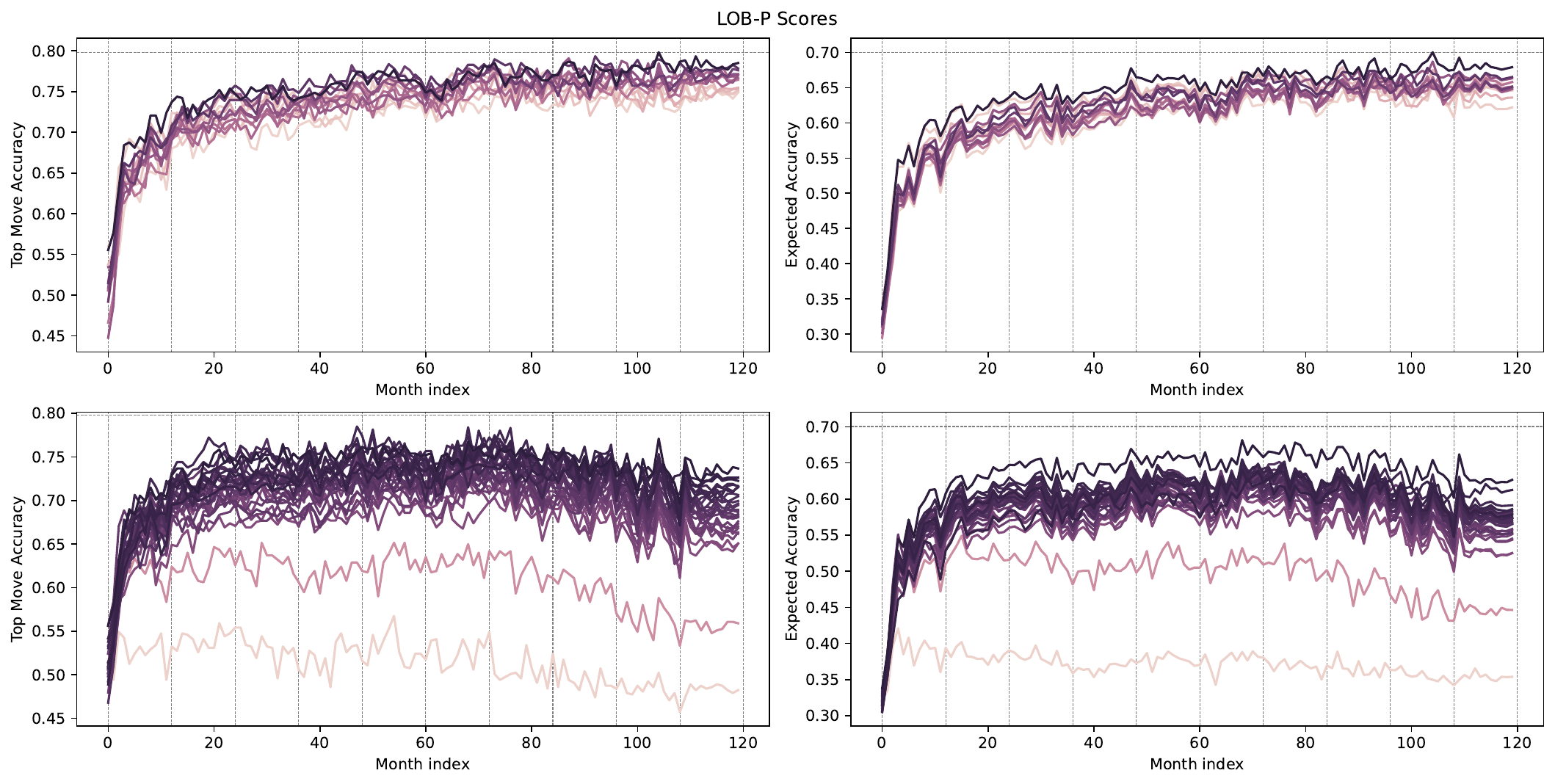}
    \caption[LOB-P training dynamics grouped by template performance.]{
    LOB-P templates partitioned by final top-move accuracy using a threshold of 0.74.
    Top: high-performing templates.
    Bottom: low-performing templates.
    Left: top-move accuracy.
    Right: expected accuracy.
    }
    \label{fig:full_training_run_lob_p_template_split}
\end{figure}

To further analyze the divergent LOB-P and LOB-C behavior observed in the main paper, we partition LOB-P templates by final top-move accuracy using a threshold of 0.74. Figure~\ref{fig:full_training_run_lob_p_template_split} shows a clear bifurcation: high-performing templates continue improving throughout training, whereas lower-performing templates gradually degrade.

This split largely aligns with whether the prompt contains an explicit opening PGN. High-performing templates primarily use semantic opening descriptions without move sequences, whereas lower-performing templates typically include partial PGN conditioning. The weakest-performing templates effectively reduce to unconstrained move prediction conditioned only on partial move history.

Table~\ref{tab:full_training_run_training_dynamics} reports benchmark scores at December checkpoints throughout training. Most benchmarks improve steadily over time, with diminishing gains in later stages. Despite year- and month-dependent fluctuations in the training loss, benchmark performance remains relatively stable overall, with all benchmarks except LOB-P exhibiting clear upward trends. Transient loss increases tend to affect expected accuracy more than top-move accuracy, although the effect is modest.

\begin{table*}[h!]
\centering
\small
\begin{tabular}{lrrrrrr}
\toprule
Month & LOB-P & LOB-C & LIF-D & LIF & M1-S & Cumulative Positions \\
\midrule
December 2013 & \num{0.6695} & \num{0.7079} & \num{0.4557} & \num{0.5187} & \num{0.5363} & \num{786432} \\
December 2014 & \num{0.7187} & \num{0.7641} & \num{0.4654} & \num{0.5374} & \num{0.5435} & \num{1572864} \\
December 2015 & \num{0.7157} & \num{0.7672} & \num{0.4680} & \num{0.5394} & \num{0.5448} & \num{2359296} \\
December 2016 & \num{0.7386} & \num{0.7934} & \num{0.4680} & \num{0.5439} & \num{0.5496} & \num{3997696} \\
December 2017 & \num{0.7274} & \num{0.7912} & \num{0.4674} & \num{0.5477} & \num{0.5477} & \num{6553600} \\
December 2018 & \num{0.7390} & \num{0.8042} & \num{0.4658} & \num{0.5508} & \num{0.5522} & \num{11141120} \\
December 2019 & \num{0.7313} & \num{0.8036} & \num{0.4683} & \num{0.5532} & \num{0.5528} & \num{18415616} \\
December 2020 & \num{0.7351} & \num{0.8185} & \num{0.4692} & \num{0.5567} & \num{0.5526} & \num{31588352} \\
December 2021 & \num{0.6966} & \num{0.8046} & \num{0.4697} & \num{0.5575} & \num{0.5520} & \num{49479680} \\
December 2022 & \num{0.7094} & \num{0.8162} & \num{0.4748} & \num{0.5587} & \num{0.5534} & \num{66977792} \\
\bottomrule
\end{tabular}
\caption{Top move accuracy at December checkpoints during UniMaia training, along with the cumulative number of positions processed.}
\label{tab:full_training_run_training_dynamics}
\end{table*}

\subsection{UniMaia-Aux}
\label{app:training_dynamics_unimaia_aux}

\begin{figure}[h!]
    \centering
    \includegraphics[width=\linewidth]{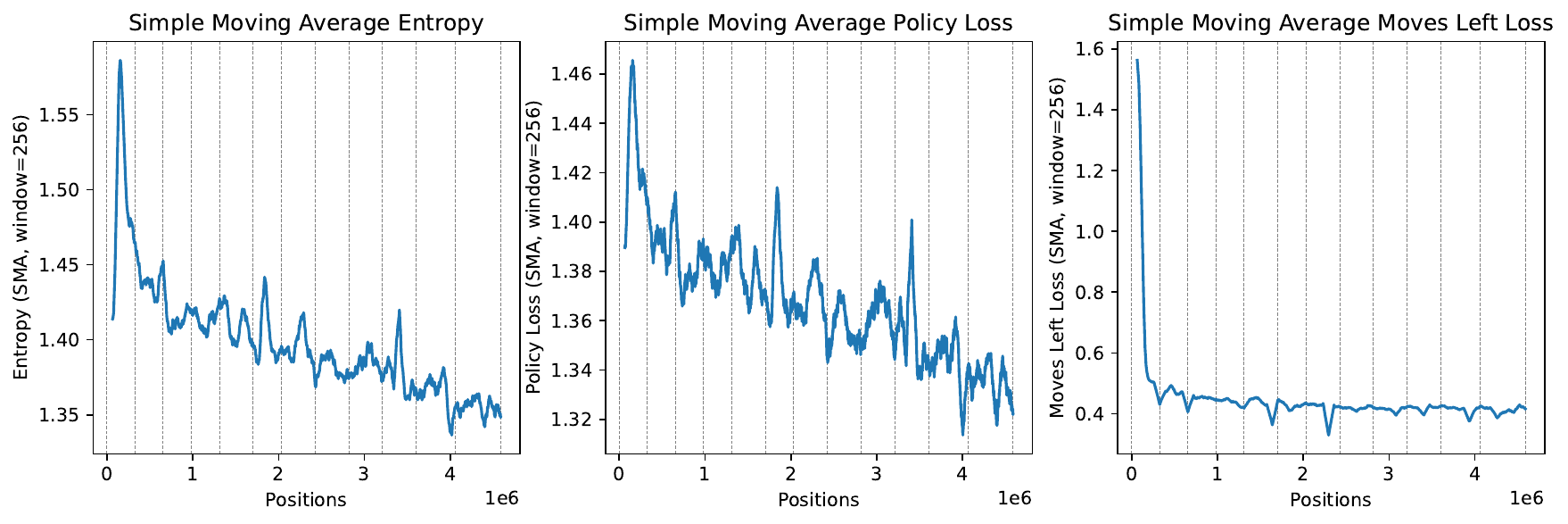}
    \caption[Optimization dynamics during UniMaia-Aux fine-tuning.]{
    Left: policy entropy.
    Middle: policy loss.
    Right: moves-left loss.
    All curves show simple moving averages.
    }
    \label{fig:full_training_run_aux_targets_losses_entropy_policy_moves_left}
\end{figure}

Figure~\ref{fig:full_training_run_aux_targets_losses_entropy_policy_moves_left} shows optimization dynamics during auxiliary-target fine-tuning. The policy loss initially decreases before briefly increasing alongside a sharp rise in policy entropy, suggesting escape from a local policy minimum. The loss then decreases again, while the moves-left loss improves steadily throughout training.

\begin{figure}[h!]
    \centering
    \includegraphics[width=\linewidth]{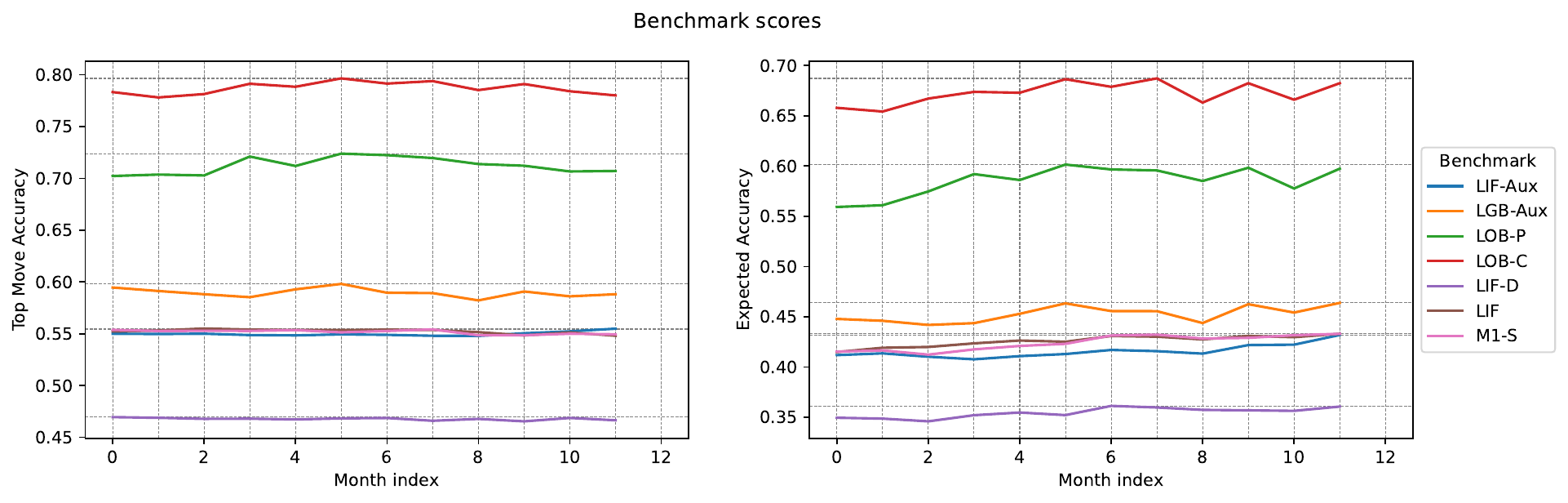}
    \caption[Benchmark trajectories during UniMaia-Aux fine-tuning.]{
    Top-move accuracy and expected accuracy across monthly checkpoints.
    Vertical lines indicate year boundaries.
    }
    \label{fig:full_training_run_aux_targets_benchmark_scores}
\end{figure}

\begin{figure}[h!]
    \centering
    \includegraphics[width=\linewidth]{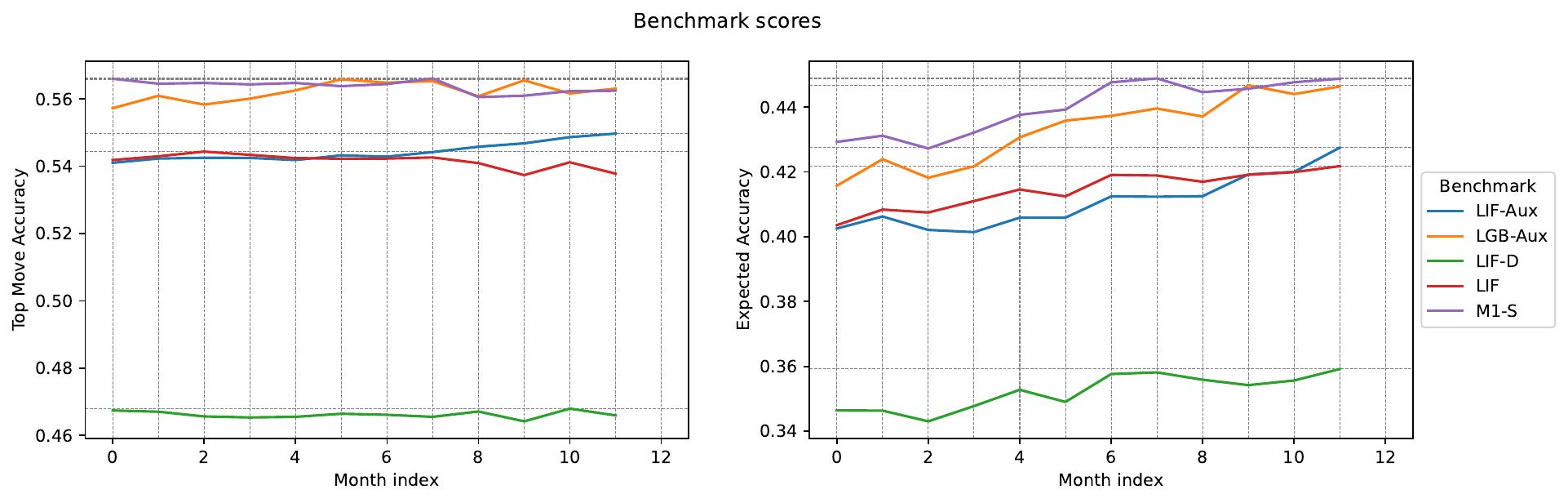}
    \caption[Benchmark trajectories during UniMaia-Aux fine-tuning with the first 10 plies omitted.]{
    Top-move accuracy and expected accuracy across monthly checkpoints with opening plies excluded from evaluation.
    }
    \label{fig:full_training_run_aux_targets_benchmark_scores_skip_first_10}
\end{figure}

\begin{figure}[h!]
    \centering
    \includegraphics[width=\linewidth]{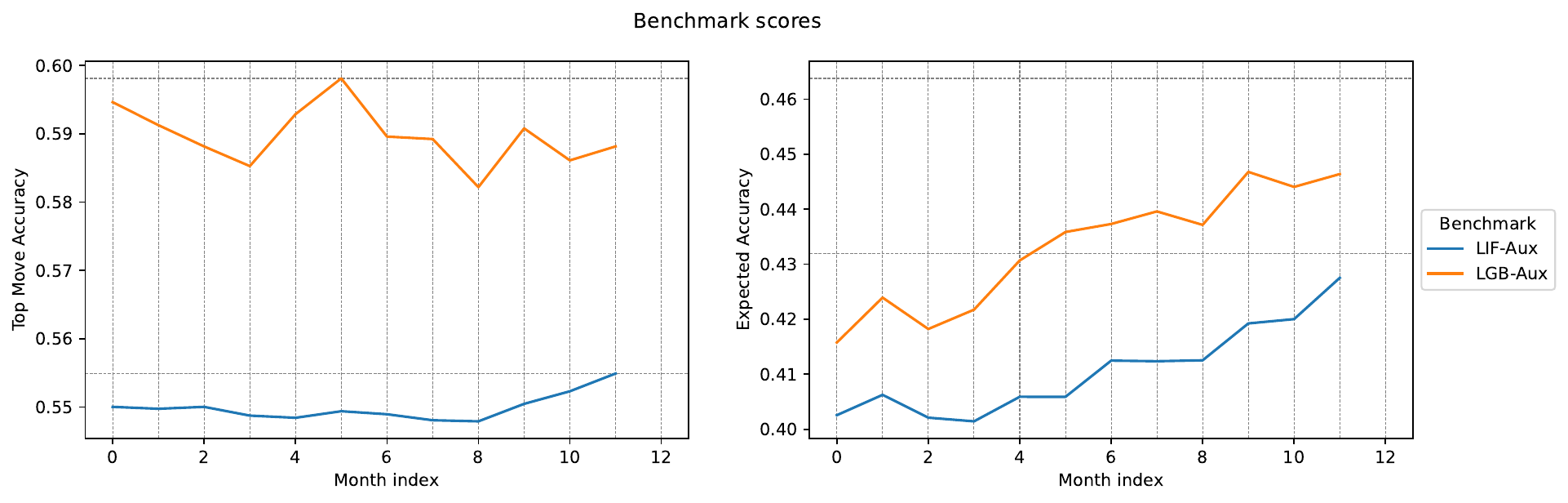}
    \caption[Training dynamics on temporally augmented benchmarks.]{
    Top-move accuracy and expected accuracy for LIF-Aux and LGB-Aux during fine-tuning.
    }
    \label{fig:full_training_run_aux_targets_benchmark_scores_aux}
\end{figure}

\begin{figure}[h!]
    \centering
    \includegraphics[width=\linewidth]{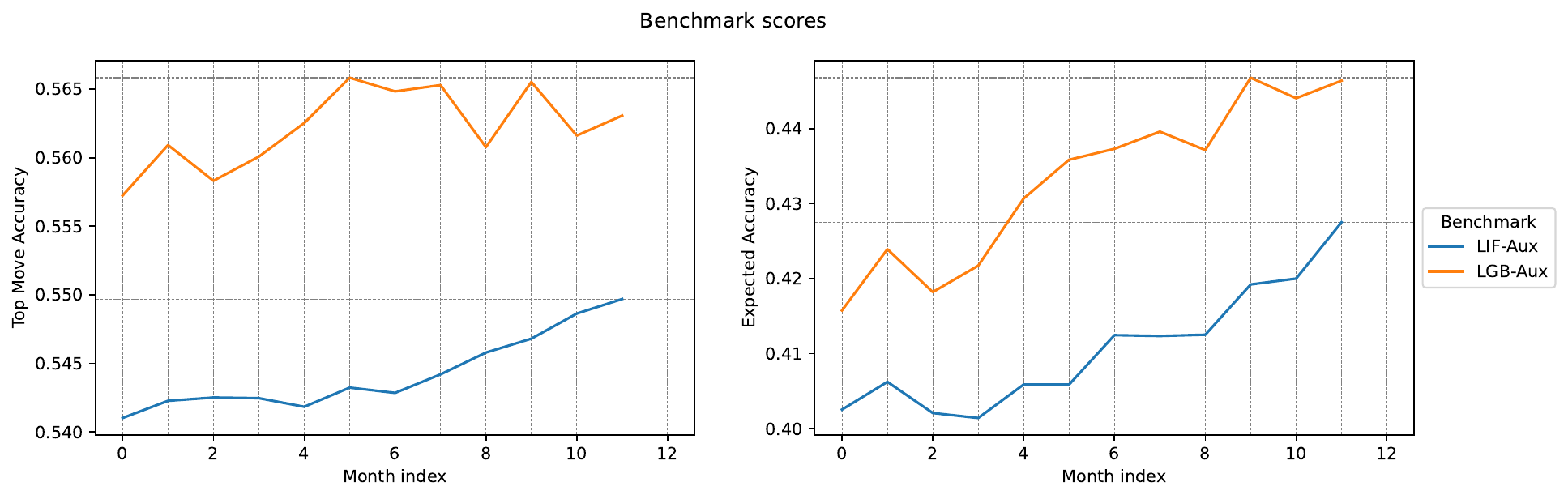}
    \caption[Training dynamics on temporally augmented benchmarks with the first 10 plies omitted.]{
    Top-move accuracy and expected accuracy for LIF-Aux and LGB-Aux with opening plies excluded.
    }
    \label{fig:full_training_run_aux_targets_benchmark_scores_skip_first_10_aux}
\end{figure}

Figures~\ref{fig:full_training_run_aux_targets_benchmark_scores} and~\ref{fig:full_training_run_aux_targets_benchmark_scores_skip_first_10} show benchmark trajectories on the original benchmarks during fine-tuning, while Figures~\ref{fig:full_training_run_aux_targets_benchmark_scores_aux} and~\ref{fig:full_training_run_aux_targets_benchmark_scores_skip_first_10_aux} report results on temporally augmented benchmarks.

Unlike UniMaia, UniMaia-Aux often continues improving expected accuracy even when top-move accuracy plateaus, suggesting that auxiliary supervision primarily improves policy calibration rather than only top predictions.

Performance on temporally augmented benchmarks generally improves throughout fine-tuning, especially when the first 10 plies are omitted. In contrast, performance on the original benchmarks slightly declines later in training, suggesting partial overfitting to the auxiliary supervision setup or prompt format.

The auxiliary suffix may also weaken fixed prompt--opening associations during the opening phase. Because prompts now include temporal metadata and move-delay information, identical openings no longer necessarily share identical prompts, encouraging reliance on broader contextual signals rather than static prompt patterns.

%% file: appendices/appendix-experimental-setup.tex
\section{Additional Experimental Details}
\label{app:experimental_details}

This appendix provides additional details on benchmark construction, evaluation protocols, baselines, and prompting strategies omitted from the main paper for brevity.

\subsection{Benchmark Details}
\label{app:benchmark_details}

\paragraph{Lichess Openings Benchmark (LOB).}
LOB evaluates opening controllability using prompts derived from curated opening databases containing ECO codes, opening names, and canonical move sequences. Positions are paired with prompts sampled from \textsc{LichessTemplates-Instruct}.

We evaluate two variants:
\begin{itemize}
    \item \textbf{LOB--Partial (LOB--P):} Prompts contain only the opening name and/or moves up to the current position.
    \item \textbf{LOB--Canonical (LOB--C):} Prompts additionally include the canonical continuation of the opening.
\end{itemize}

LOB--P avoids explicit answer leakage and therefore provides a more challenging and realistic evaluation, whereas LOB--C measures upper-bound opening-following performance.

\paragraph{Lichess Instruction-Following Benchmark (LIF).}
LIF is constructed from 2023 \textsc{LichessGames} positions formatted using templates from \textsc{LichessTemplates-Pretrain} and \textsc{LichessTemplates-Instruct}. The benchmark spans diverse player ratings, openings, and gameplay contexts, and serves as the primary prompt-conditioned evaluation benchmark.

\paragraph{Lichess Instruction-Following Diagnostic (LIF-D).}
LIF-D (19{,}606 positions) is a controlled diagnostic benchmark constructed from January 2013 positions restricted to the 900--1100 Elo range. Because it is drawn from the same temporal distribution as the training data, it is used only for ablations and training diagnostics.

\paragraph{Lichess Instruction-Following Top-10 Benchmark (LIF-T10).}
LIF-T10 (25{,}000 positions) is constructed from January 2023 games in which at least one player belongs to the most active Lichess accounts. The benchmark evaluates whether prompt-conditioned models can exploit player-specific information when identities are included in prompts.

\paragraph{Lumbra's Gigabase Benchmark (LGB).}
LGB is constructed from Lumbra's Gigabase~\cite{lumbrasgigabaseFreeChess}, a curated collection of deduplicated PGN files aggregated from multiple online and over-the-board (OTB) sources. We evaluate on the OTB 2020--2024 subset to measure generalization beyond the Lichess training distribution.

Games are formatted using \textsc{LichessTemplates-Instruct}. To partially account for systematic rating differences between FIDE and Lichess ratings, all player ratings are shifted by +500 Elo during evaluation.

\paragraph{Temporally augmented benchmarks.}
LIF-Aux and LGB-Aux append temporal metadata including time control, remaining clock time, and previous-move delay when available. These benchmarks additionally evaluate the auxiliary targets introduced during auxiliary-target fine-tuning.

\subsection{Metadata-Conditioned Benchmarks}
\label{app:metadata_conditioned_benchmarks}

\paragraph{Maia-1 (M1).}
We evaluate using the publicly released ``Maia KDD Testing Set'' from McIlroy-Young et al.~\cite{mcilroy2020aligning_maiachess}. The released dataset differs slightly from the filtering protocol described in the original paper, including a broader Elo range and a small number of low-time positions. We use the released version for reproducibility.

\paragraph{Maia-1 Subset (M1-S).}
M1-S is a stratified subset of M1 grouped by player Elo (rounded to the nearest hundred) and time control. It is used for rapid evaluation during ablations and for evaluating prompt-based baselines when full M1 evaluation is computationally expensive.

\paragraph{Maia-2 Rapid Benchmark (M2R).}
M2R follows the Maia-2 evaluation protocol introduced by Tang et al.~\cite{tang2024maia2} using Lichess Rapid games from December 2019.

\paragraph{Allie Blitz Benchmark (ABB).}
ABB consists of Lichess Blitz games from 2022~\cite{zhang2024humanallie}. Following prior work, low-time positions are filtered to reduce noise.

\subsection{Evaluation Protocol}
\label{app:evaluation_protocol}

Prior human-move-prediction benchmarks commonly exclude the first 10 plies and moves made under severe time pressure to reduce noise from opening preparation and time scrambles~\cite{mcilroy2020aligning_maiachess,zhang2024humanallie}.

Where applicable, we report both filtered and unfiltered results. For prompt-conditioned benchmarks such as LIF and LGB, we retain opening moves because some prompts explicitly target opening behavior.

We report:
\begin{itemize}
    \item \textbf{Top-move accuracy (Acc@1):} Whether the highest-probability move matches the human move.
    \item \textbf{Expected accuracy (\(\mathbb{E}[\mathrm{Acc}]\)):} The probability assigned to the human move.
    \item \textbf{Weighted F1 (\(\mathrm{F1}_{w}\)):} Used for auxiliary categorical targets.
    \item \textbf{MAE:} Used for scalar auxiliary targets.
\end{itemize}

As in the main paper, expected accuracy is reported only for models that produce normalized move distributions.

\subsection{Baseline Model Details}
\label{app:baseline_details}

\paragraph{Lc0-CF baselines.}
We evaluate multiple pretrained Lc0-CF policy networks, including the t3, t82, BT3, BT4, and BT4-spsa checkpoints~\cite{monroe2024mastering,monroe2026chessformer}. These models differ in parameter count and architectural details but all use structured board encodings based on 112 spatial feature planes.

\paragraph{Maia family.}
Maia-1 consists of separate Lc0-based models trained for different Elo ranges~\cite{mcilroy2020aligning_maiachess}. Maia-2 introduces discrete Elo tokens and separate rapid/blitz models~\cite{tang2024maia2}, while Maia-3 replaces discrete tokens with interpolated Elo embeddings using a ChessFormer backbone~\cite{monroe2026chessformer}.

\paragraph{Allie.}
Allie is a GPT-2-based human imitation model conditioned on Elo and time-control metadata encoded as tokens~\cite{zhang2024humanallie}. We evaluate the \textsc{Allie-Policy} configuration without search.

\paragraph{ChessCLIP.}
ChessCLIP scores the compatibility between prompts, candidate moves, and board states using CLIP-style contrastive representations~\cite{feng2024chessgpt}. During inference, all legal moves are enumerated and ranked according to prompt similarity.

\paragraph{ChessGPT v1.} ChessGPT is a large language model trained for chess reasoning and move prediction~\cite{feng2024chessgpt}. It is based on the RedPajama INCITE 3B model~\cite{weber2024redpajama}. ChessGPT-base is obtained by continual pretraining on a mixture of text, PGN, and FEN data, while ChessGPT-chat is further fine-tuned on conversational chess data.

\textbf{ChessGPT-base.} We evaluate ChessGPT-base using two prompting strategies unless otherwise specified:
\begin{enumerate}
    \item \textbf{Prompt}: The prompt followed by the PGN move list without headers: \texttt{\{prompt\}\textbackslash{}n\textbackslash{}n\{pgn\}}.
    \item \textbf{Elo and time control PGN header\footnote{We use \textit{PGN header} to refer to PGN-style headers that may not fully conform to the PGN specification.}}: A PGN header describing both players' Elos and the time control (Appendix~\ref{app:pgn_header_prompt_template}), followed by two newlines and the PGN move list.
\end{enumerate}

To ensure reliable move extraction, we use \texttt{outlines}~\cite{willard2023efficientoutlines} to constrain the generated output with a predefined regular expression. For ChessGPT-base, the move number may optionally be preceded by a space, and both the move number and trailing punctuation are optional.

\textbf{ChessGPT-chat.} We evaluate ChessGPT-chat using three prompting strategies unless otherwise specified:
\begin{enumerate}
    \item \textbf{Simple, duplicate}: We prompt ChessGPT-chat with \texttt{\{prompt\}\textbackslash{}n\textbackslash{}nPGN: \{pgn\}\textbackslash{}n\textbackslash{}nJust write your move.} Applying the chat template twice---once without the generation prompt and once with it---improves performance on the Lichess openings benchmark.
    \item \textbf{Simple, original}: The same prompt as above, except the chat template is applied only once with the generation prompt.
    \item \textbf{General policy}: A prompt adapted from Feng et al., consisting of a prompt to play chess in a given position, a PGN header with certain fields masked by ``??'', the PGN move list, and a final instruction to produce the next move.
\end{enumerate}

As with ChessGPT-base, we use \texttt{outlines}~\cite{willard2023efficientoutlines} to constrain generation. For ChessGPT-chat, the output must match a move number, one or three periods, a space, and a valid SAN move.

\vspace{0.5em}

\paragraph{GPT-3.5 Turbo Instruct.} GPT-3.5 Turbo Instruct~\cite{openaiGPT35Turbo} is a proprietary language model that demonstrates strong chess performance when prompted appropriately.

\textit{Prompting strategies.} We evaluate the following configurations:
\begin{enumerate}
    \item \textbf{GM PGN header}: A PGN header indicating a game between strong players (\textit{i.e.}, Magnus Carlsen and Garry Kasparov), followed by the move list. We use the prompt from the open-sourced implementations accompanying Carlini~\cite{carlini2023PlayingChess} and Allie~\cite{zhang2024humanallie}, which differs slightly from the version described in the Carlini blog and reproduced in the Allie appendix (Appendix~\ref{app:gm_pgn_header_prompt}).
    \item \textbf{GM PGN header + prompt}: A natural-language prompt inserted between the header and the move list.
    \item \textbf{Original PGN header}: The original PGN header, followed by the move list.
    \item \textbf{Elo and time control PGN header}: A PGN header specifying player ratings and time control (Appendix~\ref{app:pgn_header_prompt_template}), followed by the move list.
\end{enumerate}

For autoregressive language models, generated moves are constrained to valid chess formats where possible. Illegal moves are replaced with uniformly sampled legal moves during evaluation.

\subsection{Model Comparison}
\label{app:model_comparison}

The evaluated models differ substantially in their representations and conditioning mechanisms. Engine-based models such as Lc0 rely exclusively on structured board encodings, while metadata-conditioned models incorporate explicit Elo and time-control signals through tokens or embeddings. Prompt-conditioned language models instead encode all contextual information implicitly through text.

UniMaia combines structured Lc0 board representations with natural-language conditioning, enabling semantically flexible control while preserving domain-specific inductive biases. UniMaia-Aux further augments the model with auxiliary prediction heads for behavioral and temporal targets.

Tables~\ref{tab:model_inputs}--\ref{tab:model_training_scale_language} summarize the input representations, predicted outputs, and training configurations of the evaluated models.

\begin{table*}[h!]
\centering
\small
\small
\begin{tabular}{l l l l c}
\toprule
{Model} & {Game state} & {Elo} & {Time control} & {Prompt} \\
\midrule
Lc0 & 112 Lc0 planes & -- & -- & -- \\
Maia-1 & 112 Lc0 planes & Separate models & -- & -- \\
Maia-2 & Current board & Discrete tokens & Separate models & -- \\
Maia-3 & 112 Lc0 planes & Interp. embeddings & -- & -- \\
Allie & Move history & Interp. embeddings & Discrete tokens & -- \\
ChessCLIP & 112 Lc0 planes & Via prompt & Via prompt & \checkmark \\
ChessGPT & Via prompt & Via prompt & Via prompt & \checkmark \\
GPT-3.5 & Via prompt & Via prompt & Via prompt & \checkmark \\
UniMaia & 112 Lc0 planes & Via prompt & Via prompt & \checkmark \\
UniMaia-Aux & 112 Lc0 planes & Via prompt & Via prompt & \checkmark \\
\bottomrule
\end{tabular}
\caption[Comparison of model inputs.]{Comparison of model inputs and conditioning mechanisms.}
\label{tab:model_inputs}
\end{table*}

\begin{table*}[h!]
\centering
\small
\begin{tabular}{l c c c c c c c}
\toprule
{Model} & {Policy} & {Value} & {Moves left} & {Move delay} & {Resignation} & {Termination} & {Text} \\
\midrule
Lc0 & \checkmark & \checkmark & \checkmark & -- & -- & -- & -- \\
Maia-1 & \checkmark & \checkmark & -- & -- & -- & -- & -- \\
Maia-2 & \checkmark & \checkmark & -- & -- & -- & -- & -- \\
Maia-3 & \checkmark & \checkmark & -- & -- & -- & -- & -- \\
Allie & \checkmark & \checkmark & -- & \checkmark & \checkmark & -- & -- \\
ChessCLIP & \checkmark & -- & -- & -- & -- & -- & -- \\
ChessGPT & \checkmark & -- & -- & -- & -- & -- & \checkmark \\
GPT-3.5 & \checkmark & -- & -- & -- & -- & -- & \checkmark \\
UniMaia & \checkmark & -- & -- & -- & -- & -- & -- \\
UniMaia-Aux & \checkmark & \checkmark & \checkmark & \checkmark & \checkmark & \checkmark & -- \\
\bottomrule
\end{tabular}
\caption[Comparison of model outputs.]{Comparison of predicted outputs and auxiliary targets.}
\label{tab:model_outputs}
\end{table*}

\begin{table*}[h!]
\centering
\small
\begin{tabular}{l r r r r r}
\toprule
{Model} & {Batch size} & {Grad. acc.} & {\# GPUs} & {\# Positions} & {\# Parameters} \\
\midrule
Maia-1 & \num{1024} & -- & 1 & 409.6M & 92M \\
Maia-2 & \num{8192} & -- & 2 & 9.15B & 23M \\
Maia-3 & \num{128} & 4 & 8 & 1.024B & 5M, 23M, 79M \\
ChessCLIP & 512 & -- & 8 & 1.3M & 67M \\
UniMaia & \num{64} & -- & 2 & 66.978M & 3B+404M \\
UniMaia-Aux & \num{64} & -- & 2 & 71.565M & 3B+515M \\
\bottomrule
\end{tabular}
\caption[Training configurations for structured-input models.]{Training configurations for structured-input models.}
\label{tab:model_training_scale_structured}
\end{table*}

\begin{table*}[h!]
\centering
\small
\begin{tabular}{l r r r r}
\toprule
{Model} & {\# GPUs} & {Context} & {\# Tokens} & {\# Parameters} \\
\midrule
Allie & 8 & 512 & 262.144B & 355M \\
ChessGPT-base & 8 & 1024 & ? & 3B \\
ChessGPT-chat & 8 & 1024 & ? & 3B \\
\bottomrule
\end{tabular}
\caption[Training configurations for language-based models.]{Training configurations for language-based models. ChessGPT is trained for one epoch. The exact dataset size is unavailable due to dataset release restrictions.}
\label{tab:model_training_scale_language}
\end{table*}

%% file: appendices/appendix-prompt-templates.tex
\section{Prompt Templates}
\label{app:prompt_templates}

\subsection{Example \textsc{LichessTemplates} Prompt Templates}
\label{app:example_lichess_prompt_templates}

Table~\ref{tab:example_lichess_templates} shows representative prompt templates from \textsc{LichessTemplates-Pretrain} and \textsc{LichessTemplates-Instruct}, illustrating the linguistic variability and metadata coverage used during training and evaluation.

\begin{table*}[h!]
\centering
\small
\begin{tabular}{p{0.1\linewidth} p{0.8\linewidth}}
\toprule
\textbf{Template Set} & \textbf{Example Template} \\
\midrule

\multirow{3}{*}{\textsc{Pretrain}}
& \texttt{This \{time\_control\_lower\} chess game was played online at Lichess.org on \{date\}, with each player given \{td\_all\_seconds\} and \{text\_time\_increment\} per move. The game featured \{white\_title\_full\_quot\_name\} \{white\_alias\_alias\_paren\} as \{W\_lit\}, with an Elo rating of \{white\_elo\}, against \{black\_title\_full\_quot\_name\} \{black\_alias\_alias\_paren\} as \{B\_lit\}, who had a rating of \{black\_elo\}. The opening used was the \{opening\_commas\}, delineated by the moves: \{opening\_moves\}.} \\[0.75em]

\cmidrule(lr){2-2}

& \texttt{In a \{ratedness\_lower\} \{time\_control\_lower\} chess game hosted on Lichess.org on \{date\}, \{white\_title\_full\_paren\_title\} \{white\_alias\_known\_as\_quot\_name\}, rated \{white\_elo\}, played as \{W\_lit\} against \{black\_title\_full\_paren\_title\} \{black\_alias\_known\_as\_quot\_name\} and rated \{black\_elo\}. The game opened with the \{opening\}, starting with moves \{opening\_moves\}. The time control for the match was \{td\_all\_seconds\} \{with\_text\_time\_increment\} per move.} \\[0.75em]

\cmidrule(lr){2-2}

& \texttt{On \{date\}, at Lichess.org, \{white\_title\_full\} \{white\_alias\} (white) with an Elo rating of \{white\_elo\} played against \{black\_title\_full\} \{black\_alias\} (black) who had an Elo rating of \{black\_elo\}. This rated \{time\_control\_lower\} game had a time control of \{text\_td\} per player \{time\_increment\_words\_without\_any\_increment\}. The game featured the \{opening\}, which began with the moves \{opening\_moves\}.} \\
\midrule

\multirow{3}{*}{\textsc{Instruct}}
& \texttt{Join \{black\_title\} \{black\_alias\}, ELO \{black\_elo\}, as he plays \{B\_lit\} utilizing the \{opening\} with the opening moves \{opening\_moves\}. Challenge this \{black\_rank\_player\_a\} to a game on Lichess and test your skills against his opening strategy.} \\[0.75em]

\cmidrule(lr){2-2}

& \texttt{Play as \{white\_title\_full\} \{white\_alias\}, taking the \{w\_lit\_lower\_pieces\} and opening the game with \{opening\_moves\}, known as the \{opening\_variant\_name\_of\_the\_main\}.} \\[0.75em]

\cmidrule(lr){2-2}

& \texttt{In this chess game on Lichess, the player, controlling the \{b\_lit\_lower\_pieces\}, opens with the \{opening\} by playing \{opening\_moves\}.} \\
\bottomrule
\end{tabular}
\caption{Representative prompt templates from \textsc{LichessTemplates-Pretrain} and \textsc{LichessTemplates-Instruct}.}
\label{tab:example_lichess_templates}
\end{table*}

\subsection{Metadata-Conditioned Benchmark Prompts}
\label{app:metadata-conditioned_benchmark_prompts}

All prompt-conditioned models were provided with the following prompt template (formatted with game metadata) when evaluated on the metadata-conditioned benchmarks:

``\texttt{An anonymous white player with an ELO of \{white\_elo\} plays chess against another anonymous black player, with an ELO of \{black\_elo\} in a \{type.lower()\} game on Lichess played with \{time\_info\}.}''

Here, ``\texttt{time\_control}'' is rendered as ``\texttt{a time control of <duration as human-friendly text> with (no increment|an increment of <increment as human-friendly text>)}''

\subsection{GM PGN Header Prompt}
\label{app:gm_pgn_header_prompt}

The GM PGN header prompt used for evaluation is based on the open-sourced implementations accompanying Carlini~\cite{carlini2023PlayingChess} and Allie~\cite{zhang2024humanallie}. This version differs slightly from the prompt described in the Carlini blog and reproduced in the Allie appendix.

In particular, the blog version includes a fixed game result and lists Kasparov as White and Carlsen as Black, whereas the implementation version omits the result field and lists Carlsen as White and Kasparov as Black. We use the implementation version for all experiments.

\begin{verbatim}
[White "Magnus Carlsen"]
[Black "Garry Kasparov"]
[WhiteElo "2900"]
[BlackElo "2800"]

\end{verbatim}

\subsection{Elo and Time Control PGN Header}
\label{app:pgn_header_prompt_template}

Aside from the opening-specific benchmarks, ChessGPT-base and GPT-3.5 Turbo Instruct were prompted using a PGN header prepended to the movelist:

\begin{minipage}[t]{0.5\linewidth}
\begin{verbatim}"""[White "???"]
[Black "???"]
[WhiteElo "\{white\_elo\}"]
[BlackElo "\{black\_elo\}"]
[TimeControl "\{time\_control\}"]

"""\end{verbatim}
\end{minipage}

This format mirrors the structured metadata representation used during training of these models, in contrast to the natural-language prompts used for UniMaia.

\subsection{Auxiliary Targets Prompt Suffix}
\label{app:auxiliary_targets_prompt_suffix}

When predicting the auxiliary targets, we append the following suffix to the prompt if the time control is available:

\texttt{ The time control for this \{time\_control.lower()\} chess game was \{text\_td\} \{time\_increment\_text\}.}

where \texttt{text\_td} is the time delay in hours, minutes, and seconds, and \texttt{time\_increment\_text} is either \texttt{with no additional increment} or \texttt{with \{time\_increment\} <second or seconds> of increment}, with the correct pluralization.

If the game has \texttt{\%clk} annotations for move delay prediction, we append the following:

\texttt{ Time taken last turn by active player: \{active\_player\_last\_move\_time\}. Time taken last turn by opponent: \{opponent\_last\_move\_time\}. Time remaining for the active player: \{active\_player\_time\_remaining\}. Time remaining for the opponent: \{opponent\_time\_remaining\}.}

%% file: appendices/appendix-detailed-evaluation-results.tex
\section{Detailed Evaluation Results}
\label{app:detailed_evaluation_results}

This appendix provides the complete benchmark results corresponding to the summarized results in Section~\ref{sec:results}. We include additional prompt-conditioned and metadata-conditioned evaluations, results with the first 10 plies omitted, auxiliary-target evaluations, and comparisons across model classes under a unified evaluation harness.

Because prior work often uses different benchmark suites, prompting strategies, and legality handling procedures, direct comparison across model classes is difficult. To improve comparability, all reproduced results are evaluated using a unified benchmark harness that standardizes preprocessing, prompting, legality handling, and evaluation metrics whenever possible.

\subsection{Prompt-Conditioned Benchmarks}
\label{app:evaluation_prompt_conditioned}

Tables~\ref{tab:prompt-conditioned_top_move_accuracy_model_results}--\ref{tab:prompt-conditioned_expected_accuracy_skip_first_10_model_results}
report the full prompt-conditioned benchmark results for both \textbf{Acc@1} and expected accuracy.

\begin{table*}[h!]
\centering
\small
\begin{tabular}{l c c c c c c}
\toprule
{Model} & {LOB-P} & {LOB-C} & {LIF-D} & {LIF} & {LIF-T10} & {LGB} \\
\midrule
Lc0-CF (t3) & 0.5010 & 0.5027 & 0.3244 & 0.3942 & 0.3180 & 0.4772 \\
Lc0-CF (t82) & 0.5011 & 0.5024 & 0.3214 & 0.3908 & 0.3148 & 0.4749 \\
Lc0-CF (BT3) & 0.5025 & 0.5041 & 0.3252 & 0.3924 & 0.3162 & 0.4771 \\
Lc0-CF (BT4) & 0.5025 & 0.5039 & 0.3246 & 0.3924 & 0.3132 & 0.4770 \\
Lc0-CF (BT4-spsa) & 0.4744 & 0.4757 & 0.3217 & 0.3902 & 0.3122 & 0.4643 \\
Maia-1\textsuperscript{*} & 0.5582 & 0.5602 & 0.4411 & 0.4797 & 0.4076 & 0.4857 \\
Maia-1\textsuperscript{**} & 0.5709 & 0.5729 & 0.4411 & 0.4830 & 0.4085 & 0.5034 \\
Maia-2 (Blitz) & 0.4058 & 0.4040 & 0.4144 & 0.4763 & 0.4051 & 0.4452 \\
Maia-2 (Rapid) & 0.4129 & 0.4123 & 0.4207 & 0.4758 & 0.3943 & 0.4484 \\
Allie-Policy & 0.6150 & 0.6164 & 0.4658 & 0.5390 & 0.4662 & 0.5572 \\
\midrule
ChessCLIP & 0.2730 & 0.2696 & 0.0895 & 0.0918 & 0.0747 & 0.1040 \\
ChessGPT-base\textsuperscript{\dag} & \bftab{0.7998} & \bftab{0.8955} & 0.3808 & 0.4464 & 0.4153 & 0.4788 \\
ChessGPT-chat\textsuperscript{\ddag} & 0.3417 & 0.2272 & 0.1494 & 0.1637 & 0.1480 & 0.1501 \\
GPT-3.5\textsuperscript{\S} & 0.5748 & 0.6431 & 0.4457 & 0.5312 & 0.4778 & 0.5514 \\
UniMaia & \underline{0.7094} & \underline{0.8162} & \bftab{0.4748} & \bftab{0.5587} & \bftab{0.4954} & \bftab{0.6158} \\
UniMaia-Aux & 0.7071 & 0.7801 & \underline{0.4664} & \underline{0.5481} & \underline{0.4915} & \underline{0.6086} \\
\bottomrule
\end{tabular}
\caption[Prompt-conditioned benchmark results, measured by Acc@1.]{\textbf{Prompt-conditioned} benchmark results, measured by \textbf{Acc@1}. Best values are in bold and second-best values are underlined. \textsuperscript{*}Evaluated only with the official models. \textsuperscript{**}Evaluated with both the official and unofficial models. \textsuperscript{\dag}Opening benchmarks evaluated with the regular prompt. Remaining benchmarks evaluated with the Elo and time control PGN header. \textsuperscript{\ddag}Opening benchmarks evaluated with the duplicate simple prompt. Remaining benchmarks evaluated with the general policy prompt. \textsuperscript{\S}Opening benchmarks evaluated with the GM PGN header. Remaining benchmarks evaluated with the Elo and time control PGN header.}
\label{tab:prompt-conditioned_top_move_accuracy_model_results}
\end{table*}

\begin{table}[h!]
\centering
\small
\begin{tabular}{l c c c c}
\toprule
{Model} & {LIF-D} & {LIF} & {LIF-T10} & {LGB} \\
\midrule
Lc0-CF (t3) & 0.3360 & 0.4052 & 0.3290 & 0.5064 \\
Lc0-CF (t82) & 0.3329 & 0.4013 & 0.3265 & 0.5012 \\
Lc0-CF (BT3) & 0.3364 & 0.4036 & 0.3272 & 0.5054 \\
Lc0-CF (BT4) & 0.3357 & 0.4037 & 0.3232 & 0.5067 \\
Lc0-CF (BT4-spsa) & 0.3366 & 0.4046 & 0.3249 & 0.5021 \\
Maia-1\textsuperscript{*} & 0.4532 & 0.4845 & 0.4106 & 0.4515 \\
Maia-1\textsuperscript{**} & 0.4532 & 0.4871 & 0.4114 & 0.4694 \\
Maia-2 (Blitz) & 0.4412 & 0.5021 & 0.4278 & 0.4850 \\
Maia-2 (Rapid) & 0.4467 & 0.5019 & 0.4187 & 0.4825 \\
Allie-Policy & \bftab{0.4812} & \bftab{0.5492} & 0.4735 & 0.5452 \\
\midrule
ChessCLIP & 0.0862 & 0.0813 & 0.0633 & 0.0567 \\
ChessGPT-base\textsuperscript{\dag} & 0.3813 & 0.4440 & 0.4158 & 0.4339 \\
ChessGPT-chat\textsuperscript{\ddag} & 0.1490 & 0.1557 & 0.1491 & 0.1564 \\
GPT-3.5\textsuperscript{\dag} & 0.4608 & 0.5444 & \bftab{0.4914} & 0.5569 \\
UniMaia & \underline{0.4712} & \underline{0.5457} & \underline{0.4792} & \underline{0.5638} \\
UniMaia-Aux & 0.4657 & 0.5378 & 0.4787 & \bftab{0.5692} \\
\bottomrule
\end{tabular}
\caption[Prompt-conditioned benchmark results with the first 10 plies omitted, measured by Acc@1.]{Prompt-conditioned benchmark results, measured by \textbf{Acc@1}. We \textbf{omit} the first 10 plies from each game in the benchmark. Best values are in bold and second-best values are underlined. \textsuperscript{*}Evaluated only with the official models. \textsuperscript{**}Evaluated with both the official and unofficial models. \textsuperscript{\dag}Evaluated with the Elo and time control PGN header. \textsuperscript{\ddag}Evaluated with the general policy prompt.}
\label{tab:prompt-conditioned_top_move_accuracy_skip_first_10_model_results}
\end{table}

\begin{table*}[h!]
\centering
\small
\begin{tabular}{l c c c c c c}
\toprule
{Model} & {LOB-P} & {LOB-C} & {LIF-D} & {LIF} & {LIF-T10} & {LGB} \\
\midrule
Lc0-CF (t3) & 0.3628 & 0.3627 & 0.2310 & 0.2742 & 0.2298 & 0.3390 \\
Lc0-CF (t82) & 0.3633 & 0.3630 & 0.2311 & 0.2744 & 0.2293 & 0.3398 \\
Lc0-CF (BT3) & 0.3644 & 0.3642 & 0.2322 & 0.2756 & 0.2308 & 0.3405 \\
Lc0-CF (BT4) & 0.3657 & 0.3655 & 0.2328 & 0.2772 & 0.2311 & 0.3436 \\
Lc0-CF (BT4-spsa) & 0.3815 & 0.3813 & 0.2462 & 0.2948 & 0.2448 & 0.3616 \\
Maia-1\textsuperscript{*} & 0.4130 & 0.4138 & 0.3321 & 0.3653 & 0.3195 & 0.3651 \\
Maia-1\textsuperscript{**} & 0.4299 & 0.4307 & 0.3321 & 0.3678 & 0.3204 & 0.3827 \\
Maia-2 (Blitz) & 0.2990 & 0.2971 & 0.3227 & 0.3670 & 0.3216 & 0.3423 \\
Maia-2 (Rapid) & 0.3056 & 0.3040 & 0.3224 & 0.3650 & 0.3149 & 0.3401 \\
Allie-Policy & 0.4905 & 0.4906 & 0.3510 & 0.4104 & 0.3525 & 0.4480 \\
\midrule
ChessCLIP & 0.0402 & 0.0403 & 0.0658 & 0.0616 & 0.0496 & 0.0405 \\
UniMaia & \underline{0.5899} & \bftab{0.7140} & \underline{0.3522} & \underline{0.4297} & \bftab{0.3678} & \underline{0.4834} \\
UniMaia-Aux & \bftab{0.5973} & \underline{0.6823} & \bftab{0.3606} & \bftab{0.4331} & \underline{0.3600} & \bftab{0.4891} \\
\bottomrule
\end{tabular}
\caption[{Prompt-conditioned results, measured by \(\mathbb{E}[\mathrm{Acc}]\).}]{\textbf{Prompt-conditioned} results, measured by \(\bm{\mathbb{E}[\mathrm{Acc}}]\). Best values are in bold and second-best values are underlined. \textsuperscript{*}Evaluated only with the official models. \textsuperscript{**}Evaluated with both the official and unofficial models.}
\label{tab:prompt-conditioned_expected_accuracy_model_results}
\end{table*}

\begin{table}[h!]
\centering
\small
\begin{tabular}{l c c c c}
\toprule
{Model} & {LIF-D} & {LIF} & {LIF-T10} & {LGB} \\
\midrule
Lc0-CF (t3) & 0.2360 & 0.2763 & 0.2336 & 0.3624 \\
Lc0-CF (t82) & 0.2364 & 0.2766 & 0.2336 & 0.3651 \\
Lc0-CF (BT3) & 0.2372 & 0.2781 & 0.2349 & 0.3657 \\
Lc0-CF (BT4) & 0.2379 & 0.2800 & 0.2352 & 0.3712 \\
Lc0-CF (BT4-spsa) & 0.2530 & 0.2994 & 0.2508 & 0.3937 \\
Maia-1\textsuperscript{*} & 0.3438 & 0.3713 & 0.3248 & 0.3434 \\
Maia-1\textsuperscript{**} & 0.3438 & 0.3733 & 0.3254 & 0.3570 \\
Maia-2 (Blitz) & 0.3463 & 0.3913 & 0.3446 & 0.3842 \\
Maia-2 (Rapid) & 0.3463 & 0.3886 & 0.3375 & 0.3792 \\
Allie-Policy & \bftab{0.3652} & \underline{0.4215} & \bftab{0.3607} & \underline{0.4393} \\
\midrule
ChessCLIP & 0.0705 & 0.0662 & 0.0518 & 0.0422 \\
UniMaia & 0.3444 & 0.4129 & \underline{0.3507} & 0.4252 \\
UniMaia-Aux & \underline{0.3592} & \bftab{0.4218} & 0.3507 & \bftab{0.4521} \\
\bottomrule
\end{tabular}
\caption[{Prompt-conditioned results with the first 10 plies omitted, measured by \(\mathbb{E}[\mathrm{Acc}]\).}]{\textbf{Prompt-conditioned} results, measured by \(\bm{\mathbb{E}[\mathrm{Acc}]}\). We \textbf{omit} the first 10 plies from each game in the benchmark. Best values are in bold and second-best values are underlined. \textsuperscript{*}Evaluated only with the official models. \textsuperscript{**}Evaluated with both the official and unofficial models.}
\label{tab:prompt-conditioned_expected_accuracy_skip_first_10_model_results}
\end{table}

Overall, UniMaia achieves the strongest performance across most prompt-conditioned tasks, attaining the highest Acc@1 on LIF-D, LIF, LIF-T10, and LGB. The main exception is opening reconstruction (LOB-P and LOB-C), where ChessGPT-base performs best and UniMaia consistently ranks second.

When excluding the first 10 plies to reduce the influence of opening memorization, UniMaia retains the strongest overall performance on over-the-board prediction tasks while remaining competitive across the remaining benchmarks.

For expected accuracy, UniMaia and UniMaia-Aux achieve the strongest overall performance across nearly all prompt-conditioned settings. Although UniMaia-Aux slightly underperforms UniMaia in Acc@1 on the original prompt-conditioned benchmarks, it frequently achieves stronger expected accuracy, suggesting improved policy calibration from auxiliary temporal conditioning.

Both ChessGPT variants frequently generate illegal moves. During evaluation, illegal moves are replaced with uniformly sampled legal moves, reducing both Acc@1 and expected accuracy.

Finally, ChessCLIP rarely ranks the correct move as its top prediction despite occasionally assigning moderate probability mass to it. Its output distributions are typically diffuse after the softmax transformation, resulting in comparatively weak expected accuracy.

\subsubsection{Auxiliary Temporal Metadata}
\label{app:auxiliary_targets_results}

Tables~\ref{tab:auxiliary_targets_model_results}
and~\ref{tab:auxiliary_targets_skip_first_10_model_results}
report results for UniMaia-Aux on temporally augmented benchmarks.

\begin{table*}[h!]
\centering
\small
\makebox[\linewidth][c]{%
\begin{tabular}{l l c c c c c c c c}
\toprule
\multirow{2}{*}{Benchmark} & \multirow{2}{*}{Model} & \multirow{2}{*}{Acc@1\(\uparrow\)} & \multirow{2}{*}{\(\mathbb{E}[\mathrm{Acc}]\uparrow\)} & \multicolumn{3}{c}{\(\mathrm{F1}_{w}\uparrow\)} & \multicolumn{3}{c}{MAE\(\downarrow\)} \\
 & & & & {Result} & {Termination} & {Resigned} & {\(N\!-\!i\!-\!1\)} & {\(V\)} & {\(\Delta t_{a}\)} \\
\midrule
\multirow{2}{*}{LIF-Aux} & Allie-Policy & \underline{0.5390} & \underline{0.4104} & -- & -- & \underline{0.6409} & -- & \bftab{0.2176} & \underline{4.8651} \\
 & UniMaia-Aux & \bftab{0.5549} & \bftab{0.4318} & \bftab{0.6718} & \bftab{0.5448} & \bftab{0.7279} & \bftab{21.3624} & \underline{0.3229} & \bftab{4.0252} \\
 \midrule
\multirow{2}{*}{LGB-Aux} & Allie-Policy & \underline{0.5572} & \underline{0.4480} & -- & -- & \underline{0.0933} & -- & \bftab{0.3711} & \underline{127.7715} \\
 & UniMaia-Aux & \bftab{0.5872} & \bftab{0.4637} & \bftab{0.3814} & \bftab{0.6494} & \bftab{0.6643} & \bftab{35.6365} & \underline{0.4092} & \bftab{124.9658} \\
\bottomrule
\end{tabular}%
}
\caption[{Auxiliary targets results.}]{Auxiliary targets results. Best values are in bold and second-best values are underlined.}
\label{tab:auxiliary_targets_model_results}
\end{table*}

\begin{table*}[h!]
\centering
\small
\makebox[\linewidth][c]{%
\begin{tabular}{l l c c c c c c c c}
\toprule
\multirow{2}{*}{Benchmark} & \multirow{2}{*}{Model} & \multirow{2}{*}{Acc@1\(\uparrow\)} & \multirow{2}{*}{\(\mathbb{E}[\mathrm{Acc}]\uparrow\)} & \multicolumn{3}{c}{\(\mathrm{F1}_{w}\uparrow\)} & \multicolumn{3}{c}{MAE\(\downarrow\)} \\
 & & & & {Result} & {Termination} & {Resigned} & {\(N\!-\!i\!-\!1\)} & {\(V\)} & {\(\Delta t_{a}\)} \\
\midrule
\multirow{2}{*}{LIF-Aux} & Allie-Policy & \underline{0.5492} & \underline{0.4215} & -- & -- & \underline{0.6388} & -- & \bftab{0.2139} & \underline{5.0358} \\
 & UniMaia-Aux & \bftab{0.5495} & \bftab{0.4275} & \bftab{0.6618} & \bftab{0.5431} & \bftab{0.7270} & \bftab{20.4985} & \underline{0.3258} & \bftab{4.3000} \\
\midrule
\multirow{2}{*}{LGB-Aux} & Allie-Policy & \underline{0.5452} & \underline{0.4393} & -- & -- & \underline{0.1231} & -- & \bftab{0.3337} & \underline{122.3217} \\
 & UniMaia-Aux & \bftab{0.5622} & \bftab{0.4464} & \bftab{0.3824} & \bftab{0.6135} & \bftab{0.6223} & \bftab{26.3534} & \underline{0.3981} & \bftab{118.3532} \\
\bottomrule
\end{tabular}%
}
\caption[{Auxiliary targets results with the first 10 plies omitted.}]{Auxiliary targets results. We \textbf{omit} the first 10 plies from each game in the benchmark. Best values are in bold and second-best values are underlined.}
\label{tab:auxiliary_targets_skip_first_10_model_results}
\end{table*}

Unlike the standard prompt-conditioned benchmarks, LIF-Aux and LGB-Aux include temporal metadata such as time controls, remaining clock time, and previous move delay in the prompt, more closely matching the auxiliary-target training setup.

In addition to move prediction accuracy, we evaluate the auxiliary targets introduced in Appendix~\ref{app:unimaia_aux_final_configuration}. We report weighted F1 (\(\mathrm{F1}_w\)) for categorical targets and MAE for scalar targets.

UniMaia-Aux consistently outperforms Allie-Policy on both LIF-Aux and LGB-Aux, with and without the first 10 plies omitted. The gains are generally larger after excluding opening plies, suggesting that temporal metadata improves behavioral modeling beyond opening memorization.

UniMaia-Aux also substantially improves resignation prediction on over-the-board games, indicating stronger modeling of human gameplay behavior. Overall, it outperforms Allie-Policy on nearly all auxiliary targets except value prediction. We note that UniMaia-Aux is trained on substantially less data and uses a cross-entropy objective over discrete outcomes, whereas Allie uses MSE for value prediction, which may partially explain the MAE differences.

\subsection{Metadata-Conditioned Benchmarks}
\label{app:evaluation_metadata_conditioned}

Tables~\ref{tab:metadata-conditioned_top_move_accuracy_skip_first_10_model_results}
through~\ref{tab:metadata-conditioned_expected_accuracy_model_results}
report metadata-conditioned benchmark results, including evaluations with the first 10 plies omitted to reduce opening memorization effects.

\begin{table*}[h!]
\centering
\small
\begin{tabular}{l c c c c c}
\toprule
{Model} & {ABB (ours)} & {ABB (reported)} & {M1-S} & {M1} & {M2R} \\
\midrule
Lc0-CF (t3) & 0.4179 & -- & 0.4318 & 0.4351 & 0.4136 \\
Lc0-CF (t82) & 0.4113 & -- & 0.4285 & 0.4289 & 0.4062 \\
Lc0-CF (BT3) & 0.4157 & -- & 0.4310 & 0.4340 & 0.4127 \\
Lc0-CF (BT4) & 0.4133 & -- & 0.4255 & 0.4305 & 0.4079 \\
Lc0-CF (BT4-spsa) & 0.4145 & -- & 0.4223 & 0.4303 & 0.4091 \\
Maia-1\textsuperscript{*} & 0.5078 & 0.516 \(\pm\) 0.001 & 0.5094 & 0.5135 & 0.5135 \\
Maia-1\textsuperscript{**} & 0.5089 & -- & 0.5149 & 0.5173 & 0.5138 \\
Maia-2 (Blitz) & 0.5230 & 0.520 \(\pm\) 0.001 & 0.5246 & \underline{0.5300} & 0.5284 \\
Maia-2 (Rapid) & 0.5204 & -- & 0.5253 & 0.5294 & 0.5311 \\
Maia-3-79M (Blitz) & -- & \bftab{0.571 \(\pm\) 0.001} & -- & -- & -- \\
Allie-Policy & \bftab{0.5643} & \underline{0.557 \(\pm\) 0.001} & \bftab{0.5733} & \bftab{0.5717} & \bftab{0.5598} \\
ChessCLIP & 0.0813 & -- & 0.0813 & 0.0803 & 0.0838 \\
ChessGPT-base\textsuperscript{\dag} & 0.4336 & -- & 0.4375 & -- & 0.4305 \\
ChessGPT-chat\textsuperscript{\ddag} & 0.1697 & -- & 0.1719 & -- & 0.1647 \\
GPT-3.5\textsuperscript{\S} & 0.5410 & 0.537 \(\pm\) 0.001 & 0.5565 & -- & 0.5390 \\
GPT-3.5\textsuperscript{\dag} & 0.5498 & -- & \underline{0.5667} & -- & 0.5489 \\
UniMaia & \underline{0.5548} & -- & \underline{0.5667} & -- & \underline{0.5536} \\
UniMaia-Aux & 0.5471 & -- & 0.5624 & -- & 0.5493 \\
\bottomrule
\end{tabular}
\caption[Metadata-conditioned benchmark prediction results with the first 10 plies omitted, measured by Acc@1.]{\textbf{Metadata-conditioned} benchmark prediction results, measured by \textbf{Acc@1}. We \textbf{omit} the first 10 plies from each game in the benchmark. Best values are in bold and second-best values are underlined. \textsuperscript{*}Evaluated only with the official models. \textsuperscript{**}Evaluated with both the official and unofficial models. \textsuperscript{\dag}Evaluated with the Elo and time control PGN header. \textsuperscript{\ddag}Evaluated with the general policy prompt. \textsuperscript{\S}Evaluated with the GM PGN header.} 
\label{tab:metadata-conditioned_top_move_accuracy_skip_first_10_model_results}
\end{table*}

\begin{table}[h!]
\centering
\small
\begin{tabular}{l c c c c}
\toprule
{Model} & {ABB} & {M1-S} & {M1} & {M2R} \\
\midrule
Lc0-CF (t3) & 0.3035 & 0.3145 & 0.3165 & 0.3009 \\
Lc0-CF (t82) & 0.3028 & 0.3137 & 0.3159 & 0.3002 \\
Lc0-CF (BT3) & 0.3049 & 0.3170 & 0.3183 & 0.3027 \\
Lc0-CF (BT4) & 0.3063 & 0.3175 & 0.3198 & 0.3036 \\
Lc0-CF (BT4-spsa) & 0.3244 & 0.3360 & 0.3386 & 0.3212 \\
Maia-1\textsuperscript{*} & 0.3868 & 0.3883 & 0.3902 & 0.3898 \\
Maia-1\textsuperscript{**} & 0.3888 & 0.3916 & 0.3932 & 0.3900 \\
Maia-2 (Blitz) & 0.4102 & 0.4144 & \underline{0.4166} & 0.4137 \\
Maia-2 (Rapid) & 0.4056 & 0.4107 & 0.4131 & 0.4130 \\
Allie-Policy & \bftab{0.4409} & \underline{0.4476} & \bftab{0.4475} & \underline{0.4354} \\
ChessCLIP & 0.0700 & 0.0670 & 0.0686 & 0.0742 \\
UniMaia & 0.4224 & 0.4317 & -- & 0.4182 \\
UniMaia-Aux & \underline{0.4382} & \bftab{0.4488} & -- & \bftab{0.4427} \\
\bottomrule
\end{tabular}
\caption[{Metadata-conditioned results with the first 10 plies omitted, measured by \(\mathbb{E}[\mathrm{Acc}]\).}]{\textbf{Metadata-conditioned} results, measured by \(\bm{\mathbb{E}[\mathrm{Acc}]}\). We \textbf{omit} the first 10 plies from each game in the benchmark. Best values are in bold and second-best values are underlined. \textsuperscript{*}Evaluated only with the official models. \textsuperscript{**}Evaluated with both the official and unofficial models.}
\label{tab:metadata-conditioned_expected_accuracy_skip_first_10_model_results}
\end{table}

For ABB, we additionally report published results from prior work where available. Because the Allie and Maia-3 evaluations use broadly compatible protocols, these values are approximately comparable, although all rankings in this work are based on our reproduced results under the unified benchmark harness.

Across metadata-conditioned benchmarks, UniMaia remains competitive with specialized human move prediction models despite being optimized primarily for prompt-conditioned policy modulation. Allie-Policy achieves the strongest overall Acc@1 performance, while UniMaia-Aux attains the best expected accuracy on several benchmarks, including M1-S and M2R.

Notably, UniMaia-Aux consistently improves expected accuracy relative to UniMaia despite slightly lower Acc@1 scores, suggesting that auxiliary temporal objectives improve policy calibration even when temporal metadata is unavailable at evaluation time.

\begin{table}[h!]
\centering
\small
\begin{tabular}{l c c c c}
\toprule
{Model} & {ABB} & {M1-S} & {M1} & {M2R} \\
\midrule
Lc0-CF (t3) & 0.4091 & 0.4234 & 0.4249 & 0.4040 \\
Lc0-CF (t82) & 0.4036 & 0.4195 & 0.4199 & 0.3981 \\
Lc0-CF (BT3) & 0.4073 & 0.4231 & 0.4240 & 0.4032 \\
Lc0-CF (BT4) & 0.4053 & 0.4181 & 0.4210 & 0.3991 \\
Lc0-CF (BT4-spsa) & 0.4032 & 0.4116 & 0.4177 & 0.3972 \\
Maia-1\textsuperscript{*} & 0.5043 & 0.5072 & 0.5119 & 0.5094 \\
Maia-1\textsuperscript{**} & 0.5057 & 0.5122 & \underline{0.5158} & 0.5096 \\
Maia-2 (Blitz) & 0.4961 & 0.4998 & 0.5048 & 0.4964 \\
Maia-2 (Rapid) & 0.4956 & 0.5038 & 0.5065 & 0.5031 \\
Allie-Policy & \bftab{0.5568} & \bftab{0.5654} & \bftab{0.5650} & \bftab{0.5497} \\
ChessCLIP & 0.0799 & 0.0803 & 0.0797 & 0.0821 \\
ChessGPT-base\textsuperscript{\dag} & 0.4408 & 0.4452 & -- & 0.4380 \\
ChessGPT-chat\textsuperscript{\ddag} & 0.1671 & 0.1661 & -- & 0.1631 \\
GPT-3.5\textsuperscript{\dag} & 0.5398 & \underline{0.5554} & -- & 0.5370 \\
UniMaia & \underline{0.5447} & 0.5534 & -- & \underline{0.5413} \\
UniMaia-Aux & 0.5361 & 0.5495 & -- & 0.5356 \\
\bottomrule
\end{tabular}
\caption[Metadata-conditioned results, measured by Acc@1.]{\textbf{Metadata-conditioned} results, measured by \textbf{Acc@1}. Best values are in bold and second-best values are underlined. \textsuperscript{*}Evaluated only with the official models. \textsuperscript{**}Evaluated with both the official and unofficial models. \textsuperscript{\dag}Evaluated with the Elo and time control PGN header. \textsuperscript{\ddag}Evaluated with the general policy prompt.}
\label{tab:metadata-conditioned_top_move_accuracy_model_results}
\end{table}

\begin{table}[h!]
\centering
\small
\begin{tabular}{l c c c c}
\toprule
{Model} & {ABB} & {M1-S} & {M1} & {M2R} \\
\midrule
Lc0-CF (t3) & 0.2985 & 0.3093 & 0.3104 & 0.2955 \\
Lc0-CF (t82) & 0.2977 & 0.3088 & 0.3098 & 0.2948 \\
Lc0-CF (BT3) & 0.2997 & 0.3118 & 0.3120 & 0.2970 \\
Lc0-CF (BT4) & 0.3009 & 0.3123 & 0.3133 & 0.2979 \\
Lc0-CF (BT4-spsa) & 0.3176 & 0.3295 & 0.3306 & 0.3140 \\
Maia-1\textsuperscript{*} & 0.3816 & 0.3840 & 0.3862 & 0.3827 \\
Maia-1\textsuperscript{**} & 0.3840 & 0.3877 & 0.3897 & 0.3829 \\
Maia-2 (Blitz) & 0.3860 & 0.3921 & \underline{0.3938} & 0.3861 \\
Maia-2 (Rapid) & 0.3819 & 0.3890 & 0.3907 & 0.3858 \\
Allie-Policy & \bftab{0.4310} & \bftab{0.4388} & \bftab{0.4386} & \underline{0.4225} \\
ChessCLIP & 0.0652 & 0.0627 & 0.0640 & 0.0685 \\
UniMaia & 0.4084 & 0.4181 & -- & 0.4025 \\
UniMaia-Aux & \underline{0.4222} & \underline{0.4330} & -- & \bftab{0.4234} \\
\bottomrule
\end{tabular}
\caption[{Metadata-conditioned results, measured by \(\mathbb{E}[\mathrm{Acc}]\).}]{\textbf{Metadata-conditioned} results, measured by \(\bm{\mathbb{E}[\mathrm{Acc}]}\). Best values are in bold and second-best values are underlined. \textsuperscript{*}Evaluated only with the official models. \textsuperscript{**}Evaluated with both the official and unofficial models.}
\label{tab:metadata-conditioned_expected_accuracy_model_results}
\end{table}

Tables~\ref{tab:metadata-conditioned_top_move_accuracy_model_results}
and~\ref{tab:metadata-conditioned_expected_accuracy_model_results}
contain nearly identical trends, although Maia-2 scores are lower because the first 10 plies were excluded during training.
 
\subsection{Inference Strategy Ablations}
\label{sec:evaluation_opening_instruction_following}

This section provides detailed results for different prompting and inference strategies used with large language models, including ChessGPT and GPT-3.5 Turbo Instruct (Appendix~\ref{app:baseline_details}), illustrating the sensitivity of prompt-conditioned models to input formatting and prompt design.


\begin{table*}[h!]
\centering
\small
\begin{tabular}{l c c c c c c}
\toprule
{Model} & {LOB-P} & {LOB-C} & {LIF-D} & {LIF} & {LIF-T10} & {LGB} \\
\midrule
ChessGPT-base (prompt) & \bftab{0.7998} & \bftab{0.8955} & \underline{0.3801} & \bftab{0.4503} & \underline{0.4060} & \bftab{0.5436} \\
ChessGPT-base (Elo + time control PGN) & \underline{0.6100} & \underline{0.6186} & \bftab{0.3808} & \underline{0.4464} & \bftab{0.4153} & \underline{0.4788} \\
\midrule
ChessGPT-chat (simple, duplicate) & 0.3417 & 0.2272 & 0.0981 & 0.1097 & 0.0964 & 0.1532 \\
ChessGPT-chat (simple, original) & 0.2601 & 0.1730 & 0.0892 & 0.0992 & 0.0831 & 0.1243 \\
ChessGPT-chat (general policy) & 0.1630 & 0.1503 & 0.1494 & 0.1637 & 0.1480 & 0.1501 \\
\bottomrule
\end{tabular}
\caption[{ChessGPT prompt-conditioned results, measured by Acc@1.}]{ChessGPT \textbf{prompt-conditioned} results, measured by \textbf{Acc@1}. Best values are in bold and second-best values are underlined.}
\label{tab:chessgpt_prompt-conditioned_top_move_accuracy_model_results}
\end{table*}

\begin{table*}[h!]
\centering
\small
\begin{tabular}{l c c c c}
\toprule
{Model} & {LIF-D} & {LIF} & {LIF-T10} & {LGB} \\
\midrule
ChessGPT-base (prompt) & \underline{0.3672} & \underline{0.4303} & \underline{0.3888} & \underline{0.4336} \\
ChessGPT-base (Elo + time control PGN) & \bftab{0.3813} & \bftab{0.4440} & \bftab{0.4158} & \bftab{0.4339} \\
\midrule
ChessGPT-chat (simple, duplicate) & 0.0933 & 0.1020 & 0.0874 & 0.1129 \\
ChessGPT-chat (simple, original) & 0.0873 & 0.0939 & 0.0727 & 0.0918 \\
ChessGPT-chat (general policy) & 0.1490 & 0.1557 & 0.1491 & 0.1564 \\
\bottomrule
\end{tabular}
\caption[{ChessGPT prompt-conditioned results with the first 10 plies omitted, measured by Acc@1.}]{ChessGPT \textbf{prompt-conditioned} results, measured by \textbf{Acc@1}. We \textbf{omit} the first 10 plies from each game in the benchmark. Best values are in bold and second-best values are underlined.}
\label{tab:chessgpt_prompt-conditioned_top_move_accuracy_skip_first_10_model_results}
\end{table*}

\begin{table*}[h!]
\centering
\small
\begin{tabular}{l c c c}
\toprule
{Model} & {ABB} & {M1-S} & {M2R} \\
\midrule
ChessGPT-base (prompt) & \underline{0.4330} & \underline{0.4330} & \underline{0.4310} \\
ChessGPT-base (Elo + time control PGN) & \bftab{0.4408} & \bftab{0.4452} & \bftab{0.4380} \\
\midrule
ChessGPT-chat (simple, duplicate) & 0.1128 & 0.1006 & 0.1146 \\
ChessGPT-chat (simple, original) & 0.0933 & 0.0889 & 0.0998 \\
ChessGPT-chat (general policy) & 0.1671 & 0.1661 & 0.1631 \\
\bottomrule
\end{tabular}
\caption[{ChessGPT metadata-conditioned results, measured by Acc@1.}]{ChessGPT \textbf{metadata-conditioned} results, measured by \textbf{Acc@1}. Best values are in bold and second-best values are underlined.}
\label{tab:chessgpt_metadata-conditioned_top_move_accuracy_model_results}
\end{table*}

\begin{table*}[h!]
\centering
\small
\begin{tabular}{l c c c}
\toprule
{Model} & {ABB} & {M1-S} & {M2R} \\
\midrule
ChessGPT-base (prompt) & \underline{0.4260} & \underline{0.4257} & \underline{0.4248} \\
ChessGPT-base (Elo + time control PGN) & \bftab{0.4336} & \bftab{0.4375} & \bftab{0.4305} \\
\midrule
ChessGPT-chat (simple, duplicate) & 0.0980 & 0.0898 & 0.1071 \\
ChessGPT-chat (simple, original) & 0.0874 & 0.0821 & 0.0932 \\
ChessGPT-chat (general policy) & 0.1697 & 0.1719 & 0.1647 \\
\bottomrule
\end{tabular}
\caption[{ChessGPT metadata-conditioned results with the first 10 plies omitted, measured by Acc@1.}]{ChessGPT \textbf{metadata-conditioned} results, measured by \textbf{Acc@1}. We \textbf{omit} the first 10 plies from each game in the benchmark. Best values are in bold and second-best values are underlined.}
\label{tab:chessgpt_metadata-conditioned_top_move_accuracy_skip_first_10_model_results}
\end{table*}


\begin{table*}[h!]
\centering
\small
\begin{tabular}{l c c c c c c}
\toprule
{Model} & {LOB-P} & {LOB-C} & {LIF-D} & {LIF} & {LIF-T10} & {LGB} \\
\midrule
GPT-3.5 (GM PGN header) & \bftab{0.5748} & \bftab{0.6431} & \underline{0.4344} & \underline{0.5218} & \underline{0.4684} & \bftab{0.5525} \\
GPT-3.5 (GM PGN header + Prompt) & 0.5118 & \underline{0.5186} & 0.4341 & 0.5166 & 0.4607 & 0.5138 \\
GPT-3.5 (Prompt + Original PGN header) & \underline{0.5196} & 0.5183 & 0.4338 & 0.5166 & 0.4593 & 0.5147 \\
GPT-3.5 (Elo + Time Control PGN header) & -- & -- & \bftab{0.4457} & \bftab{0.5312} & \bftab{0.4778} & \underline{0.5514} \\
\bottomrule
\end{tabular}
\caption[{GPT-3.5 prompt-conditioned results, measured by Acc@1.}]{GPT-3.5 \textbf{prompt-conditioned results}, measured by \textbf{Acc@1}. Best values are in bold and second-best values are underlined.}
\label{tab:gpt-3.5_prompt-conditioned_top_move_accuracy_model_results}
\end{table*}

\begin{table*}[h!]
\centering
\small
\begin{tabular}{l c c c c}
\toprule
{Model} & {LIF-D} & {LIF} & {LIF-T10} & {LGB} \\
\midrule
GPT-3.5 (GM PGN header) & 0.4531 & 0.5352 & 0.4841 & \underline{0.5562} \\
GPT-3.5 (GM PGN header + Prompt) & \underline{0.4584} & \underline{0.5411} & \underline{0.4849} & 0.5555 \\
GPT-3.5 (Prompt + Original PGN header) & 0.4579 & 0.5410 & \underline{0.4849} & 0.5543 \\
GPT-3.5 (Elo + Time Control PGN header) & \bftab{0.4608} & \bftab{0.5444} & \bftab{0.4914} & \bftab{0.5569} \\
\bottomrule
\end{tabular}
\caption[{GPT-3.5 prompt-conditioned results with the first 10 plies omitted, measured by Acc@1.}]{GPT-3.5 \textbf{prompt-conditioned} results, measured by \textbf{Acc@1}. We \textbf{omit} the first 10 plies from each game in the benchmark. Best values are in bold and second-best values are underlined.}
\label{tab:gpt-3.5_prompt-conditioned_top_move_accuracy_skip_first_10_model_results}
\end{table*}

\begin{table*}[h!]
\centering
\small
\begin{tabular}{l c c c}
\toprule
{Model} & {ABB} & {M1-S} & {M2R} \\
\midrule
GPT-3.5 (GM PGN header) & \underline{0.5298} & \underline{0.5451} & \underline{0.5248} \\
GPT-3.5 (GM PGN header + Prompt) & 0.5271 & 0.5409 & 0.5212 \\
GPT-3.5 (Elo + Time Control PGN header) & \bftab{0.5398} & \bftab{0.5554} & \bftab{0.5370} \\
\bottomrule
\end{tabular}
\caption[{GPT-3.5 metadata-conditioned results, measured by Acc@1.}]{GPT-3.5 \textbf{metadata-conditioned} results, measured by \textbf{Acc@1}. Best values are in bold and second-best values are underlined.}
\label{tab:gpt-3.5_metadata-conditioned_top_move_accuracy_model_results}
\end{table*}

\begin{table*}[h!]
\centering
\small
\begin{tabular}{l c c c}
\toprule
{Model} & {ABB} & {M1-S} & {M2R} \\
\midrule
GPT-3.5 (GM PGN header) & 0.5410 & 0.5565 & 0.5390 \\
GPT-3.5 (GM PGN header + Prompt) & \underline{0.5493} & \underline{0.5640} & \underline{0.5461} \\
GPT-3.5 (Elo + Time Control PGN header) & \bftab{0.5498} & \bftab{0.5667} & \bftab{0.5489} \\
\bottomrule
\end{tabular}
\caption[{GPT-3.5 metadata-conditioned results with the first 10 plies omitted, measured by Acc@1.}]{GPT-3.5 \textbf{metadata-conditioned} results, measured by \textbf{Acc@1}. We \textbf{omit} the first 10 plies from each game in the benchmark. Best values are in bold and second-best values are underlined.}
\label{tab:gpt-3.5_metadata-conditioned_top_move_accuracy_skip_first_10_model_results}
\end{table*}

%% file: appendices/appendix-extended-analysis.tex
\section{Extended Analysis}
\label{app:extended_analysis}

This appendix provides additional analysis of UniMaia, focusing on policy continuity, internal residual dynamics, cross-attention behavior, and prompt sensitivity.

\subsection{Policy Continuity Across Elo}
\label{app:policy_continuity}

We provide additional policy-continuity analyses beyond those shown in Section~\ref{sec:analysis}. All evaluations use Elo values from 0--3400, including regions below the training distribution (Appendix~\ref{app:rating_system}).

Figure~\ref{fig:elo_plot_various_stages} in Section~\ref{sec:analysis} evaluates four representative positions spanning the early opening, late opening, middlegame, and endgame. The corresponding PGNs and prompts are provided in Appendix~\ref{app:policy_continuity_across_positions_prompts}.

We additionally analyze continuity under opening conditioning by fixing the initial position and varying the opening prompt. We consider prompts corresponding to the King's Pawn, Queen's Pawn, English, and Zukertort openings. Full prompts are provided in Appendix~\ref{app:policy_continuity_under_opening_conditioning_prompts}.

\begin{figure}[t]
    \centering
    \includegraphics[width=\linewidth]{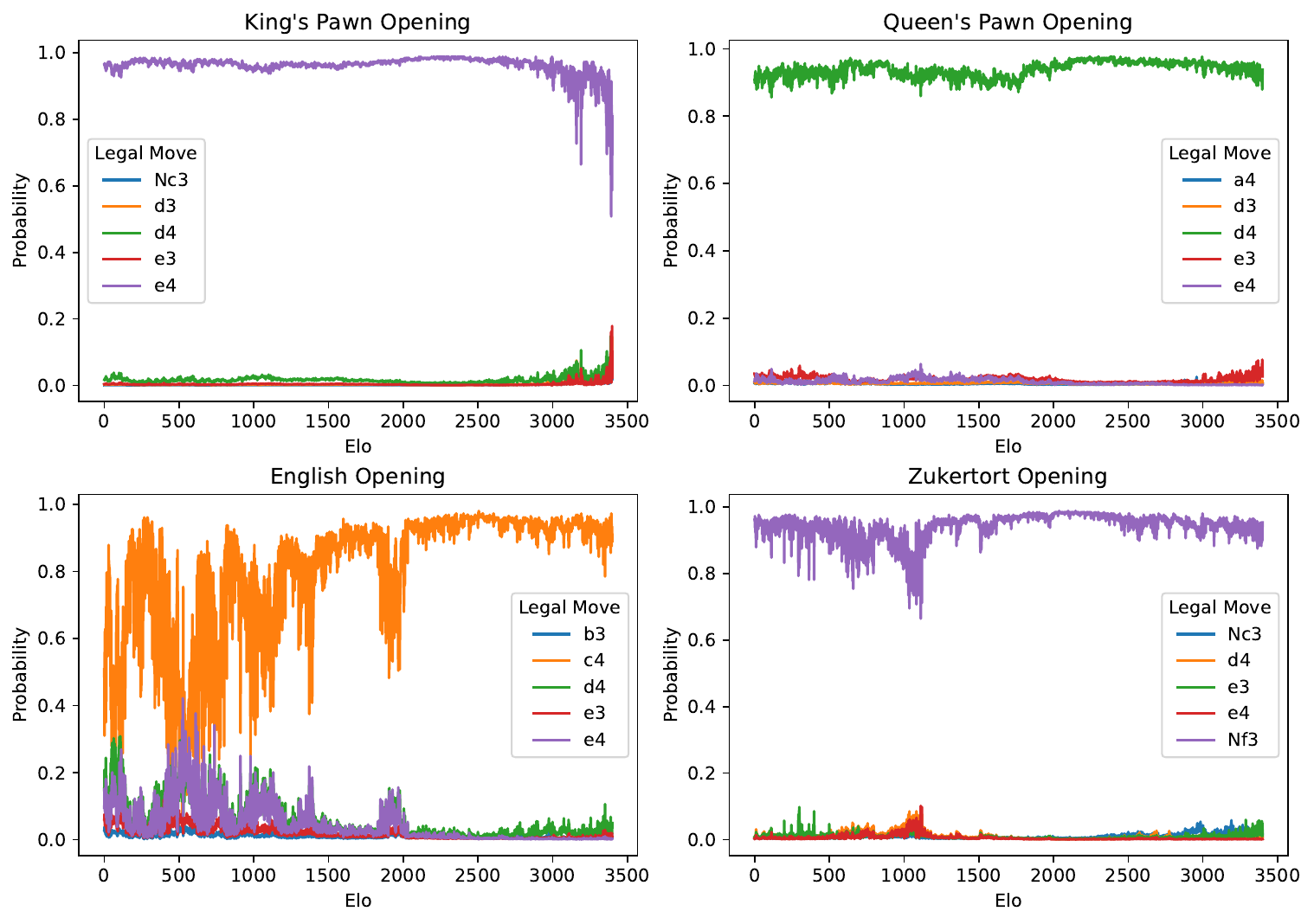}
    \caption{Top-5 move probabilities predicted by UniMaia as a function of Elo for opening-conditioned prompts.}
    \label{fig:elo_plot_opening_appendix}
\end{figure}

Most openings exhibit relatively stable top-move probabilities across Elo, while the English Opening shows larger variation in the range 400--1300 Elo, likely reflecting common transpositions and differing opening-selection behavior across rating levels.

\subsubsection{Quantitative Diagnostics of Policy Continuity}
\label{app:quantitative_policy_continuity_plots}

\begin{figure}[h!]
    \centering
    \includegraphics[width=\linewidth]{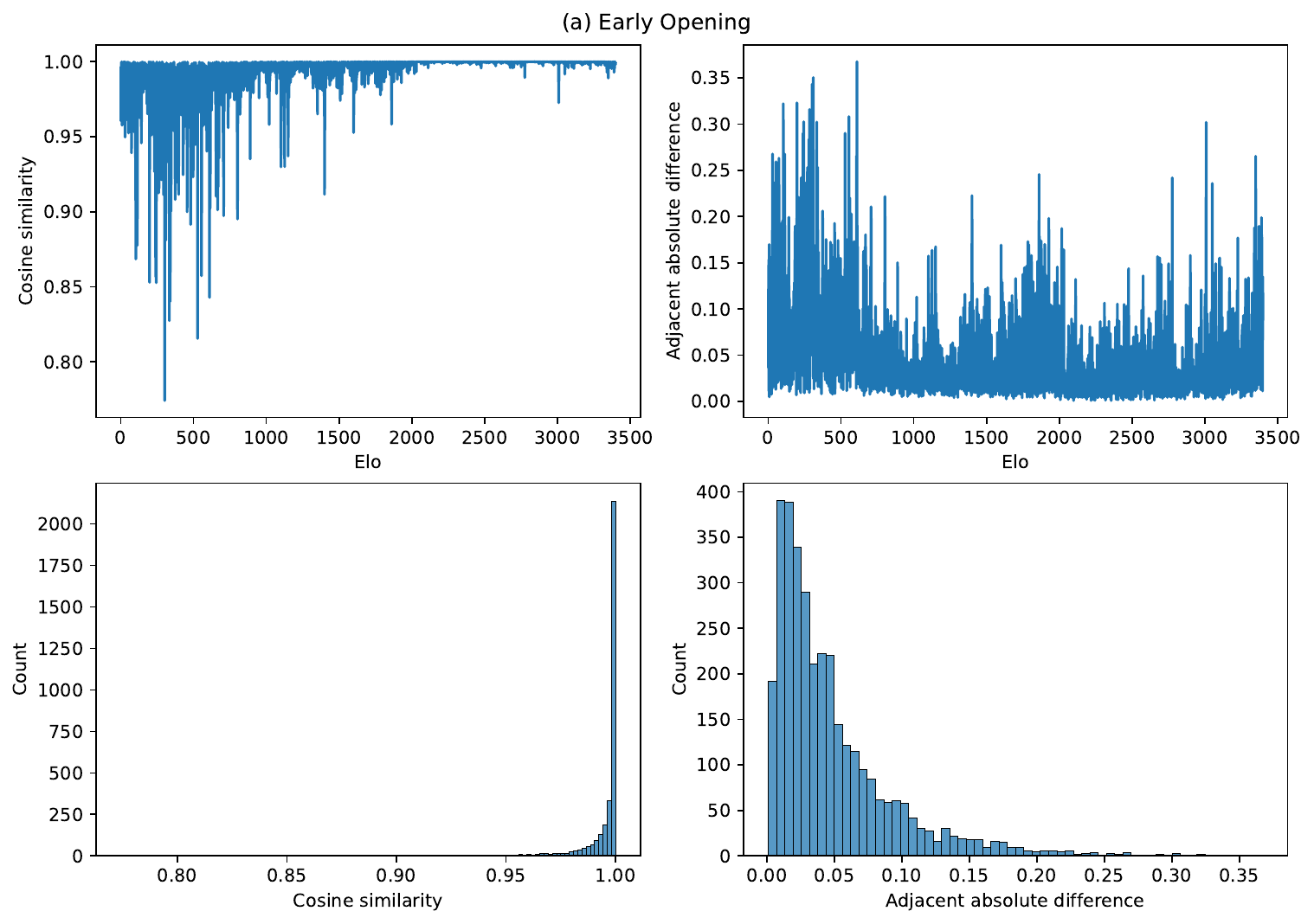}
    \caption[Quantitative policy continuity diagnostics across Elo values (early opening).]{Quantitative policy continuity diagnostics for an early-opening position.}
    \label{fig:elo_smoothness_sicilian}
\end{figure}

\begin{figure}[h!]
    \centering
    \includegraphics[width=\linewidth]{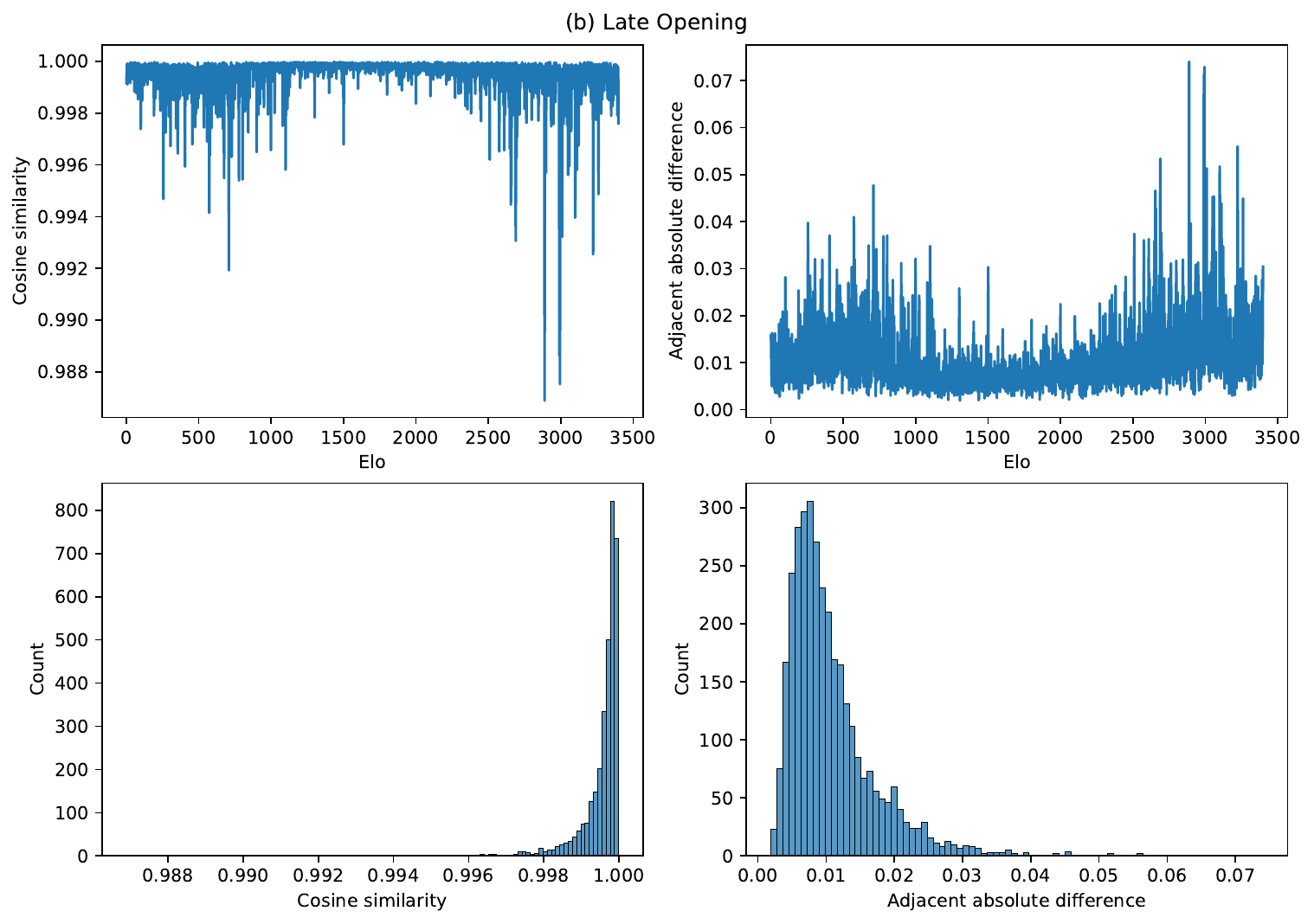}
    \caption[Quantitative policy continuity diagnostics across Elo values (late opening).]{Quantitative policy continuity diagnostics for a late-opening position.}
    \label{fig:elo_smoothness_tarrasch}
\end{figure}

\begin{figure}[h!]
    \centering
    \includegraphics[width=\linewidth]{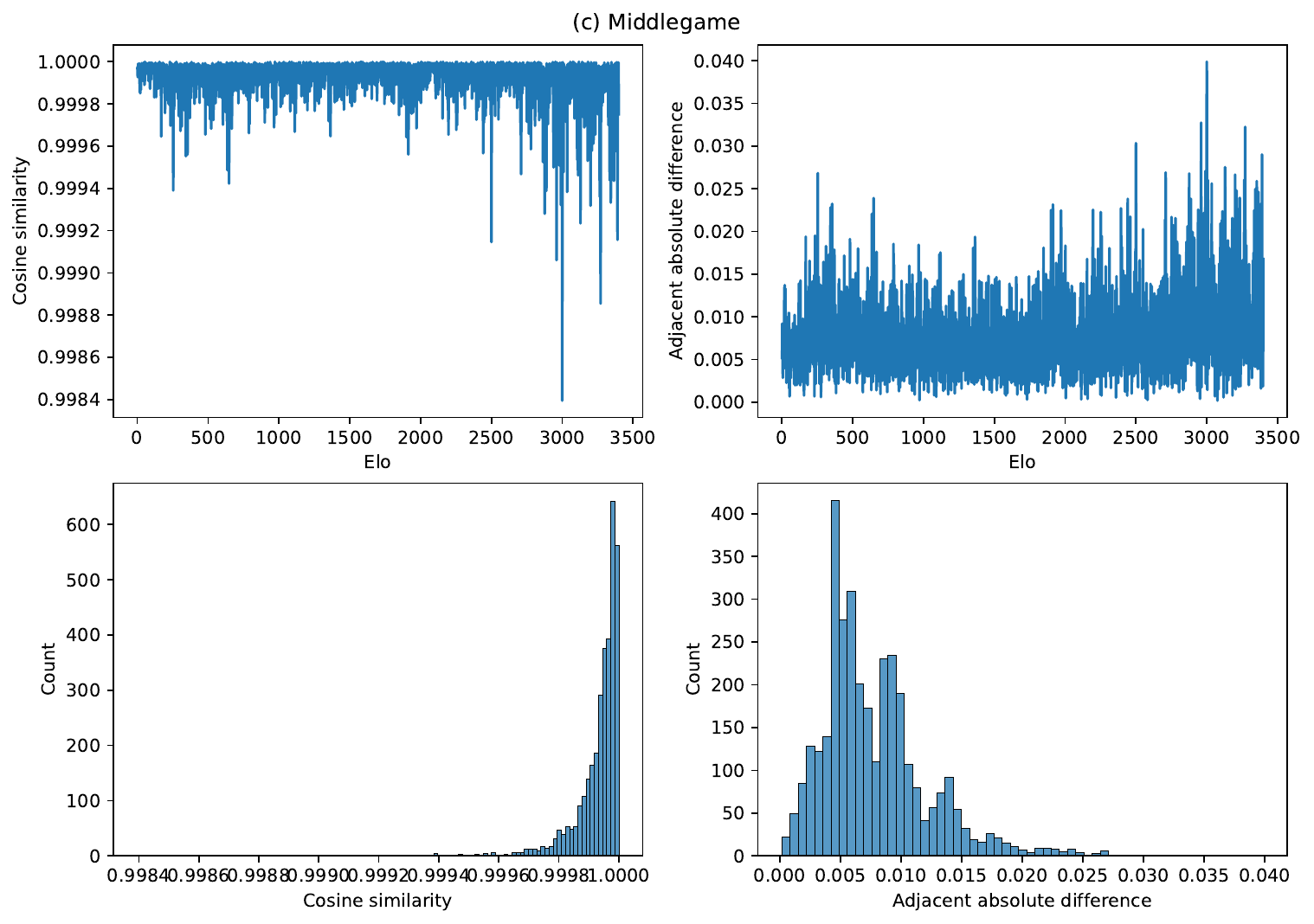}
    \caption[Quantitative policy continuity diagnostics across Elo values (middlegame).]{Quantitative policy continuity diagnostics for a middlegame position.}
    \label{fig:elo_smoothness_middlegame}
\end{figure}

\begin{figure}[h!]
    \centering
    \includegraphics[width=\linewidth]{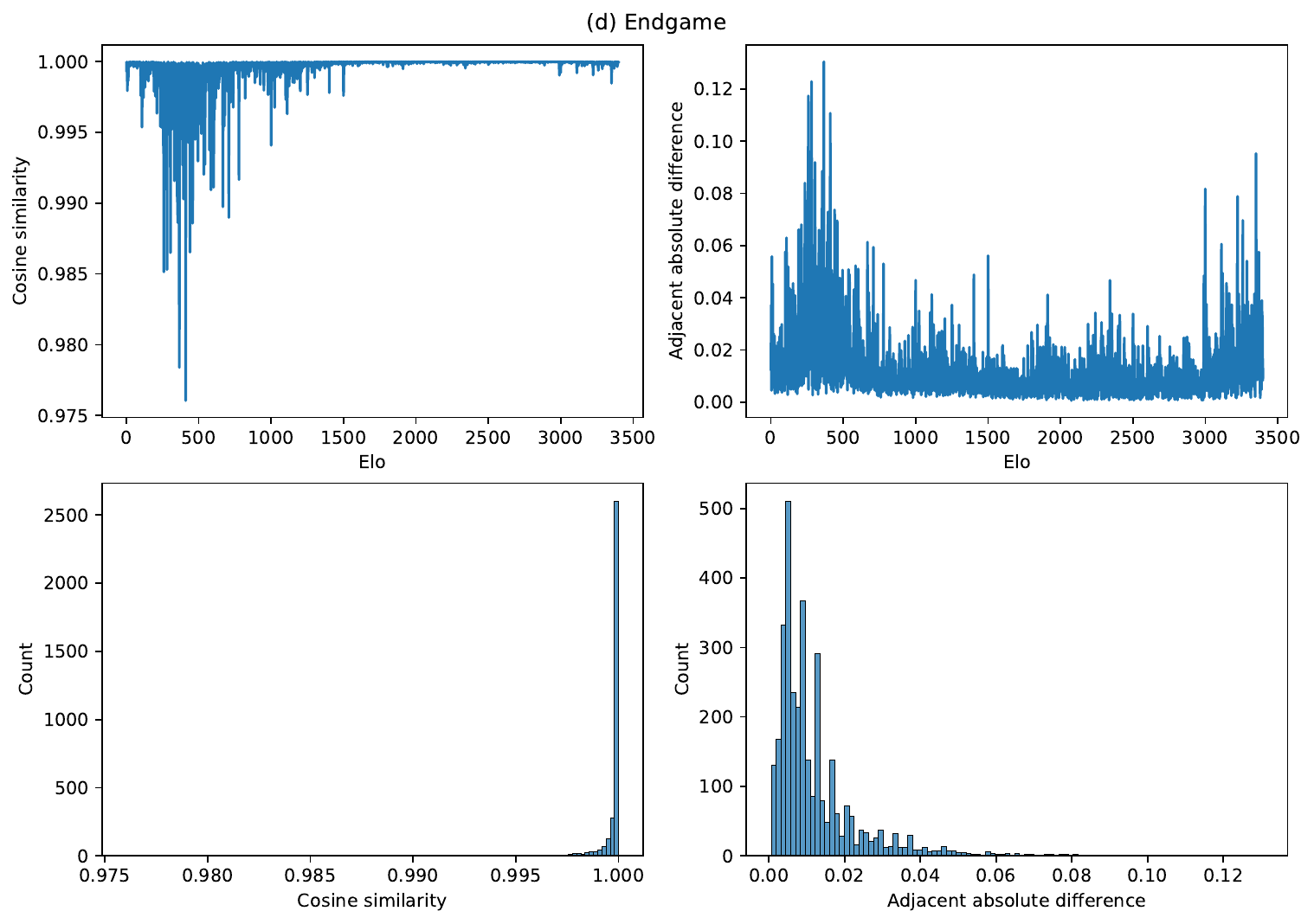}
    \caption[Quantitative policy continuity diagnostics across Elo values (endgame).]{Quantitative policy continuity diagnostics for an endgame position.}
    \label{fig:elo_smoothness_endgame}
\end{figure}

Figures~\ref{fig:elo_smoothness_sicilian}--\ref{fig:elo_smoothness_endgame}
provide quantitative diagnostics of policy continuity across Elo values, complementing the qualitative analysis in Section~\ref{sec:analysis} and Appendix~\ref{app:policy_continuity}. For each position, we measure cosine similarity and L2 distance between policies at adjacent Elo values, together with their corresponding distributions.

Overall, adjacent policies remain highly similar across Elo ranges, while policy differences vary smoothly rather than exhibiting abrupt transitions. This behavior is consistent across opening, middlegame, and endgame positions, supporting the interpretation that Elo conditioning induces a largely continuous modulation of the policy distribution.

\subsection{Mechanistic Interpretability}
\label{app:mechanistic_interpretability}

To better understand how prompt conditioning modifies the pretrained chess policy, we analyze the internal behavior of UniMaia on the first 256 examples of LIF.

\subsubsection{Layer-wise Residual Contributions}
\label{app:layerwise_contribution}

We extract the outputs of the zero-initialized projection layers in the ControlNet branch, which we refer to as \emph{ControlNet residual updates}. These residuals quantify the magnitude of the modifications introduced relative to the frozen Lc0 backbone.

\begin{figure}[t]
    \centering
    \includegraphics[width=\linewidth]{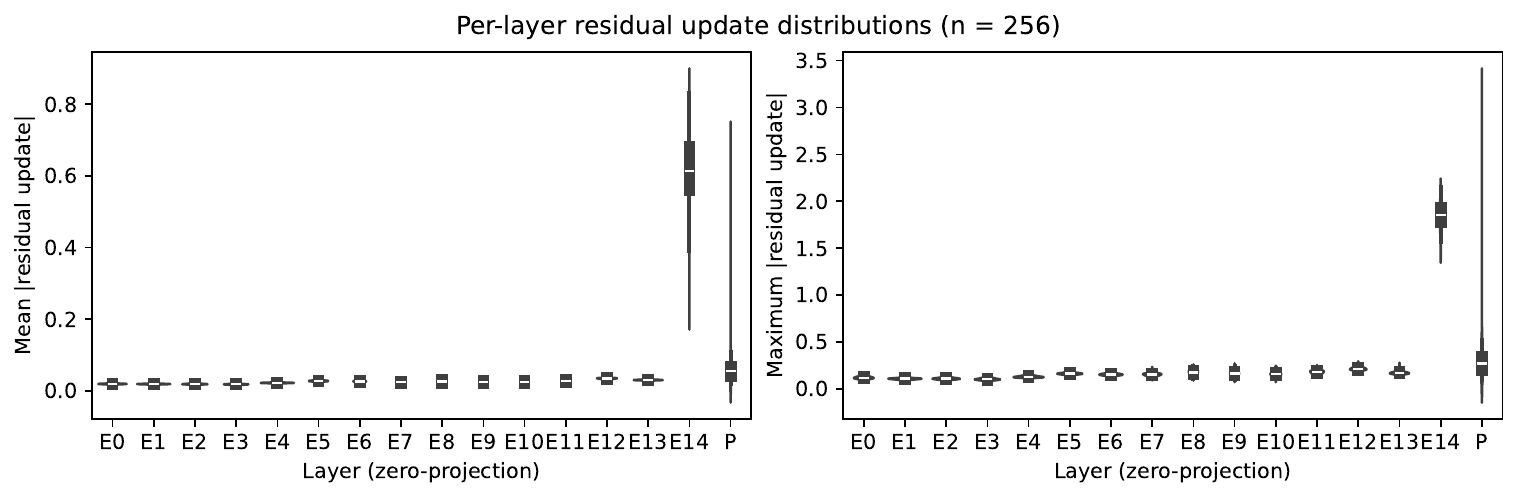}
    \caption{Distribution of mean and maximum absolute residual updates for each encoder layer and the controllable policy head.}
    \label{fig:activation_values_appendix}
\end{figure}

Figure~\ref{fig:activation_values_appendix} shows that encoder layers 0--13 produce relatively small residual updates, whereas encoder layer 14 and the controllable policy head produce substantially larger updates. This suggests that prompt-conditioned modulation is concentrated near the output of the network rather than uniformly distributed across all layers.

This behavior is consistent with the zero-initialization design of ControlNet-style conditioning, which encourages earlier layers to remain close to the pretrained Lc0 representations while allowing later layers to introduce controllable modifications.

\begin{figure}[t]
    \centering
    \includegraphics[width=\linewidth]{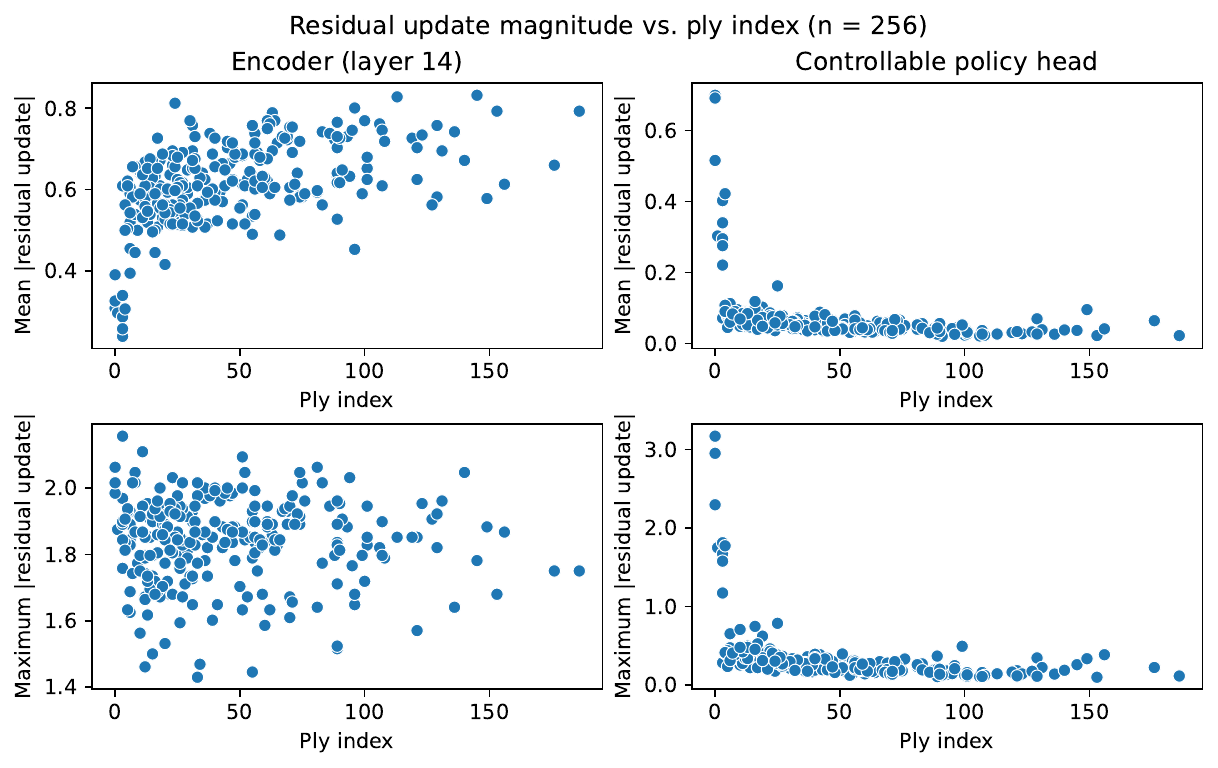}
    \caption{Mean and maximum absolute residual updates as a function of ply index for encoder layer 14 and the controllable policy head.}
    \label{fig:activation_values_vs_ply_index_appendix}
\end{figure}

Figure~\ref{fig:activation_values_vs_ply_index_appendix} further shows that the controllable policy head contributes most strongly in the early opening, while encoder layer 14 becomes increasingly influential later in the game.

One possible explanation is that earlier layers are more tightly constrained by architectural biases inherited from Lc0-CF, such as the geometric attention bias (GAB)~\cite{monroe2024mastering,monroe2026chessformer}. Later layers and the controllable policy head interface more directly with the policy output and may therefore admit greater flexibility for prompt-conditioned modulation. We emphasize, however, that this interpretation remains speculative and is not directly verified in this work.

To analyze how concentrated the residual updates are, we define the \emph{residual concentration ratio} as the ratio of the maximum absolute residual update to the mean absolute residual update.

\begin{figure}[t]
    \centering
    \includegraphics[width=\linewidth]{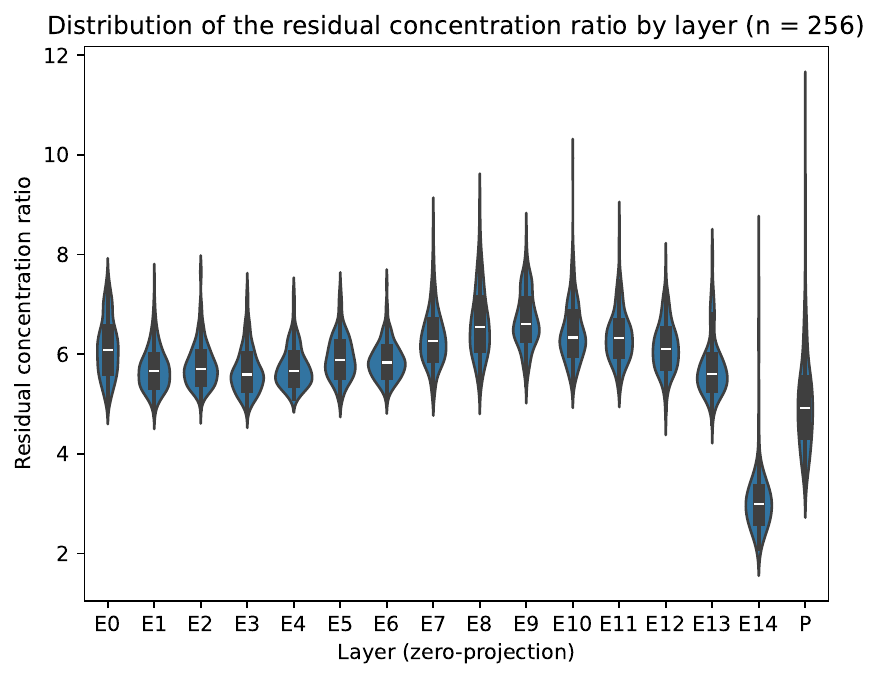}
    \caption{Residual concentration ratio for each encoder layer and the controllable policy head.}
    \label{fig:activation_values_ratio_appendix}
\end{figure}

As shown in Figure~\ref{fig:activation_values_ratio_appendix}, the ratio is approximately 6 for encoder layers 0--13, approximately 3 for encoder layer 14, and approximately 5 for the controllable policy head. This suggests that residual updates in the early encoder layers and policy head are concentrated in relatively few dimensions, whereas encoder layer 14 exhibits more distributed modifications.

Because residual updates in layers 0--13 are small in magnitude, we further compare the encoder representations of UniMaia and the frozen Lc0 backbone using cosine similarity.

\begin{figure}[t]
    \centering
    \includegraphics[width=\linewidth]{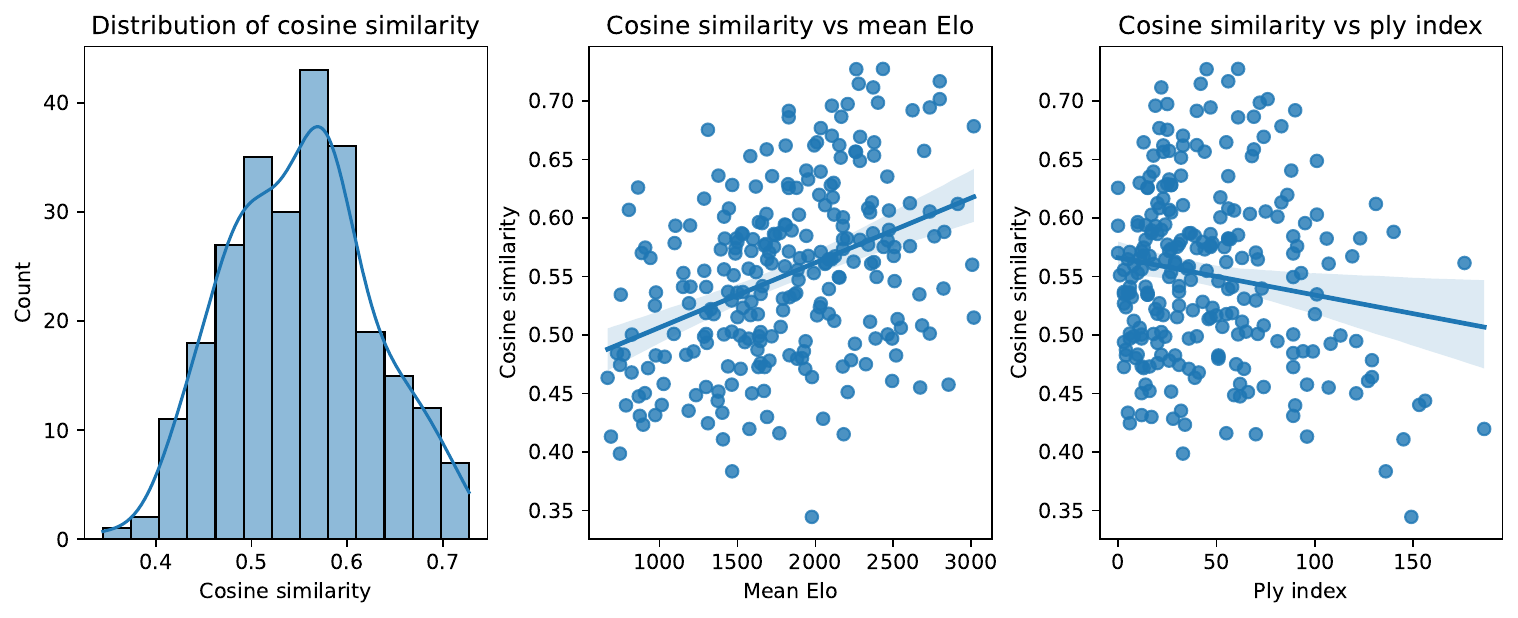}
    \caption{Cosine similarity between flattened encoder layer 13 representations of UniMaia and the frozen Lc0 backbone. Left: distribution over examples. Middle: similarity as a function of mean Elo. Right: similarity as a function of ply index.}
    \label{fig:cosine_similarity_layer_13_appendix}
\end{figure}

The cosine similarity has mean \(0.551\) and standard deviation \(0.075\), indicating moderate similarity between the induced representations.

We observe a moderate positive correlation between cosine similarity and mean player Elo (\(r = 0.396\), \(p = \num{4.70e-11}\)), suggesting that higher-rated games induce representations closer to the pretrained Lc0 policy. This is consistent with the interpretation that stronger play aligns more closely with the pretrained backbone, whereas lower-rated or stylistically distinct play requires larger prompt-conditioned deviations. We additionally observe a weak negative correlation with ply index (\(r = -0.157\), \(p = 0.0119\)), indicating that representations diverge slightly from the base model as games progress.

\subsubsection{Cross-attention Sinks}
\label{app:cross_attention_sinks}

To analyze the interaction between prompts and board representations, we examine cross-attention behavior in the ControlNet branch.

Following prior work on attention sinks~\cite{xiao2023efficientattentionsinks}, we compute the proportion of examples for which the largest cross-attention weight is assigned to the first token. Attention weights are averaged across heads and board squares before computing this statistic. Note that the ChessGPT tokenizer does not include a beginning-of-sequence token, so the first token is part of the prompt itself.

\begin{figure}[t]
    \centering
    \includegraphics[width=\linewidth]{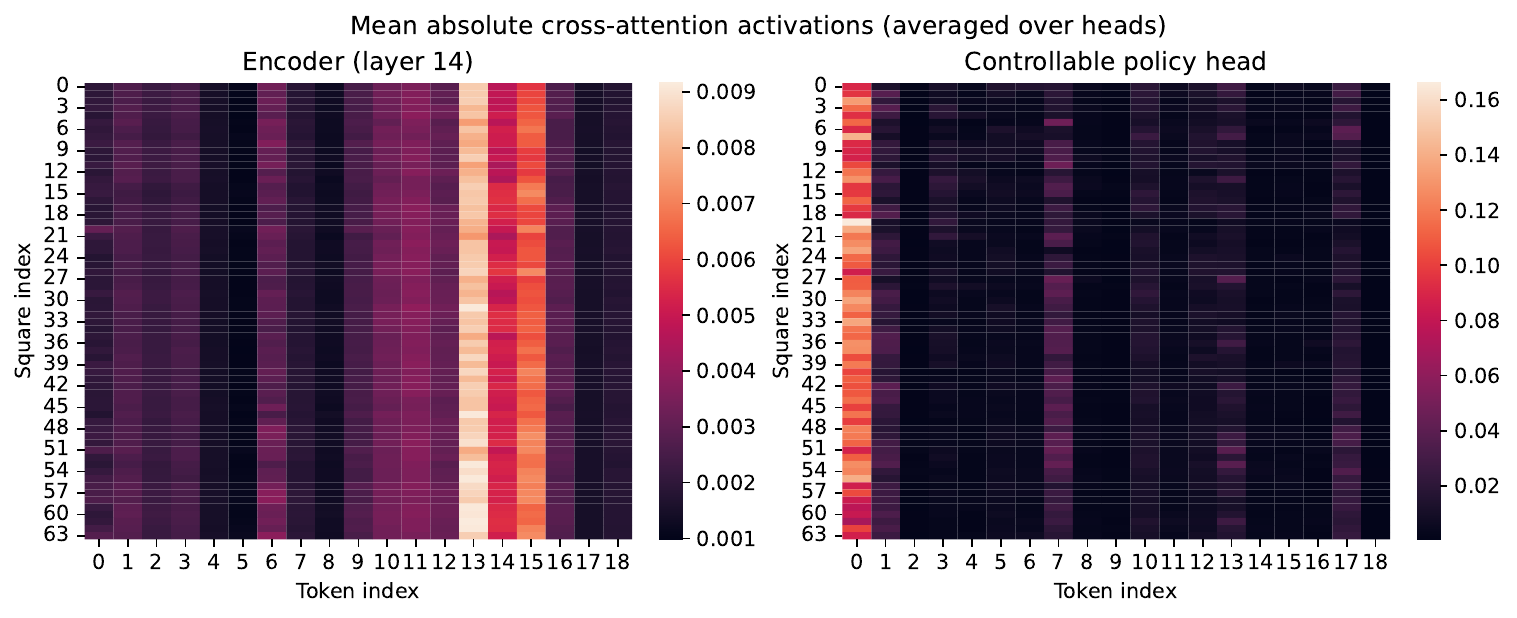}
    \caption{Mean absolute cross-attention activations averaged across heads. Left: encoder layer 14. Right: controllable policy head.}
    \label{fig:cross_attention_activations_appendix}
\end{figure}

Most layers exhibit strong attention-sink behavior, whereas a small subset rarely or never does so. In particular, encoder layers 2 and 14 almost never exhibit attention sinks (\(\approx 1\%\) and \(0\%\), respectively), while most remaining layers and the controllable policy head do so in more than \(90\%\) of evaluated prompts. The controllable policy head, together with encoder layers 8 and 13, exhibits attention sinks for all evaluated prompts.

Cross-attention activations are also highly similar across board squares. For a given text token, activations from different spatial locations remain nearly identical, as illustrated in Figure~\ref{fig:cross_attention_activations_appendix}.

Attempts to mitigate this behavior did not improve performance. Masking padding tokens in the cross-attention keys (Appendix~\ref{app:key_padding_mask_ablations}) slightly degraded performance, and adding gated attention~\cite{qiu2025gated} (Appendix~\ref{app:architecture_design}) did not recover this loss. We leave further investigation of these effects to future work.

\subsection{Failure Modes and Stress Cases}
\label{app:failure_modes}

UniMaia remains sensitive to relatively small prompt variations, even when the underlying instruction is semantically unchanged.

For example, the prompts ``\texttt{Open with the English Opening}'' and ``\texttt{Open with the English Opening.}'' can produce noticeably different move distributions despite differing only by punctuation.

Similarly, lexical variations can substantially affect opening-following behavior. After the position ``\texttt{1. e4 c5}'', the prompts ``\texttt{Play the Alapin Sicilian}'' and ``\texttt{Play the Alapin Sicilian Defense}'' both assign highest probability to \(\texttt{c3}\), the defining move of the Alapin Sicilian. In contrast, the prompt ``\texttt{Open with the Alapin Sicilian}'' instead assigns highest probability to \(\texttt{Nf3}\), with \(\texttt{c3}\) becoming only the second-ranked move.

One contributing factor is the distribution of prompt templates used during training. None of the pretraining templates, and only approximately \(33.4\%\) (\(\frac{684}{2048}\)) of instruction templates, begin with the verb ``Play'', while none begin with ``Open''. Many templates instead begin with contextual prefixes such as player names or titles. Consequently, prompts that more closely match the training distribution tend to produce more stable behavior than semantically equivalent but distributionally different phrasing.

This sensitivity is also qualitatively consistent with the prevalence of attention sinks (Appendix~\ref{app:cross_attention_sinks}). If a small subset of tokens disproportionately captures attention, minor tokenization or phrasing differences may propagate through the network and induce large downstream policy changes. While this does not establish causality, it provides a plausible explanation for the observed instability.

These examples are intended to illustrate stress cases rather than typical behavior. Nevertheless, they highlight an important limitation of prompt-conditioned chess policies: relatively small prompt perturbations can induce disproportionately large policy changes.

%% file: appendices/appendix-continuity-evaluation-details.tex
\section{Policy Continuity Evaluation Details}
\label{app:policy_continuity_evaluation}

This appendix summarizes the evaluation prompts used in the policy continuity analyses of Section~\ref{sec:analysis}. All evaluations sweep Elo values from 0--3400 to probe both in-distribution and out-of-distribution behavior.

\subsection{Policy Continuity Evaluation Prompts}
\label{app:policy_continuity_evaluation_prompts}

We evaluate policy continuity in two settings: (1) across representative positions spanning different game stages and (2) under opening-conditioned prompts from the initial position.

\subsubsection{Policy Continuity Across Positions}
\label{app:policy_continuity_across_positions_prompts}

For the analysis in Figure~\ref{fig:elo_plot_various_stages}, we evaluate four representative positions spanning the early opening, late opening, middlegame, and endgame. The corresponding prompts and PGNs are listed in Table~\ref{tab:policy_continuity_prompts}.

\begin{table*}[h!]
\centering
\small
\begin{tabular}{p{2.5cm} p{5.5cm} p{6cm}}
\toprule
\textbf{Stage} & \textbf{PGN} & \textbf{Prompt Template} \\
\midrule
Early Opening & 1. e4 c5 & \texttt{You are an anonymous white player rated \{elo\}, playing against an anonymous black player rated \{elo\}, using the Sicilian Alapin} \\
Late Opening & 1. d4 d5 2. c4 e6 3. Nc3 c5 4. cxd5 exd5 5. Nf3 Nf6 6. g3 Nc6 7. Bg2 & \texttt{You are an anonymous black player rated \{elo\}, playing against an anonymous white player rated \{elo\}, opening with the main line of the Tarrasch Defense's Prague Variation in the QGD.} \\
Middlegame & 1. e4 c5 2. Nf3 Nc6 3. Bc4 g6 4. c3 Bg7 5. Qb3 e6 6. d4 cxd4 7. cxd4 Nxd4 8. Nxd4 Bxd4 9. Qd3 Qb6 10. Be3 Bxe3 11. fxe3 Ne7 12. Nc3 & \texttt{Username prashu\_20091, a beginner with an ELO of \{elo\}, confidently plays as White, utilizing the Sicilian Defense: Old Sicilian with the opening moves 1. e4 c5 2. Nf3 Nc6.} \\
Endgame & 1. e4 d5 2. exd5 Qxd5 3. Nc3 Qa5 4. d4 c6 5. Nf3 Bg4 6. Be2 Bxf3 7. Bxf3 Nf6 8. O-O e6 9. Bf4 Bb4 10. Ne4 Nxe4 11. Bxe4 O-O 12. c3 Be7 13. b4 Qd8 14. Qh5 g6 15. Qh3 Nd7 16. Bh6 Re8 17. a3 Nf6 18. Bf3 Bf8 19. Bxf8 Rxf8 20. a4 Rc8 21. Rad1 Nd7 22. c4 Nb6 23. d5 cxd5 24. cxd5 Nxd5 25. Bxd5 exd5 26. Qb3 Qc7 27. Rxd5 Rfd8 28. Rxd8+ Rxd8 29. h3 Rd2 30. Qe3 Qd8 31. Qxa7 Rd1 32. Rxd1 Qxd1+ 33. Kh2 Qd6+ 34. g3 Qf6 35. Qe3 Qd6 36. Qc5 Qd2 37. a5 Kg7 38. Kg2 Kf6 39. Qb6+ Kg7 40. Qxb7 h5 41. a6 & \begin{minipage}[t]{\linewidth}\ttfamily The game in question was a rated blitz chess match hosted on the online chess platform Lichess.org on January 1, 2023. The match featured a rapid time control of five minutes with no additional time per move, denoted as "300+0." Playing with the white pieces was a player using the username "Talca" who held a rating of \{elo\}. The black pieces were commanded by player "pierobaldo" who held an identical rating of \{elo\} at the time of the game.

The opening played in this game was the Scandinavian Defense, specifically the Main Line variation, which is introduced with the moves 1. e4 d5 2. exd5 Qxd5 3. Nc3 Qa5. The game featured an average of \{elo\} Elo with a 0-point difference\normalfont\end{minipage} \\
\bottomrule
\end{tabular}
\caption{PGN and prompt templates used for policy continuity evaluation across representative positions (early opening, late opening, middlegame, and endgame).}
\label{tab:policy_continuity_prompts}
\end{table*}

\subsubsection{Policy Continuity Under Opening Conditioning}
\label{app:policy_continuity_under_opening_conditioning_prompts}

For the analysis in Figure~\ref{fig:elo_plot_opening_appendix} in Appendix~\ref{app:policy_continuity}, we evaluate policy continuity from the initial position under different opening-conditioned prompts:

\begin{enumerate}
    \item \textbf{King's Pawn Opening}: \texttt{You are an anonymous white player rated \{elo\}, playing against an anonymous black player rated \{elo\}, using the King's Pawn Opening}
    \item \textbf{Queen's Pawn Opening}: \texttt{You are an anonymous white player rated \{elo\}, playing against an anonymous black player rated \{elo\}, using the Queen's Pawn Opening}
    \item \textbf{English Opening}: \texttt{You are an anonymous white player rated \{elo\}, playing against an anonymous black player rated \{elo\}, using the English Opening}
    \item \textbf{Zukertort Opening}: \texttt{You are an anonymous white player rated \{elo\}, playing against an anonymous black player rated \{elo\}, using the Zukertort Opening}
\end{enumerate}

%% file: appendices/appendix-performance-elo-range.tex
\section{Performance by Elo Range}
\label{app:evaluation_performance_elo_range}

\begin{figure*}[h]
    \centering
    \includegraphics[width=\linewidth]{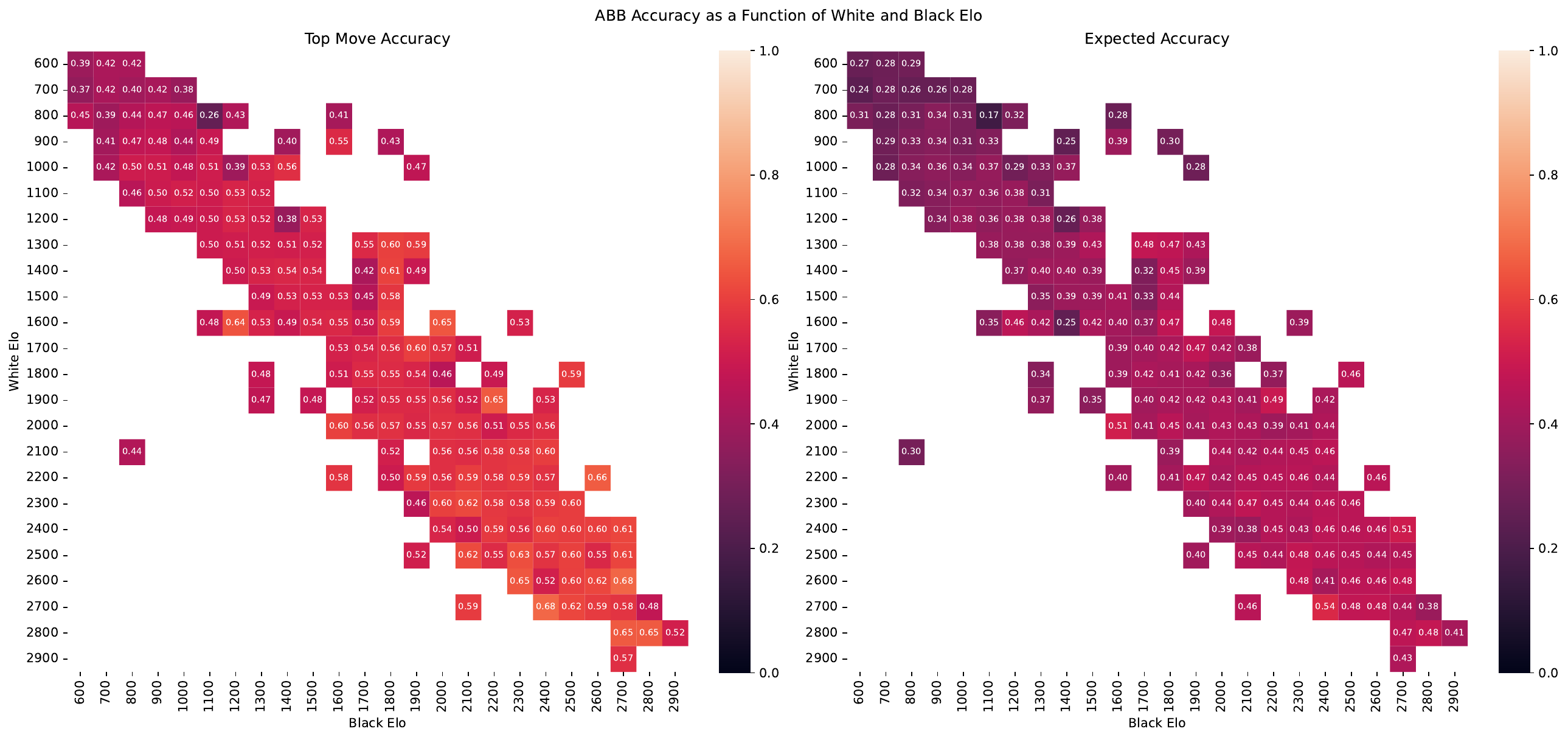}
    \caption{ABB accuracy as a function of White and Black Elo. Left: top-move accuracy. Right: expected accuracy.}
    \label{fig:abb_accuracy_heatmap_white_and_black_elo}
\end{figure*}

\begin{figure*}[h]
    \centering
    \includegraphics[width=\linewidth]{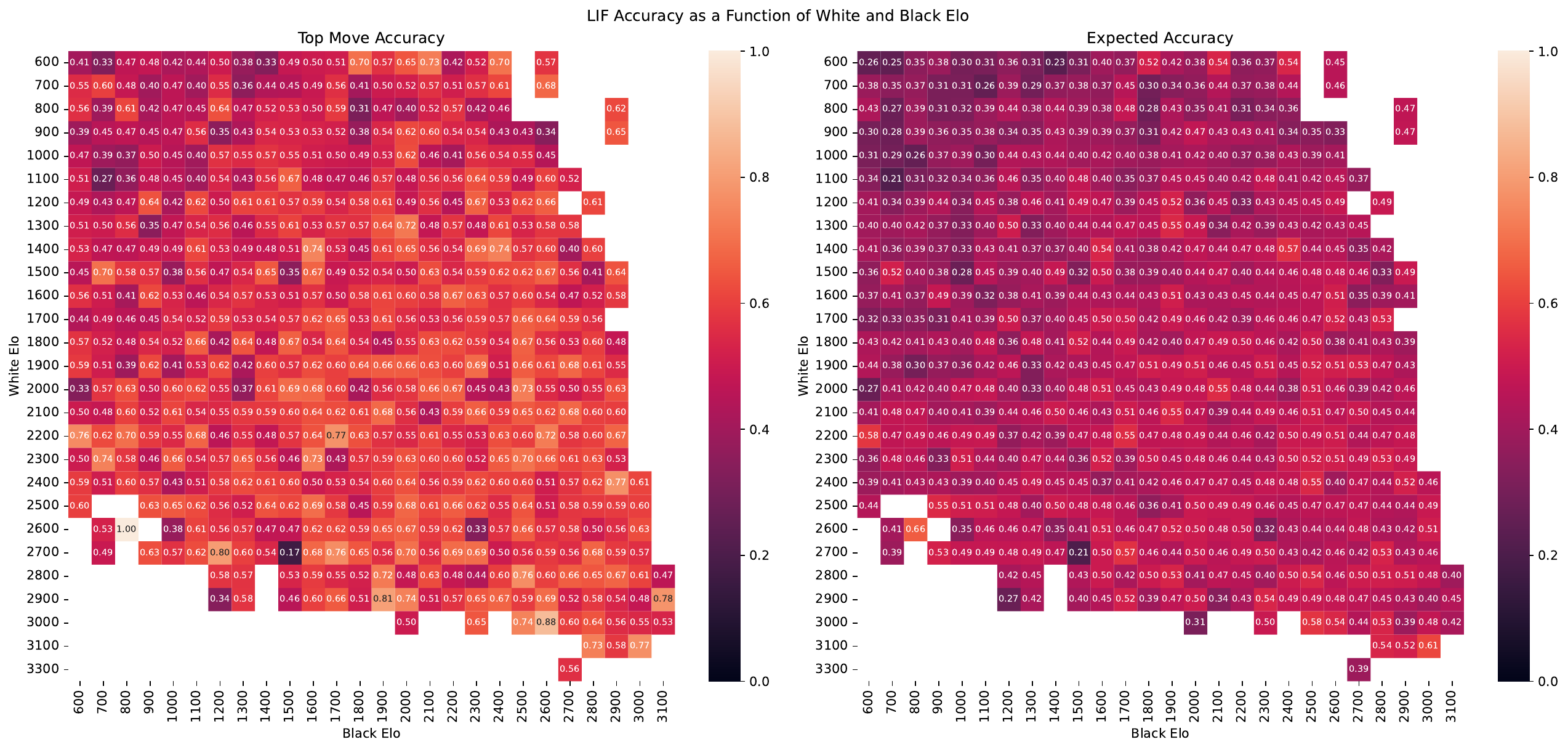}
    \caption{LIF accuracy as a function of White and Black Elo. Left: top-move accuracy. Right: expected accuracy.}
    \label{fig:lif_accuracy_heatmap_white_and_black_elo}
\end{figure*}

Figures~\ref{fig:abb_accuracy_heatmap_white_and_black_elo}
and~\ref{fig:lif_accuracy_heatmap_white_and_black_elo}
show benchmark performance as a function of White and Black Elo.

Both top-move and expected accuracy generally increase with player Elo, reflecting the greater consistency of higher-rated play. Lower-rated games exhibit more stochasticity and stylistic variation, making moves harder to predict.

The benchmark families also exhibit different heatmap patterns. LIF shows greater variability across Elo combinations because it uses randomly sampled prompt templates with varying metadata, whereas ABB uses a fixed template with consistent metadata, producing more stable patterns.

We present the accuracy and size heatmaps for all benchmarks. Together, these visualizations illustrate how performance varies across player skill levels and dataset distributions.

\begin{figure}[h]
    \centering
    \includegraphics[width=\linewidth]{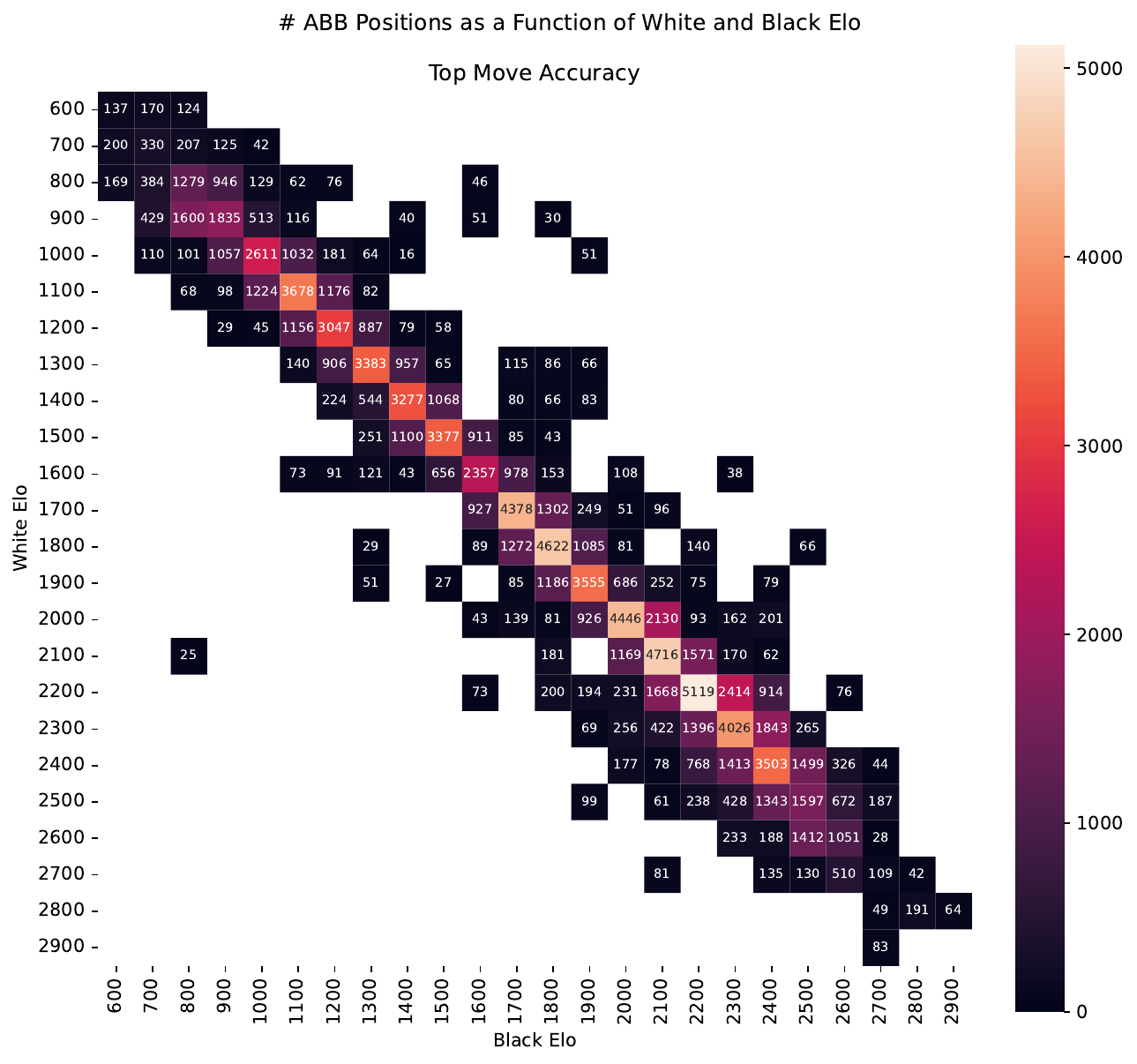}
    \caption{Number of ABB entries in each rounded White and Black Elo bucket.}
    \label{fig:abb_size_heatmap_white_and_black_elo}
\end{figure}

\begin{figure}[h]
    \centering
    \includegraphics[width=\linewidth]{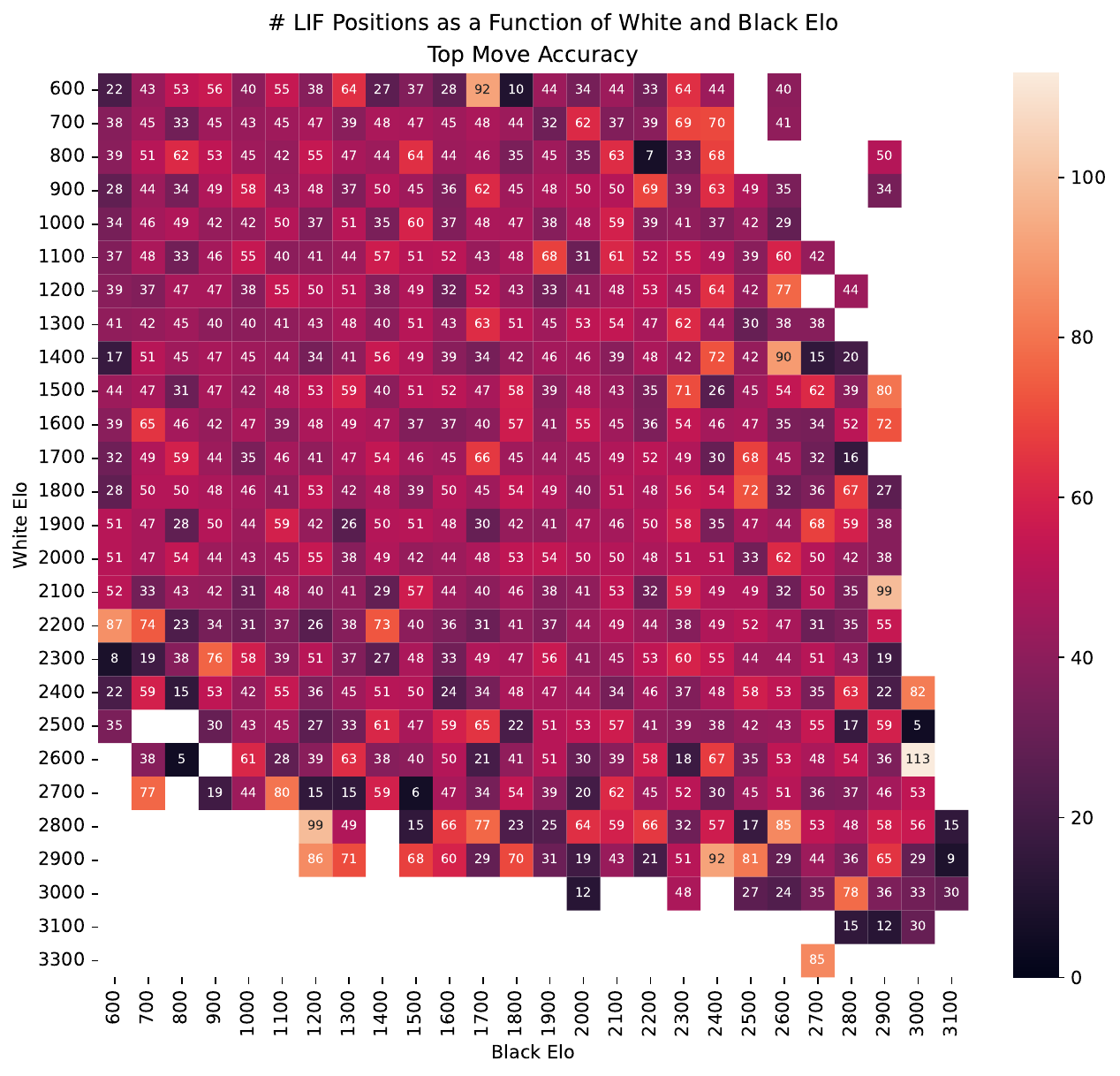}
    \caption{Number of LIF entries in each rounded White and Black Elo bucket.}
    \label{fig:lif_size_heatmap_white_and_black_elo}
\end{figure}

\begin{figure*}[h]
    \centering
    \includegraphics[width=\linewidth]{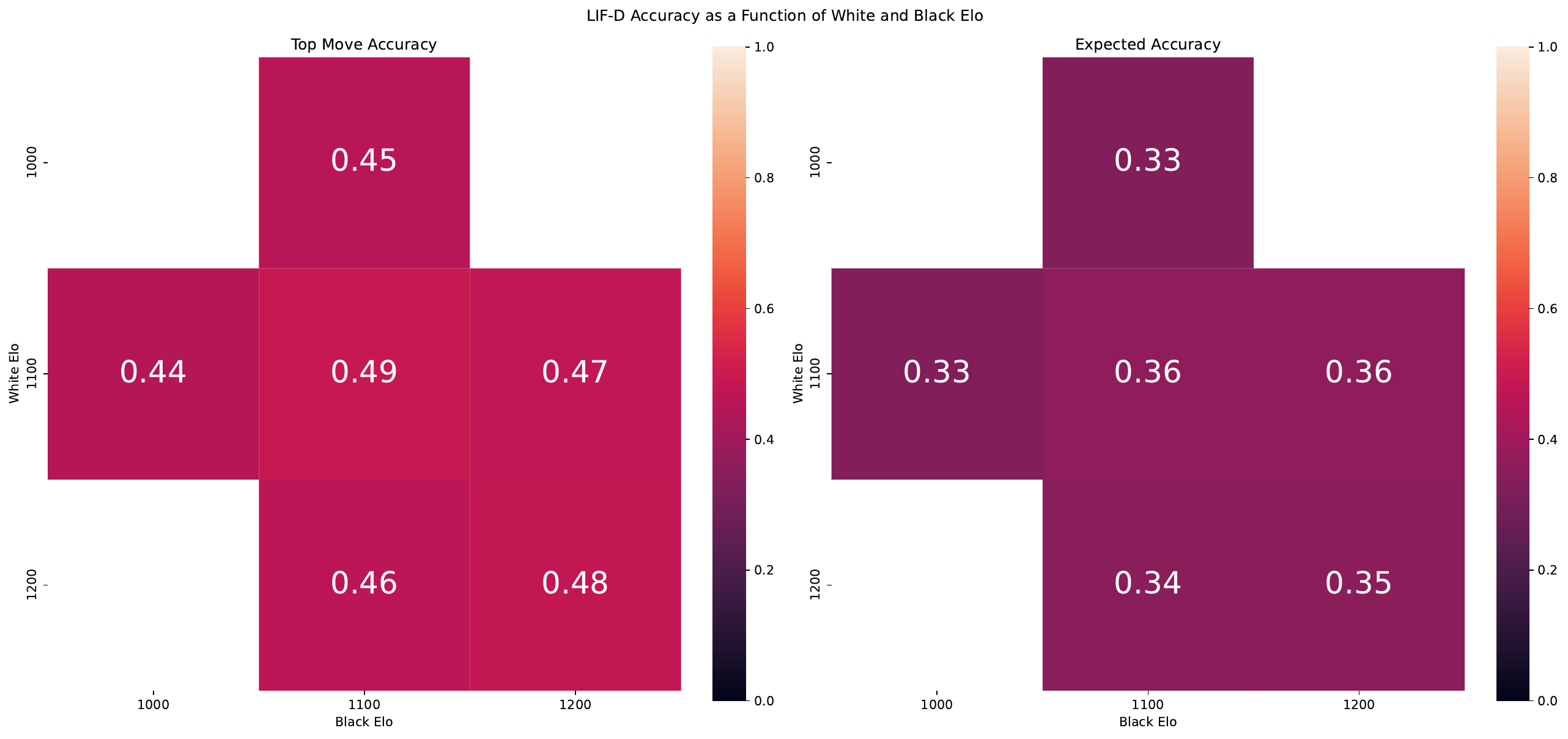}
    \caption{LIF-D accuracy as a function of White and Black Elo. Left: top-move accuracy. Right: expected accuracy.}
    \label{fig:lif-d_accuracy_heatmap_white_and_black_elo}
\end{figure*}

\begin{figure}[h]
    \centering
    \includegraphics[width=\linewidth]{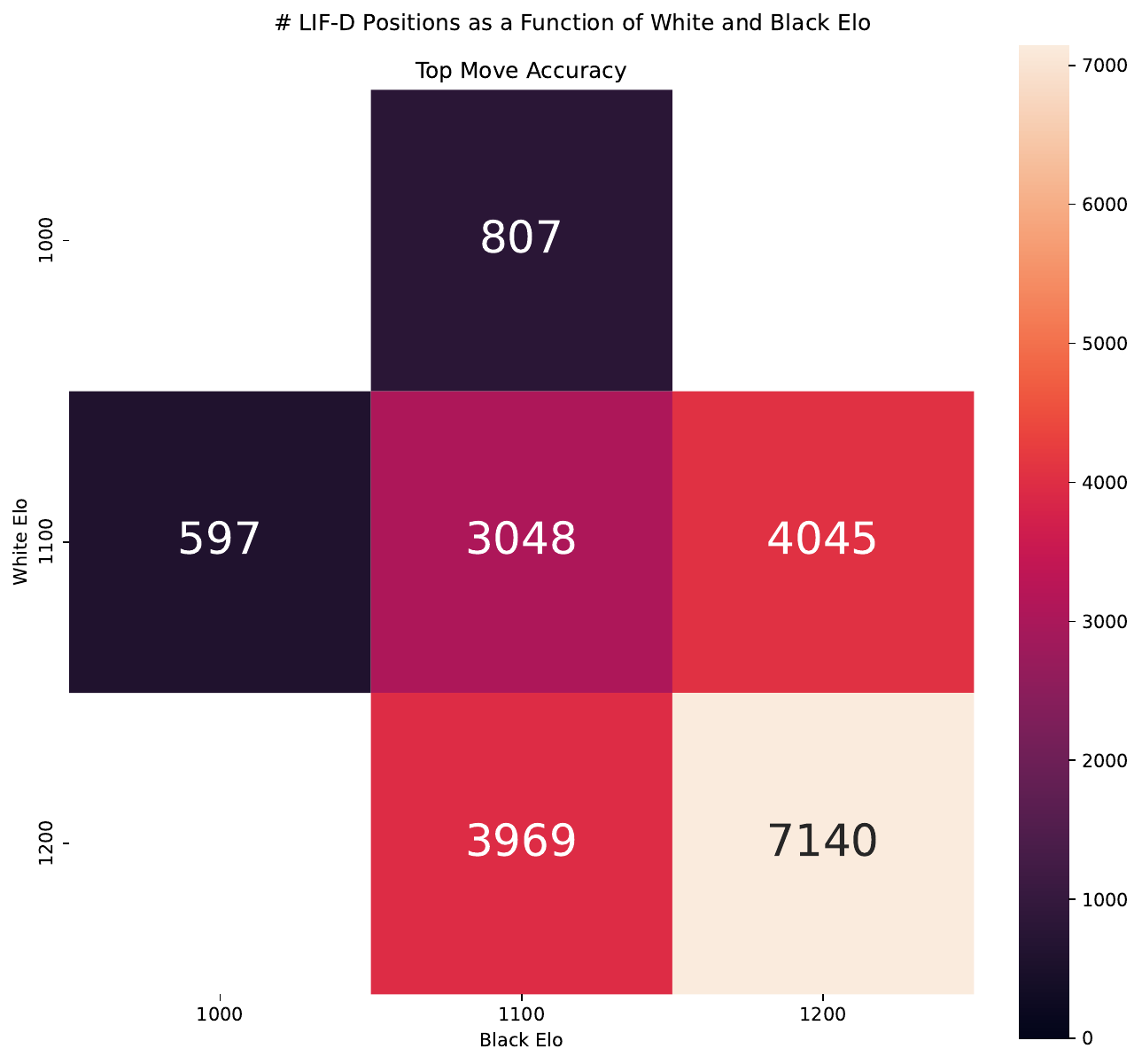}
    \caption{Number of LIF-D entries in each rounded White and Black Elo bucket.}
    \label{fig:lif-d_size_heatmap_white_and_black_elo}
\end{figure}

\begin{figure*}[h]
    \centering
    \includegraphics[width=\linewidth]{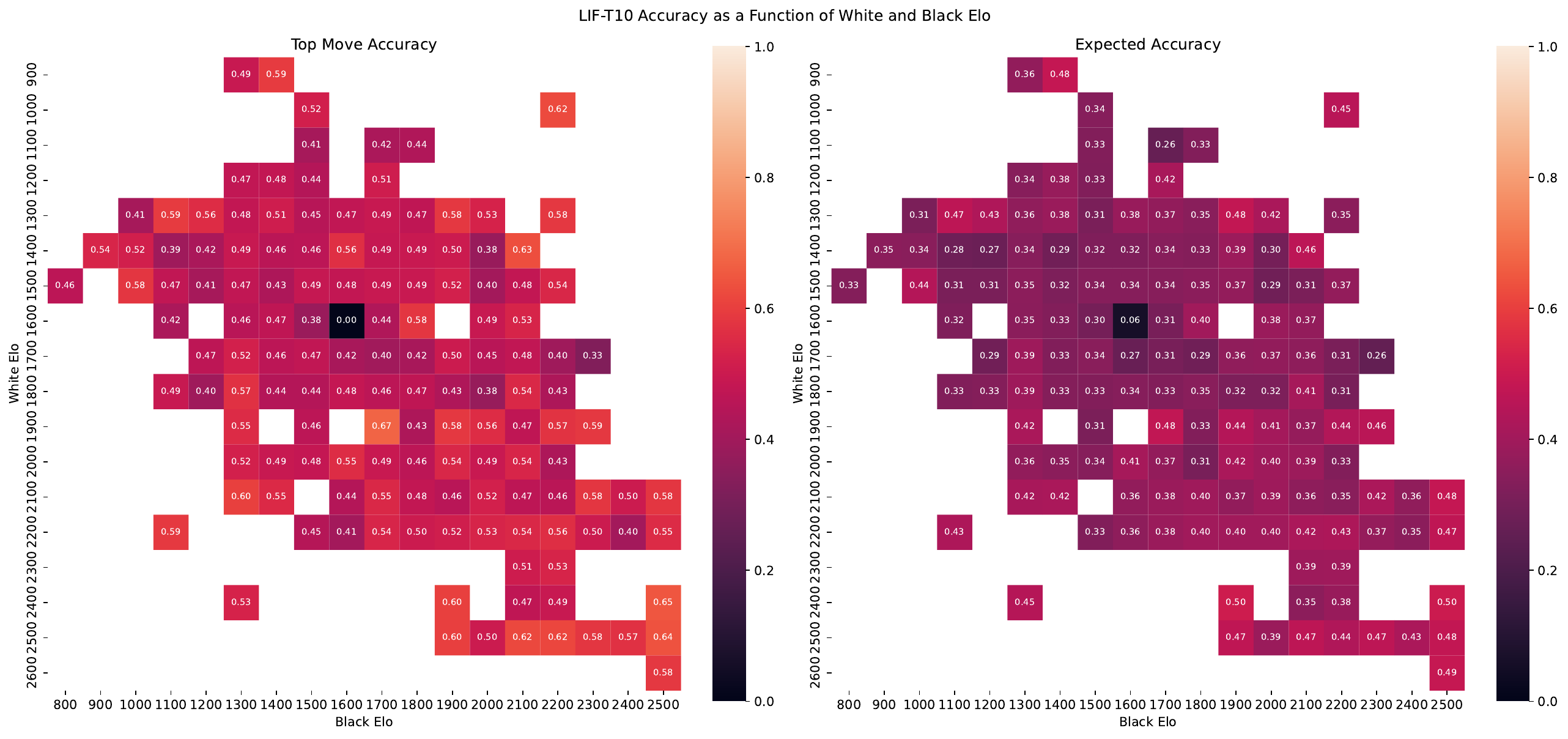}
    \caption{LIF-T10 accuracy as a function of White and Black Elo. Left: top move accuracy. Right: expected accuracy}
    \label{fig:lif-t10_accuracy_heatmap_white_and_black_elo}
\end{figure*}

\begin{figure}[h]
    \centering
    \includegraphics[width=\linewidth]{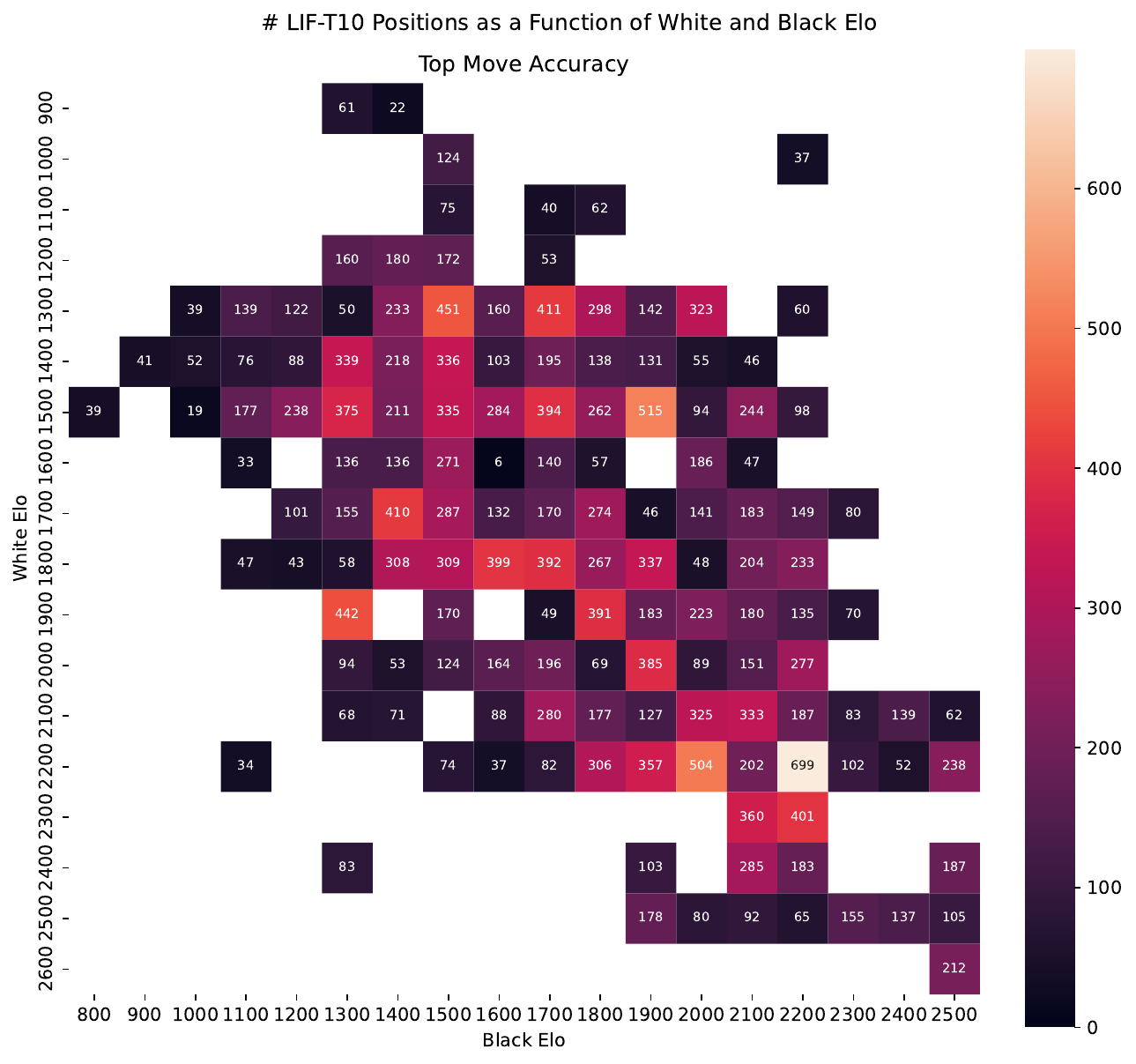}
    \caption{Number of LIF-T10 entries in each rounded White and Black Elo bucket.}
    \label{fig:lif-t10_size_heatmap_white_and_black_elo}
\end{figure}

\begin{figure*}[h]
    \centering
    \includegraphics[width=\linewidth]{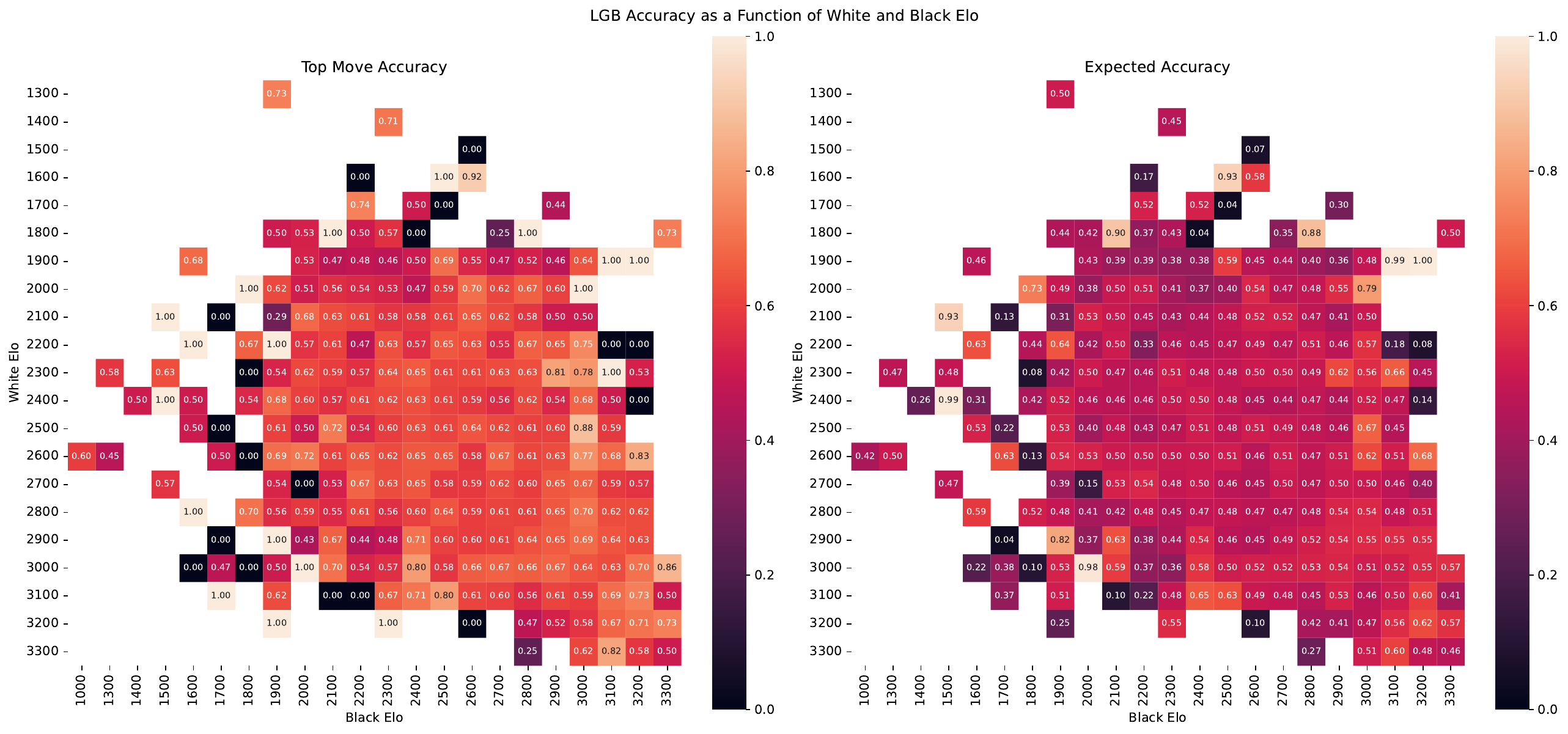}
    \caption{LGB accuracy as a function of White and Black Elo. Left: top move accuracy. Right: expected accuracy}
    \label{fig:lgb_accuracy_heatmap_white_and_black_elo}
\end{figure*}

\begin{figure}[h]
    \centering
    \includegraphics[width=\linewidth]{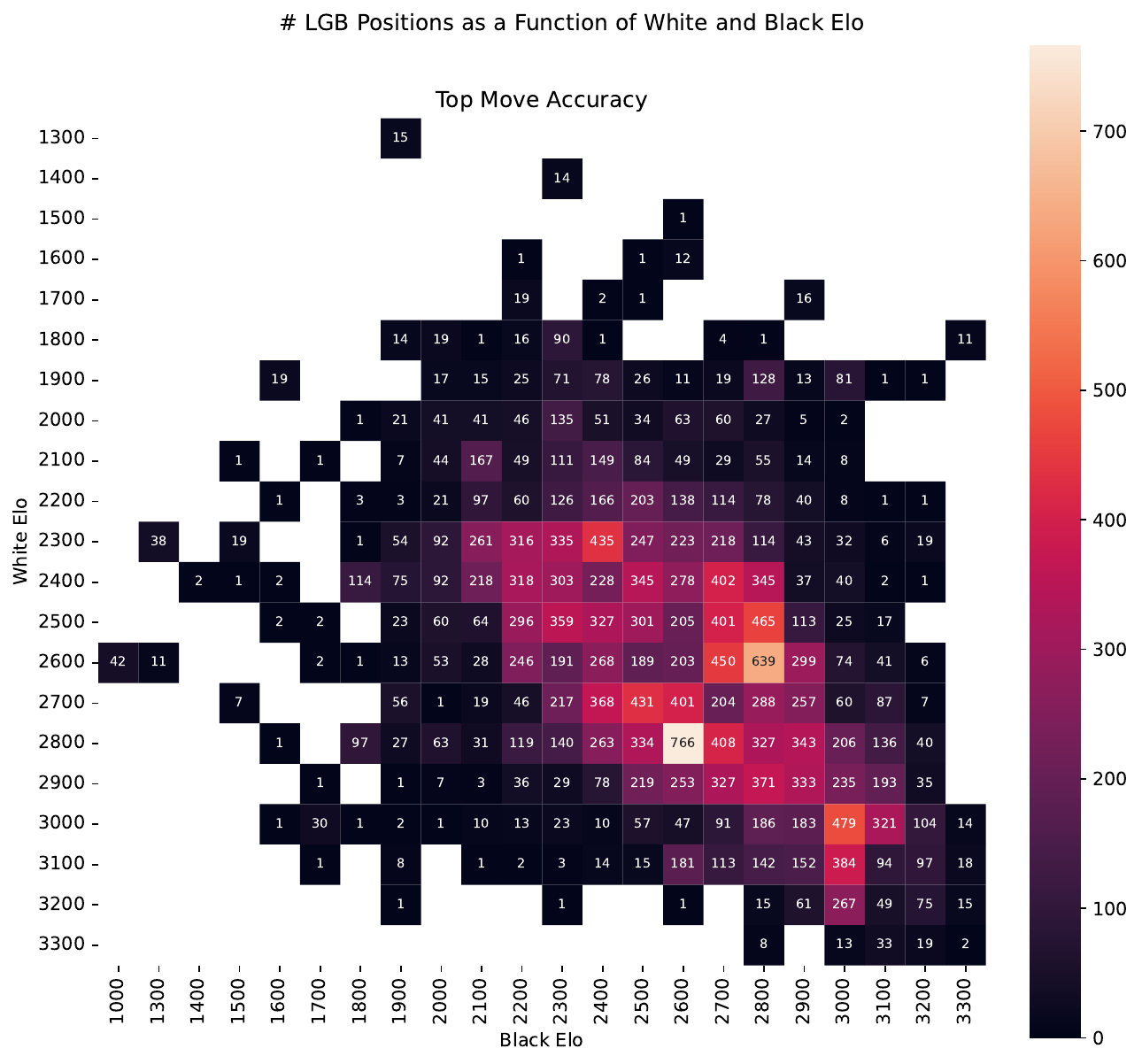}
    \caption{Number of LGB entries in each rounded White and Black Elo bucket.}
    \label{fig:lgb_size_heatmap_white_and_black_elo}
\end{figure}

\begin{figure*}[h]
    \centering
    \includegraphics[width=\linewidth]{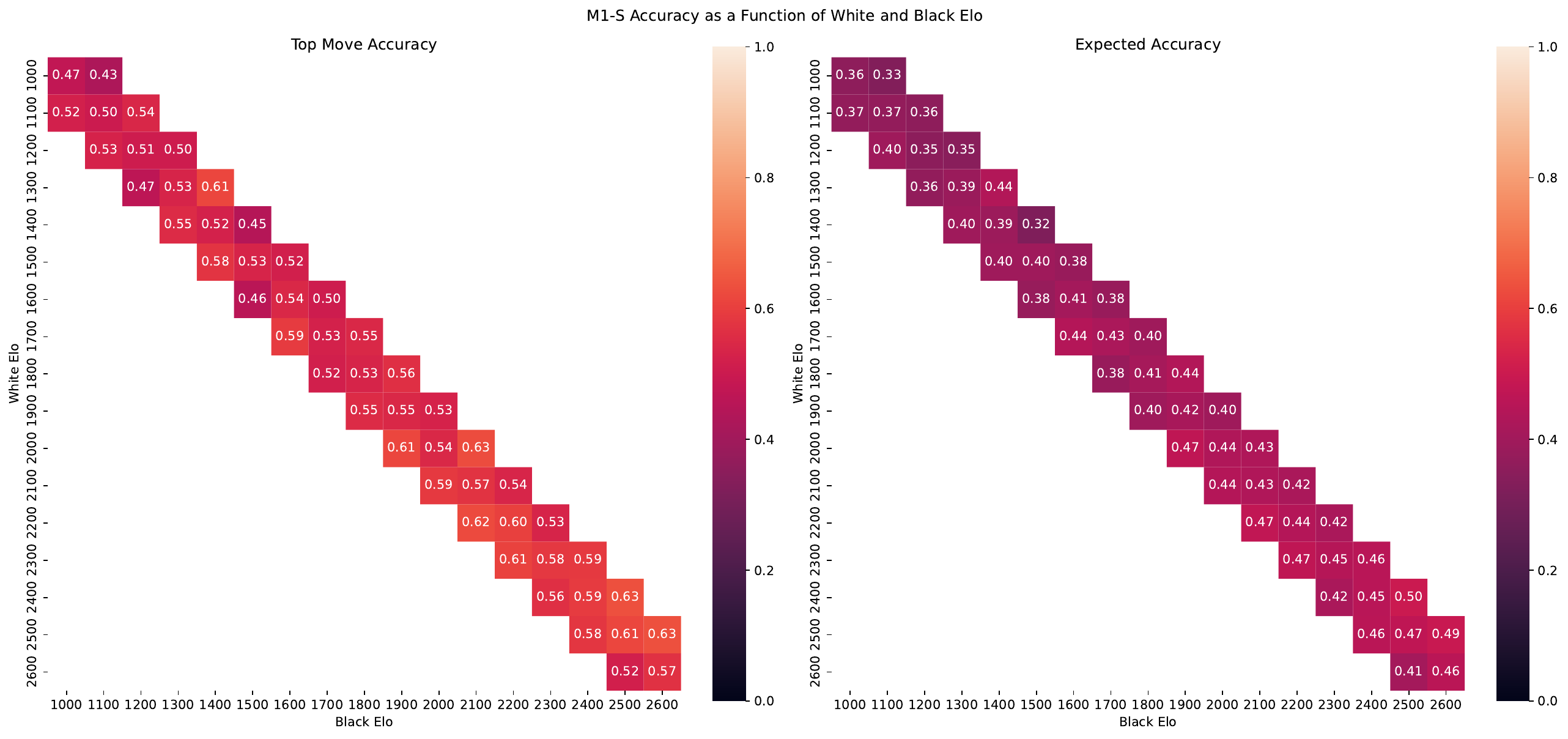}
    \caption{M1-S accuracy as a function of White and Black Elo. Left: top move accuracy. Right: expected accuracy}
    \label{fig:m1-s_accuracy_heatmap_white_and_black_elo}
\end{figure*}

\begin{figure}[h]
    \centering
    \includegraphics[width=\linewidth]{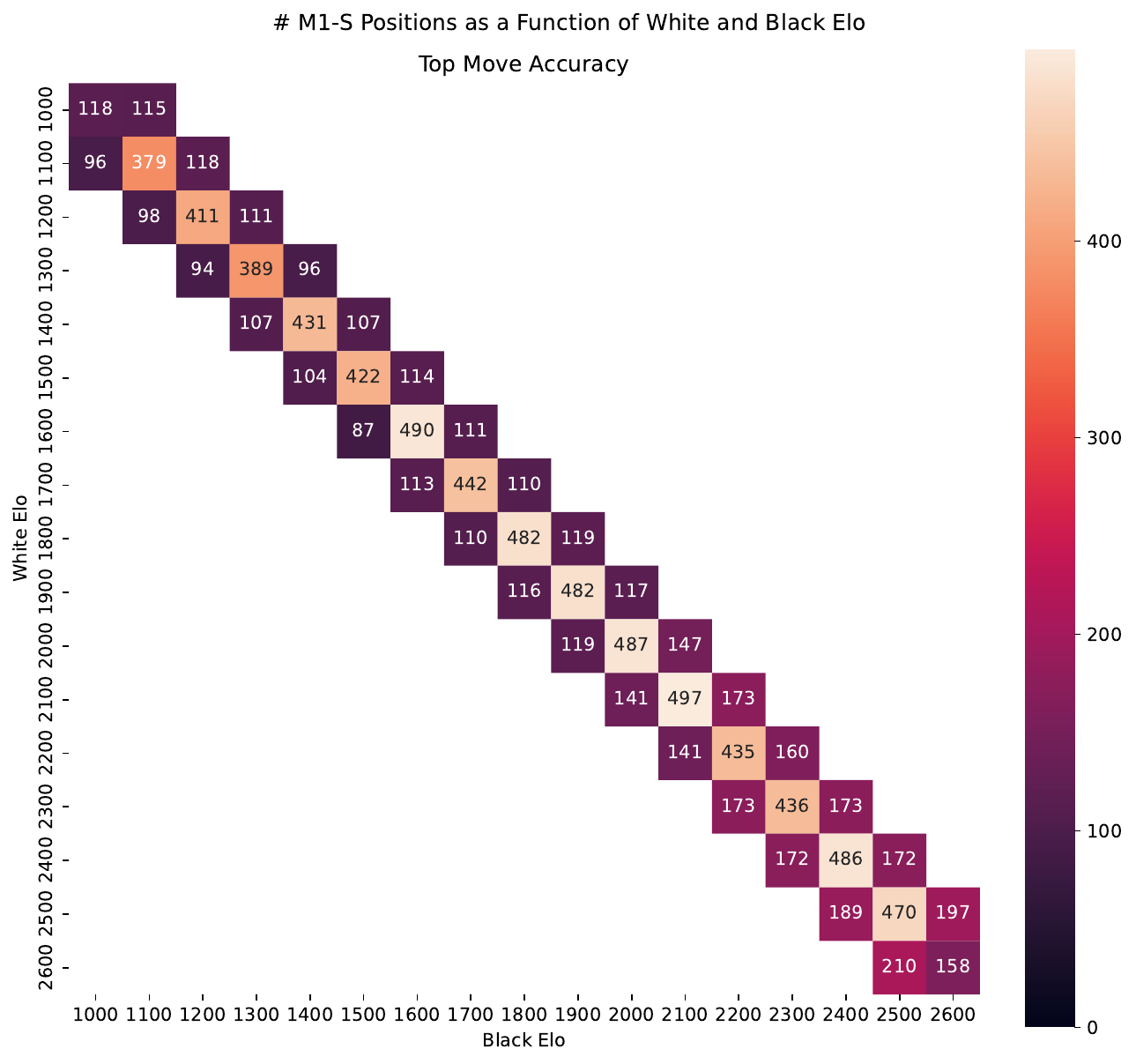}
    \caption{Number of M1-S entries in each rounded White and Black Elo bucket.}
    \label{fig:m1-s_size_heatmap_white_and_black_elo}
\end{figure}

\begin{figure*}[h]
    \centering
    \includegraphics[width=\linewidth]{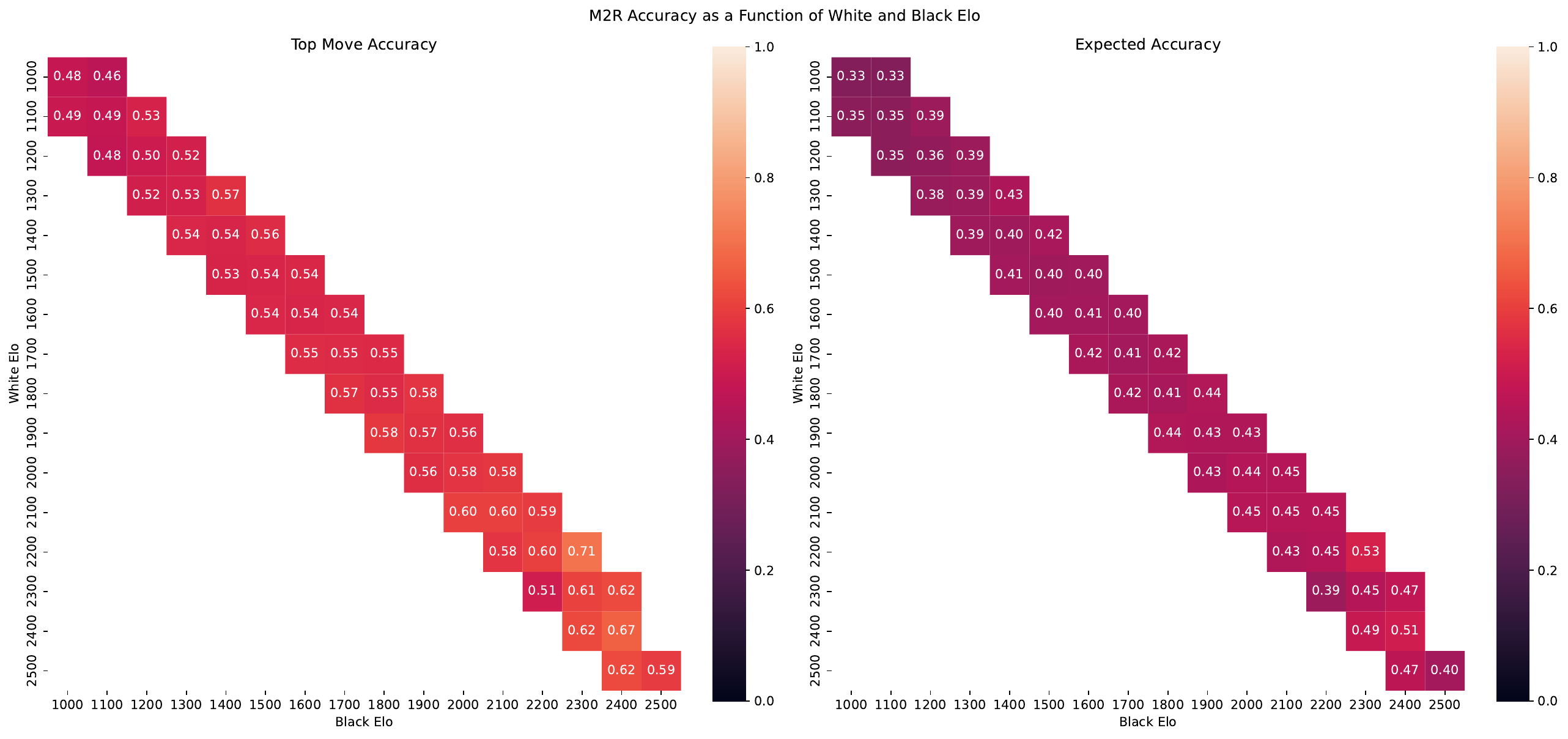}
    \caption{M2R accuracy as a function of White and Black Elo. Left: top move accuracy. Right: expected accuracy}
    \label{fig:m2r_accuracy_heatmap_white_and_black_elo}
\end{figure*}

\begin{figure}[h]
    \centering
    \includegraphics[width=\linewidth]{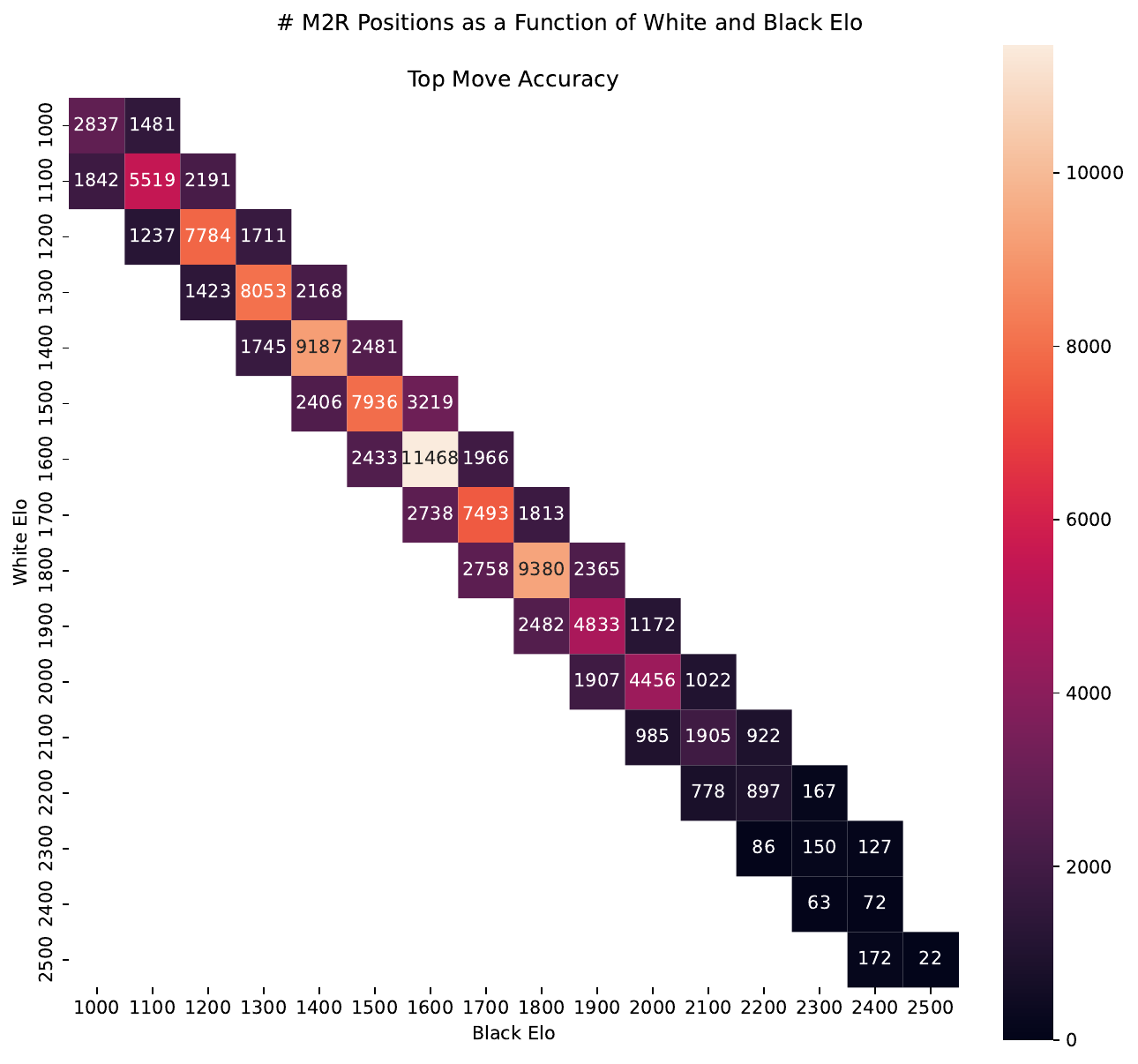}
    \caption{Number of M2R entries in each rounded White and Black Elo bucket.}
    \label{fig:m2r_size_heatmap_white_and_black_elo}
\end{figure}

%% file: appendices/appendix-future-work.tex
\section{Future Work}
\label{app:future_work}

Several directions may further improve controllable chess policy modeling.

Future work should improve the realism and robustness of the prompt distribution. Although the current pipeline uses LLMs for prompt generation and paraphrasing, it still relies on a largely fixed template set and manual validation. Incorporating natural user prompts, broader linguistic variation, and richer stylistic descriptions may improve generalization and controllability. Automated validation methods, including heuristic filters or LLM-based judges, could also help scale template verification while reducing prompt artifacts.

Another important direction is hybrid conditioning. Rather than relying solely on natural language or structured metadata, future systems could combine both. Natural language provides flexible high-level control, while structured inputs such as Elo, time control, clock usage, and player identity may improve consistency and behavioral realism. The results of UniMaia-Aux suggest that temporal and behavioral metadata are particularly promising conditioning signals.

Architecturally, the strong performance of deep fusion under full fine-tuning motivates further comparisons between ControlNet-style conditioning, deep fusion, early fusion, and hybrid approaches. Smaller embedding-focused text encoders may also provide more efficient conditioning than large language models, particularly when adapted to chess-specific corpora (Appendix~\ref{app:llm_ablations}). Other conditioning techniques, including classifier-free guidance, distilled guidance, and retrieval-augmented personalization, may further improve controllability and human move prediction.

Future work should also expand the training objective beyond next-move prediction. Bidirectional, multi-task, or self-supervised objectives could strengthen the relationship between prompts, metadata, and gameplay behavior.

Evaluation and training data may likewise be extended beyond Lichess. Incorporating over-the-board games, additional online platforms, curated engine-analysis datasets, and human evaluations would enable more robust assessment of behavioral realism and prompt controllability in real-world settings.

Finally, future interpretability work should move beyond descriptive analysis toward causal intervention. Techniques such as activation patching, residual ablations, prompt counterfactuals, and sparse autoencoders may help identify which components mediate opening control, Elo conditioning, and human move prediction, as well as how language conditioning modifies pretrained Lc0 representations. In particular, sparse autoencoders could be used to disentangle interpretable latent features within the cross-attention and policy-modulation layers, enabling analysis of how specific semantic concepts influence the pretrained chess representations.